\documentclass{article}

\usepackage[preprint]{neurips_2026}
\usepackage[utf8]{inputenc}
\usepackage[T1]{fontenc}
\usepackage{hyperref}
\usepackage{url}
\usepackage{booktabs}
\usepackage{amsfonts}
\usepackage{nicefrac}
\usepackage{microtype}
\usepackage[table]{xcolor}
\usepackage{graphicx}
\usepackage{amsmath}
\usepackage{multirow}
\usepackage{tabularx}
\usepackage{array}
\usepackage{makecell}
\usepackage{subcaption}
\usepackage{enumitem}
\usepackage{adjustbox}
\usepackage{float}

\usepackage{wrapfig}
\usepackage{fontawesome5}
\graphicspath{{figures/}}
\hypersetup{
  colorlinks=false,
  pdfborder={0 0 1},
  linkbordercolor={1 0 0},
  citebordercolor={0 1 0},
  urlbordercolor={0 1 1}
}
\title{ACE-Compression: An Ability-Centric Evaluation of Compressed Large Language Models}

\author{
Jiahui Ning\textsuperscript{1}\quad
Liang Chen\textsuperscript{2}\quad
Yuchen Liang\textsuperscript{3}\quad
Yi Zhang\textsuperscript{4}\quad
Fenglei Fan\textsuperscript{1}\\[0.35em]
{\normalfont\small\textsuperscript{1}City University of Hong Kong\quad
\textsuperscript{2}Tongji University}\\
{\normalfont\small\textsuperscript{3}Rensselaer Polytechnic Institute\quad
\textsuperscript{4}Sichuan University}
}

\begin{document}
\maketitle

\begin{abstract}
Model compression makes large language models deployable, but today's evaluation tells an incomplete story. Standard practice measures retained task performance and stops there. This misses a critical fact: compression perturbs the distributed knowledge embedded across billions of weights, and task scores alone cannot reveal which underlying abilities survive and which quietly collapse. To solve this question, we introduce ACE-Compression, an ability-centric evaluation framework. Instead of asking only “does the compressed model still answer correctly?”, we ask “which of its core abilities remain intact?” Our framework probes four deployment-facing dimensions: procedural reasoning and constraint execution, knowledge grounding and truthfulness, social cognition and affective understanding, and persona fidelity and stylistic consistency. Across more than 35 compressed checkpoints from six open-weight LLMs—covering quantization and pruning—we find that compression redistributes ability profiles. Coarse, local, and surface-level signals often persist, while fine-grained, long-tail, and context-dependent abilities become fragile. Quantization consistently outperforms pruning in retention, but the practical meaning of that gap depends entirely on which ability an application requires. Based on our findings, deployers can choose compression methods by the ability profile their application needs, not just by a leaderboard average. And algorithm designers gain diagnostic signals to build compression methods that preserve what matters. We believe that ACE-Compression turns compression from a black-box accuracy drop into an interpretable intervention—one we can measure, understand, and ultimately improve.

\end{abstract}

\section{Introduction}

Model compression has become increasingly important for serving LLMs under memory, latency, and cost constraints \citep{frantar2022gptq,lin2023awq,frantar2023sparsegpt,sun2024wanda,zhu2024survey}. The general goal of model compression is to preserve the original model's behavior, which raises the question of how to evaluate compressed LLMs at multiple levels \citep{zhu2024survey}. Current LLM evaluation can be viewed through two complementary lenses: task evaluation and ability evaluation. Task evaluation asks whether a model can complete concrete tasks under standardized benchmarks and metrics, providing a practical criterion for comparing model quality \citep{liang2023helm,srivastava2023beyond}. Ability evaluation, in contrast, seeks to identify the underlying capabilities that support task performance, such as reasoning, knowledge grounding, instruction following, truthfulness, social understanding, and persona consistency \citep{ye2024flask,cui2026measuring}. These two levels are not equivalent: a task score may reflect a mixture of abilities, and the same ability may contribute to multiple tasks.

LLMs store learned information distributively across their weights, while model compression introduces local or global perturbations to the weights. Compression can therefore change not only how the model performs on an individual task, but also how ability is shaped in a model. Modern post-training compression methods perturb the original parameters only mildly. We therefore conjecture that the compressed model might retain decent accuracy on a benchmark task while losing a core ability (e.g., multi-step reasoning or counterfactual understanding) that the task implicitly relied on. Thus, evaluating compressed LLMs should assess not only how much task performance is retained but also which underlying abilities remain intact after compression. Moreover, such joint diagnostics are useful for algorithmic improvement, since observed failure patterns can guide where compression methods should preserve sensitive weights, adjust sparsity patterns, use more targeted calibration data, or apply post-compression repair.

Recent compression evaluation studies have compared efficiency, task performance, and reliability across different model families and different compression methods such as quantization, pruning, and distillation \citep{li2024evaluating,yang2024llmcbench,vonrad2026unicomp}. In this work, we propose \textbf{ACE-Compression} (Ability-Centric Evaluation for Large Model Compression) that instantiates this ability-centric view with four deployment-facing dimensions: \textbf{(i) Procedural reasoning and constraint execution} captures whether a model can recognize requirements, coordinate constraints, and realize them in answerable or executable form. \textbf{(ii) Knowledge grounding and truthfulness} captures whether a model can preserve factual knowledge, ground responses in the relevant context, and avoid truth-sensitive failures. \textbf{(iii) Social cognition and affective understanding} captures whether a model can interpret beliefs, intentions, emotional cues, ambiguity, and socially situated meanings. \textbf{(iv) Persona fidelity and stylistic consistency} captures whether a model can maintain role-grounded behavior, character consistency, and stable stylistic or persona traits. These dimensions are not intended as a complete ontology of language-model abilities; rather, they provide a diagnostic lens for capabilities \begin{wrapfigure}{r}{0.4\textwidth}
    \centering
    \vspace{-1em}  
    \includegraphics[width=\linewidth]{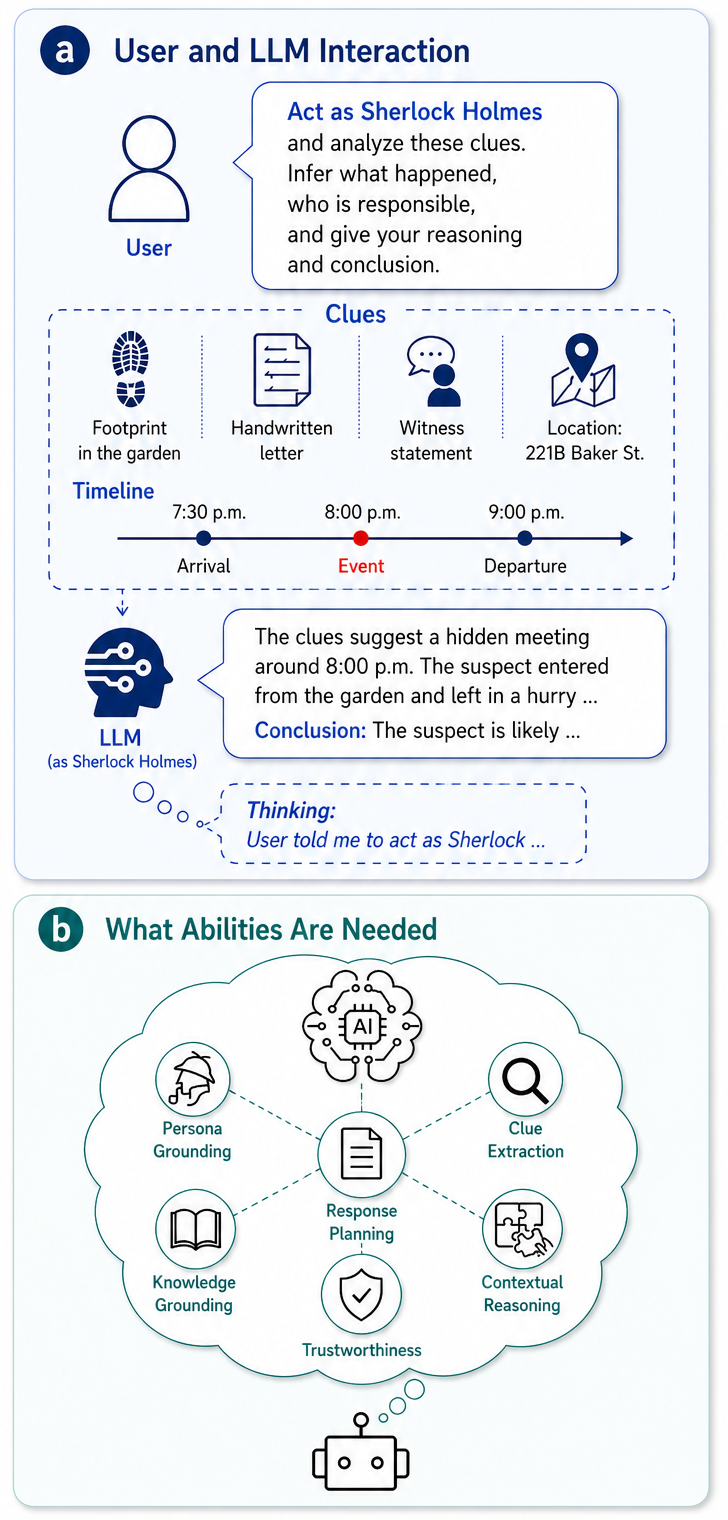}
    \captionsetup{font=scriptsize}
    \caption{Motivating scenario for ability-centric compression evaluation. A single role-grounded request requires the LLM to interact with abilities including persona grounding, clue extraction, response planning, knowledge grounding, contextual reasoning, and trustworthiness.}
    \label{fig:motivation}
    \vspace{-4em}
\end{wrapfigure}that are often entangled in realistic deployment scenarios. Figure~\ref{fig:motivation} corroborates this taxonomy. A seemingly simple role-grounded interaction requires multiple abilities: constraint following, contextual reasoning, factual grounding, and persona maintenance. This example also motivates our ability-centric view: benchmarks should be interpreted as probes of the abilities that support task behavior, rather than as isolated task scores.

\begin{figure*}[t]
    \centering
    \includegraphics[width=0.995\textwidth]{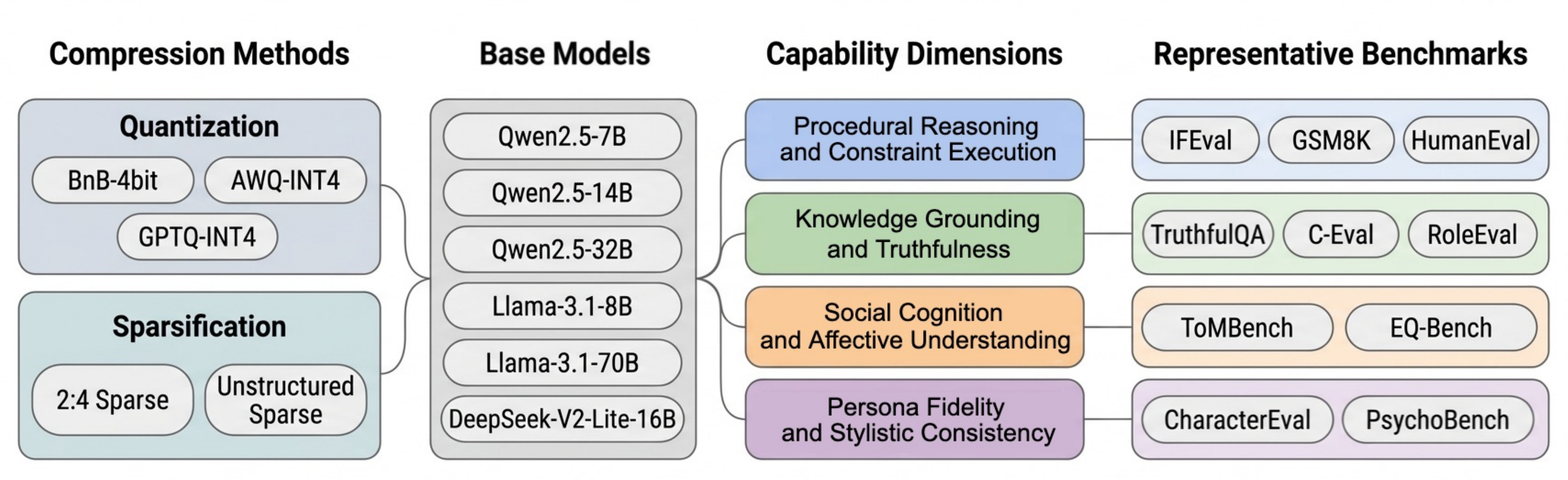}
    \caption{Evaluation design. We compare quantization methods (BnB-4bit, AWQ-INT4, GPTQ-INT4) and pruning methods (2:4 structured sparsity and unstructured sparsity) across six base models. Ten benchmarks are mapped to four ability dimensions so that pre/post-compression changes can be read as procedural, knowledge, social, or persona profile shifts.}
    \label{fig:setup}
    \vspace{-0.4cm}
\end{figure*}

Furthermore, as Figure~\ref{fig:setup} summarizes, we choose 6 representative open-weight LLMs including Qwen2.5, Llama-3.1 and Deepseek-V2-Lite checkpoints from the 7B range to 70B-scale models. We focus on mainstream post-training compression methods for efficient LLM deployment, covering both weight quantization and sparsity-based compression. We exclude distillation because it entangles compression with new architectures, supervision, data, and objectives, making capability preservation difficult to attribute to the FP16 parent \citep{hinton2015distilling,sanh2019distilbert,gu2024minillm,hsieh2023distilling}. We select ten benchmarks not as independent tasks, but as diagnostic probes for the four deployment-facing ability dimensions, which allows us to connect task-level measurements to ability-level changes. By combining task scores with benchmark-native readouts, we obtain a more comprehensive view of how compression reshapes each model's ability profile. Our key findings are 
\begin{itemize}
    \item Compression redistributes ability profile rather than causing a uniform performance decline.
    \item Coarse and surface-facing abilities are more stable than fine-grained and context-dependent abilities.
    \item Quantization is generally more resilient to ability degradation than pruning.
\end{itemize}

This paper makes three contributions. {1)} By interpreting benchmark outcomes as evidence about underlying capability profiles, we propose a comprehensive and systematic ability-centric protocol for evaluating pre- and post-compression changes in LLMs, organized around four deployment-facing dimensions. {2)} In ACE-Compression, we evaluate more than 35 model checkpoints across 10 diagnostic benchmarks, covering over 25,000 questions, 1,785 multi-turn role-playing dialogues, and 13 psychometric scales. {3)} Our findings shall provide actionable signals for compressed-model deployment and compression algorithm refinement for developers and practitioners. 

\section{Related Work}
\label{sec:related}

\paragraph{Holistic and fine-grained LLM evaluation.}
In recent years, a plethora of work motivates the evaluation of LLMs at the scenario, skill, and ability level. BIG-bench shows that aggregate task performance can conceal qualitatively different scaling behavior across diverse tasks \citep{srivastava2023beyond}. HELM argues for broad scenario coverage, multi-metric measurement, and standardized reporting when studying capabilities, limitations, and risks \citep{liang2023helm}. FLASK decomposes coarse instruction-following judgments into alignment skill sets, making model behavior easier to interpret \citep{ye2024flask}. Task-specific benchmarks provide native ability readouts that are useful for this purpose: IFEval probes formal instruction following, ToMBench exposes theory-of-mind tasks and subabilities, CharacterEval measures role-playing quality, and PsychoBench measures psychological profile expression \citep{zhou2023ifeval,chen2024tombench,tu2024charactereval,huang2024psychobench}. To the best of our knowledge, ACE-Compression is the first to apply this fine-grained perspective to model compression, analyzing how ability profiles move from FP16 parent models to compressed variants.

\paragraph{LLM compression evaluation.}
Post-training compression evaluation work has gradually moved from isolated score reporting toward systematic evaluation of retained performance, efficiency, and reliability. Quantization studies show that compression tolerance varies across tensor types, model scales, and task families, with multi-step reasoning often more fragile than simpler instruction-following readouts \citep{li2024evaluating}. LLMCBench expands the empirical scope by benchmarking compressed LLMs with deployment-oriented performance and efficiency measurements \citep{yang2024llmcbench}. UniComp is capability-aware in benchmark coverage, but method-centered in evaluation design: its goal is to compare pruning, quantization, and distillation across performance, reliability, and efficiency\citep{vonrad2026unicomp}. Although ACE-Compression shares probes such as GSM8K and TruthfulQA with UniComp, this overlap reflects common reference points in LLM evaluation rather than the organizing principle of our study. To the best of our knowledge, ACE-Compression is the first LLM compression evaluation designed from the outset to be ability-centric.

\section{ACE-Compression Setup}
\label{sec:ACE-Compression}

\subsection{Study Panel and Compression Settings}

We evaluate 6 chat- or instruction-tuned base models: Qwen2.5-7B, Qwen2.5-14B, Qwen2.5-32B, Llama-3.1-8B, Llama-3.1-70B, and DeepSeek-V2-Lite-16B. The core compression panel covers five mainstream methods: BitsAndBytes 4-bit quantization, AWQ-INT4, GPTQ-INT4, 2:4 structured sparsity, and unstructured sparsity. Overall, the study covers more than 35 model checkpoints; the appendix lists the exact model compression combinations.

For quantization, BnB-4bit is applied on the fly at inference time from the dense checkpoint, while AWQ and GPTQ variants are loaded from officially released quantized checkpoints in community \citep{dettmers2023qlora,lin2023awq,frantar2022gptq}. For pruning, we create sparse models from the corresponding FP16 checkpoints with one-shot SparseGPT pruning \citep{frantar2023sparsegpt}.  SparseGPT can be viewed as a layer-wise reconstruction method: given a dense weight matrix \(W\) and calibration activations \(X\), it searches for a sparse matrix \(\widetilde{W}\) that preserves the original layer output under a target sparsity constraint,
\begin{equation}
    \min_{\widetilde{W}\in \mathcal{S}} \left\|WX-\widetilde{W}X\right\|_F^2,
\end{equation}
where \(\mathcal{S}\) denotes the target sparsity pattern. In our study, we use two instantiations of \(\mathcal{S}\): The 2:4 structured sparsity constrains each consecutive group of four weights in the pruning dimension to retain two weights and zero out the other two; the unstructured sparsity imposes only the same overall sparsity ratio, allowing zeros to be placed at arbitrary weight positions selected by the SparseGPT reconstruction criterion. We mainly prune the main attention and MLP projection matrices to 50\% sparsity. We obtain \(X\) from calibration samples drawn from \texttt{yahma/alpaca-cleaned}, a cleaned public Alpaca-style instruction dataset derived from the Self-Instruct pipeline \citep{wang2023selfinstruct}.

All model states are evaluated on NVIDIA RTX PRO 6000 Blackwell GPUs through a unified \texttt{vLLM} backend \citep{kwon2023vllm}. Benchmarks are executed through \texttt{lm-eval} and custom OpenAI-compatible API wrappers, which share the same serving stack. This keeps hardware and serving conditions consistent across dense, quantized, and sparse model states.

\subsection{Ability Taxonomy and Benchmark Mapping}

The ACE-Compression dimensions are designed to connect fine-grained LLM evaluation with the central question under compression: which abilities are preserved, weakened, or redistributed after compression. This follows holistic evaluation work showing that aggregate scores can hide heterogeneous behavior across scenarios, tasks, and metrics \citep{srivastava2023beyond,liang2023helm}. In the compression setting, this view is especially important because a compressed model may preserve some task outcomes while weakening the abilities that support them. 
Prior work such as FLASK decomposes open-ended instruction following into multiple skill components, including logical thinking, background knowledge, problem handling, and user alignment, which is consistent with the ability-oriented view that LLM evaluation should look beyond aggregate task outcomes \citep{ye2024flask}. Moreover, recent LLM evaluation work has increasingly examined abilities such as social cognition, role-playing behavior, and psychometric profiling \citep{chen2024tombench,paech2023eqbench,tu2024charactereval,huang2024psychobench}. We therefore organize ACE-Compression around four diagnostic ability dimensions: \textbf{Procedural reasoning and constraint execution}, \textbf{Knowledge grounding and truthfulness}, \textbf{Social cognition and affective understanding}, and \textbf{Persona fidelity and stylistic consistency}.

Table~\ref{tab:benchmark_summary} specifies how ACE-Compression groups benchmarks into four ability dimensions and which benchmark-native readouts are retained for analysis.  \textbf{Procedural} evaluation combines benchmarks that stress different stages of execution: IFEval probes explicit constraint satisfaction, GSM8K probes multi-step answer realization, and HumanEval probes executable output generation. The \textbf{knowledge} dimension is designed to separate knowledge retention from knowledge-grounded output behavior. C-Eval represents broad disciplinary knowledge, while RoleEval provides a representative probe of long-tail knowledge. TruthfulQA tests whether retained knowledge is expressed as trustworthy output. For \textbf{social} cognition, ToMBench provides the primary diagnostic signal, while EQ-Bench supplies auxiliary evidence about affective understanding under socially grounded dialogue contexts. The \textbf{persona} dimension combines an outward-facing role-play view with an inward-facing psychometric profile view. CharacterEval captures visible role enactment, character consistency, and stylistic continuity, while PsychoBench provides a questionnaire-based psychometric view of the model's psychological portrayal.

The ACE-Compression mapping is intentionally compact and diagnostic. We do not treat any benchmark as a pure test of a single ability; instead, each benchmark is used as evidence for a behaviorally meaningful element within a broader ability profile. For this reason, ACE-Compression preserves benchmark-native readouts.

\begin{table*}[t]
\centering
\scriptsize
\setlength{\tabcolsep}{4pt}
\renewcommand{\arraystretch}{1.12}
\definecolor{ACEHeader}{HTML}{E8EEF4}
\definecolor{ACEHeaderText}{HTML}{1F2D3D}
\definecolor{ProcBg}{HTML}{F5FAFF}
\definecolor{KnowBg}{HTML}{F7FBF3}
\definecolor{SocialBg}{HTML}{FFF8F0}
\definecolor{PersonaBg}{HTML}{FAF6FD}
\definecolor{ProcAccent}{HTML}{2F75B5}
\definecolor{KnowAccent}{HTML}{5B8C3A}
\definecolor{SocialAccent}{HTML}{C87522}
\definecolor{PersonaAccent}{HTML}{7B5EA7}
\definecolor{MetricBg}{HTML}{FFFFFF}
\definecolor{MetricBorder}{HTML}{C9D3DC}
\definecolor{MetricText}{HTML}{263238}
\definecolor{ACERule}{HTML}{B8C2CC}
\definecolor{ACEInnerRule}{HTML}{BFC8D2}
\newcommand{\procicon}{\textcolor{ProcAccent}{\faTasks}}
\newcommand{\knowicon}{\textcolor{KnowAccent}{\faBookOpen}}
\newcommand{\socialicon}{\textcolor{SocialAccent}{\faUsers}}
\newcommand{\personaicon}{\textcolor{PersonaAccent}{\faUserTag}}
\newcommand{\metric}[1]{\begingroup\setlength{\fboxsep}{1.25pt}\colorbox{MetricBg}{\textcolor{MetricText}{\scriptsize\textbf{#1}}}\endgroup}
\newcommand{\dimlabel}[2]{#1\hspace{0.45em}\textbf{#2}}
\newcommand{\benchcite}[2]{\mbox{\textbf{#1}~\citep{#2}}}
\newcommand{\diagrow}[3]{\cellcolor{#1}#2 & \cellcolor{#1}#3 \\}
\newcommand{\aceTopRule}{\arrayrulecolor{ACEHeaderText}\toprule[0.55pt]\arrayrulecolor{ACERule}}
\newcommand{\aceHeadRule}{\arrayrulecolor{ACEHeaderText}\midrule[0.45pt]\arrayrulecolor{ACERule}}
\newcommand{\aceBlockRule}{\arrayrulecolor{ACEInnerRule}\specialrule{0.32pt}{1.2pt}{1.2pt}\arrayrulecolor{ACERule}}
\newcommand{\aceBottomRule}{\arrayrulecolor{ACEHeaderText}\bottomrule[0.55pt]\arrayrulecolor{black}}
\newcommand{\aceDiagHead}{%
    \begingroup
    \setlength{\tabcolsep}{4pt}%
    \begin{tabularx}{\linewidth}{@{}>{\raggedright\arraybackslash}p{0.34\linewidth} X@{}}
    \textbf{Benchmark} & \textbf{Main Diagnostic Readout}
    \end{tabularx}%
    \endgroup
}
\newcommand{\diagtable}[1]{%
    \begingroup
    \renewcommand{\arraystretch}{1.16}%
    \setlength{\tabcolsep}{4pt}%
    \begin{tabularx}{\linewidth}{@{}>{\raggedright\arraybackslash}p{0.34\linewidth} X@{}}
    #1
    \end{tabularx}%
    \endgroup
}

\caption{Ability dimensions, benchmarks, and native readouts in ACE-Compression. We preserve benchmark-native views when a single aggregate score would hide diagnostically important distinctions.}
\label{tab:benchmark_summary}
\begin{tabularx}{\textwidth}{@{} >{\raggedright\arraybackslash}m{0.25\textwidth} >{\raggedright\arraybackslash}p{0.15\textwidth} X @{}}
\aceTopRule
\textbf{Dimension} &
\multicolumn{2}{@{}p{0.73\textwidth}@{}}{\aceDiagHead} \\
\aceHeadRule
\dimlabel{\procicon}{Procedural reasoning\newline and constraint execution} &
\multicolumn{2}{@{}>{\columncolor{ProcBg}}p{0.73\textwidth}@{}}{\diagtable{
    \diagrow{ProcBg}{\benchcite{IFEval}{zhou2023ifeval}}{Prompt-/instruction-level strict and loose compliance}
    \diagrow{ProcBg}{\benchcite{GSM8K}{cobbe2021gsm8k}}{Strict versus flexible answer matching}
    \diagrow{ProcBg}{\benchcite{HumanEval}{chen2021humaneval}}{\metric{pass@k} and sampling recoverability}
}} \\
\aceBlockRule
\dimlabel{\knowicon}{Knowledge grounding\newline and truthfulness} &
\multicolumn{2}{@{}>{\columncolor{KnowBg}}p{0.73\textwidth}@{}}{\diagtable{
    \diagrow{KnowBg}{\benchcite{TruthfulQA}{lin2022truthfulqa}}{\metric{MC1/MC2} and generation-side \metric{BLEURT} behavior}
    \diagrow{KnowBg}{\benchcite{C-Eval}{huang2023ceval}}{Accuracy retention}
    \diagrow{KnowBg}{\benchcite{RoleEval}{shen2023roleeval}}{Role-grounded knowledge retention over shot and language slices}
}} \\
\aceBlockRule
\dimlabel{\socialicon}{Social cognition and\newline affective understanding} &
\multicolumn{2}{@{}>{\columncolor{SocialBg}}p{0.73\textwidth}@{}}{\diagtable{
    \diagrow{SocialBg}{\benchcite{ToMBench}{chen2024tombench}}{\metric{ToM profiles} across task, ability, and subability views}
    \diagrow{SocialBg}{\benchcite{EQ-Bench}{paech2023eqbench}}{Benchmark score and \metric{parseability} as auxiliary affective evidence}
}} \\
\aceBlockRule
\dimlabel{\personaicon}{Persona fidelity and\newline stylistic consistency} &
\multicolumn{2}{@{}>{\columncolor{PersonaBg}}p{0.73\textwidth}@{}}{\diagtable{
    \diagrow{PersonaBg}{\benchcite{CharacterEval}{tu2024charactereval}}{\metric{role-play quality} across grouped persona metrics}
    \diagrow{PersonaBg}{\benchcite{PsychoBench}{huang2024psychobench}}{\metric{psychometric drift} and \metric{16P} dominant-type stability}
}} \\
\aceBottomRule
\end{tabularx}

\vspace{-0.1cm}
\end{table*}

For higher-is-better readouts in Table~\ref{tab:benchmark_summary}, we report retention relative to the corresponding FP16 parent.
\begin{equation}
R_{m,b}=S^{\mathrm{comp}}_{m,b}/S^{\mathrm{fp16}}_{m,b},
\end{equation}
where \(S^{\mathrm{comp}}_{m,b}\) is the score of compressed model \(m\) on benchmark or metric \(b\), and \(S^{\mathrm{fp16}}_{m,b}\) is the score of its FP16 parent. Thus, \(R=1\) indicates full retention, \(R<1\) indicates degradation, and \(R>1\) indicates a relative gain under the same evaluation protocol. PsychoBench is handled separately: its questionnaire responses define psychometric profiles rather than higher-is-better scores, and compression effects are measured by how these profiles shift from the FP16 parent.

\section{Results and Findings}
\label{sec:results}

\begin{figure*}[t]
    \centering
    \includegraphics[width=0.995\textwidth]{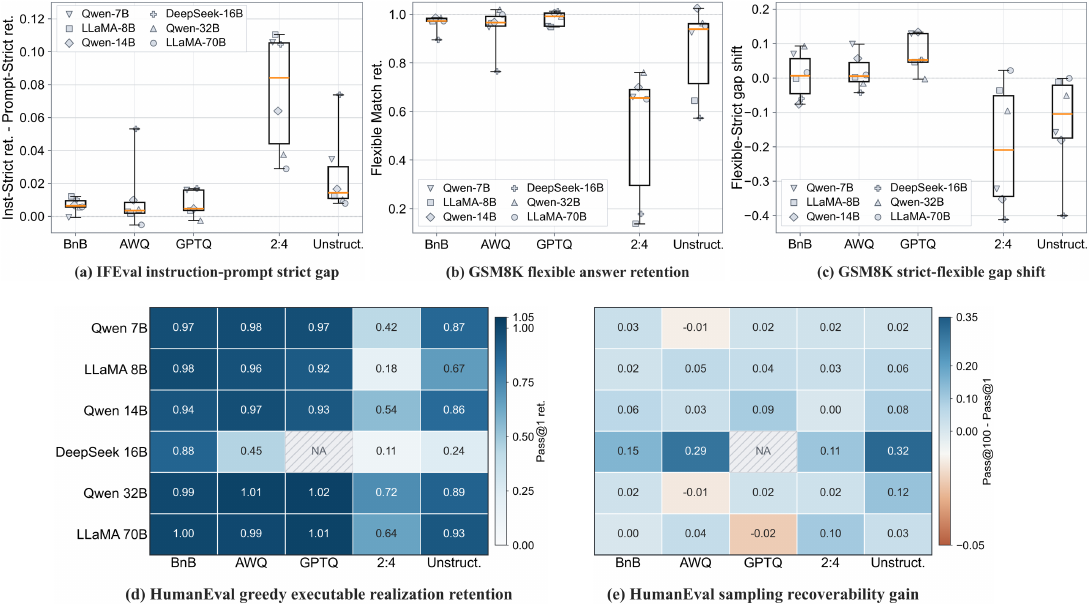}
    \caption{Procedural reasoning and constraint execution. (a) IFEval instruction-level strict retention minus prompt-level strict retention, diagnosing local constraint recognition versus full-response coordination. (b) GSM8K flexible-match retention. (c) GSM8K flexible-versus-strict retention gap shift, diagnosing answer-form brittleness. (d) HumanEval greedy pass@$1$ retention by base model and compression method. (e) HumanEval pass@$100$ minus pass@$1$ retention, diagnosing sampling-based recoverability; hatched cells mark unavailable combinations.}
    \label{fig:procedural}
    \vspace{-0.4cm}
\end{figure*}

\subsection{Procedural Reasoning and Constraint Execution}

The procedural dimension separates three related questions: whether compressed models could follow local constraints, whether multi-step reasoning still reaches the intended final answer, and whether the model can realize procedural competence as executable outputs. There is a cascade from local requirement recognition to multi-step answer realization and executable output generation.

\textbf{IFEval} separates local instruction recognition from full-response constraint coordination. In Figure~\ref{fig:procedural}(a), instruction-level strict retention usually exceeds prompt-level strict retention across methods. This indicates that local requirements tend to survive longer than their joint coordination in a full response. Compression, therefore, often breaks global constraint satisfaction before it erases recognition of individual instructions. The size of this gap also varies across model families: under the same compression setting, local compliance can remain strong for one base model, whereas full-response control can weaken more clearly for another.

As shown in Figure~\ref{fig:procedural}(b,c), we check the flexible-match retention and the gap between strict-match and flexible-match. The flexible-match view shows that \textbf{GSM8K} retention is large under quantization but becomes more fragile under pruning. The flexible–strict gap-shift view measures whether compression changes \emph{answer correctness} and \emph{answer realization} at the same time: under pruning, strict-match directly extracts the number following a  specific answer marker, while flexible extraction searches more broadly over generated numeric strings and may select a later or irrelevant number instead. This reversal therefore suggests reduced consistency between answer recovery and canonical final-answer realization under pruning.

\textbf{HumanEval} evaluates the executable-output side of procedural ability. Pass@1 reflects first-sample executable consistency, while pass@100 indicates whether valid programs remain accessible under repeated sampling. In Figure~\ref{fig:procedural}(d), quantization largely preserves first-sample success, whereas pruning causes larger drops. Figure~\ref{fig:procedural}(e) further shows how much additional executable success remains under sampling beyond the first decoded program. These two panels indicate that compression changes the first decoded solution while still preserving part of the valid program space under sampling.

\subsection{Knowledge Grounding and Truthfulness}

\begin{figure*}[t]
    \centering
    \includegraphics[width=0.995\textwidth]{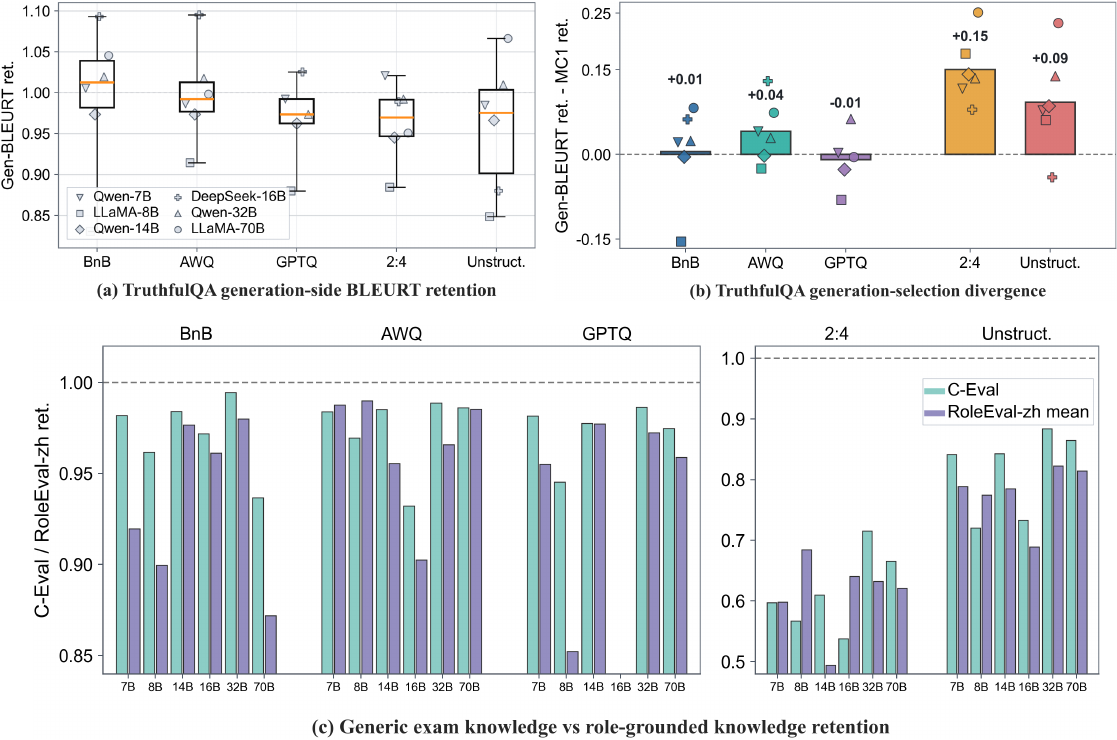}
    \caption{Knowledge grounding and truthfulness. (a) TruthfulQA generation-side BLEURT-accuracy retention. (b) TruthfulQA generation-selection divergence, computed as generation-side BLEURT-accuracy retention minus MC1 retention. (c) Generic knowledge versus role-grounded knowledge retention, using C-Eval and RoleEval-ZH. }
    \label{fig:knowledge}
    \vspace{-0.5cm}
\end{figure*}

The knowledge dimension separates three related questions: whether broad knowledge is retained, whether long-tail knowledge remains undisrupted, and whether retained knowledge translates into truthful output behavior.

Comparing \textbf{C-Eval} with \textbf{RoleEval} shows that broad disciplinary knowledge is usually retained better than role-specific long-tail knowledge. In Figure~\ref{fig:knowledge}(c), C-Eval retention stays above RoleEval-Zh retention across most models and compression settings. The gap is modest under quantization but widens under sparse compression, where RoleEval declines more steeply. This suggests that role-specific long-tail knowledge is more fragile than broad disciplinary knowledge.

\textbf{TruthfulQA} asks whether retained knowledge is expressed as truthful behavior. BLEURT is a learned reference-based metric for text-generation evaluation \citep{sellam2020bleurt}. We read BLEURT retention as generation-side reference compatibility, and MC1 retention as constrained truthful selection. Figure~\ref{fig:knowledge}(a) shows that generation-side BLEURT retention remains relatively high across compression settings, while Figure~\ref{fig:knowledge}(b) shows that the generation--selection gap is small under quantization but becomes positive under sparse compression. This phenomenon should not be read as improved truthfulness. Lower MC1 retention shows that constrained truthful selection has weakened. Thus the positive generation--selection gap suggests a shift toward reference-style, BLEURT-friendly generation after compression, rather than a genuine improvement in truthfulness.

\subsection{Social Cognition and Affective Understanding}

\begin{figure*}[t]
    \centering
    \includegraphics[width=0.995\textwidth]{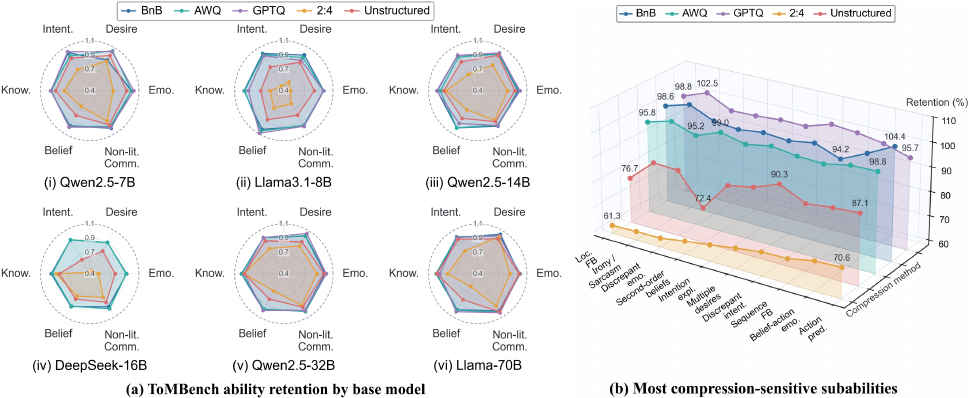}
    \caption{Social cognition and affective understanding. (a) ToMBench ability retention by base model, with one radar plot per FP16 parent and compression-method traces relative to that parent. (b) Most compression-sensitive ToMBench subabilities, averaged by method and displayed as retention percentages; abbreviated labels denote subabilities such as location false beliefs, irony/sarcasm, second-order beliefs, intentions explanations, and belief-action links. Compression primarily affects hidden-state, belief-, intention-, and ambiguity-centered reasoning.}
    \label{fig:social}
    \vspace{-0.5cm}
\end{figure*}

The social dimension is designed to diagnose how much social-cognitive performance is retained after compression. Compression does not affect all social-cognitive signals to the same degree. Quantization remains relatively stable overall, but the quantization--pruning gap is less pronounced than in the procedural and knowledge dimensions.

\textbf{ToMBench} exposes compression effects at task-, ability-, and subability-level views rather than only as a single aggregate ToM score, so we use them as the main diagnostic for compression. In Figure~\ref{fig:social}(a), compression does not reduce belief, intention, knowledge, emotion, and non-literal communication by the same amount; the weakest axes vary by models and compression methods. As Figure~\ref{fig:social}(b) shows, the subability panel makes this unevenness more explicit: the largest losses cluster around hidden mental states and socially implied meaning, including \emph{location false beliefs}, \emph{second-order beliefs}, \emph{intention explanations}, \emph{belief-action links}, and \emph{irony/sarcasm}. Together, these results show that social-cognitive degradation under compression is uneven and local.

\textbf{EQ-Bench} manifests a different failure signal. Its overall benchmark score often stays close to the FP16 baseline, but parseability declines: some compressed models stop answering within the emotion label space expected by the benchmark. DeepSeek-16B under unstructured sparsity is the most visible case in our results: parseable outputs drop from an FP16 baseline of 169.7 to 108.2, and the benchmark score becomes unavailable. This pattern suggests that compression weakens emotional judgment: the issue is not only whether the model understands emotion, but also whether it can express the intended emotion precisely.

\subsection{Persona Fidelity and Stylistic Consistency}

\begin{figure*}[t]
    \centering
    \includegraphics[width=0.995\textwidth]{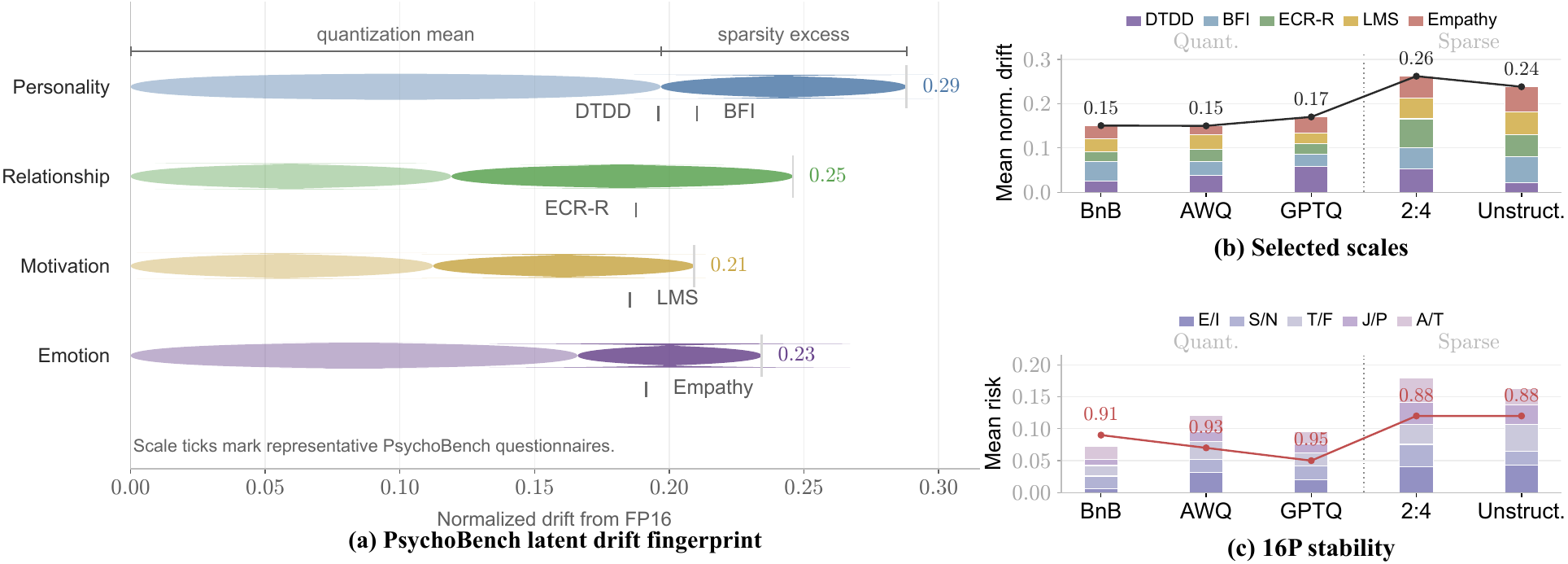}
    \caption{Persona fidelity and stylistic consistency. (a) PsychoBench latent drift fingerprint from each FP16 parent; light segments summarize the quantization mean and darker extensions indicate additional sparse-setting drift. (b) Representative psychometric scales grouped by DTDD, BFI, ECR-R, LMS, and Empathy, with the line marking mean normalized drift by method. (c) 16P stability: stacked bars indicate axis-level drift/risk and the red line labels dominant-type stability. The panels connect visible persona flattening to latent psychometric and identity-profile movement.}
    \label{fig:persona}
    \vspace{-0.5cm}
\end{figure*}

The persona dimension first examines whether compressed models still maintain visible role enactment, character consistency, and stylistic continuity. It then turns to a different question: whether questionnaire-elicited psychological profiles and identity-style readouts shift after compression.

\textbf{CharacterEval} captures the visible role-play side of persona fidelity. Using CharacterEval's native metrics, we construct within-dimension diagnostic contrasts to identify which visible persona signals are more compression-sensitive. Table~\ref{tab:charactereval} compares these sensitive signals with more stable reference signals in CharacterEval. It shows that more compression-sensitive signals are concentrated in expressive vividness, role enactment, and multi-turn continuity. Character-knowledge anchors and readability-oriented controls remain comparatively more stable, especially outside the sparse settings. This suggests that compression does not first erase the role; it flattens how the role is enacted. Under sparse settings, the loss extends further to persona behavior, persona utterance, and conversational consistency.

% Compact CharacterEval visible-persona diagnostics table.
% Requires: booktabs, array, makecell, xcolor, adjustbox.

\begin{table}[t]
\centering
\footnotesize
\setlength{\tabcolsep}{4pt}
\renewcommand{\arraystretch}{1.08}

\definecolor{PersonaAccent}{HTML}{7B5EA7}
\definecolor{PersonaBg}{HTML}{FAF6FD}
\definecolor{PersonaTag}{HTML}{FBF7FC}
\definecolor{SubText}{HTML}{555555}

\newcommand{\cehead}[1]{\textcolor{PersonaAccent}{\textbf{#1}}}
\newcommand{\sig}[1]{\textcolor{PersonaAccent}{\textbf{#1}}}
\newcommand{\subsig}[1]{\textcolor{SubText}{\footnotesize #1}}
\newcommand{\reading}[1]{%
  \begingroup
  \setlength{\fboxsep}{3pt}%
  \colorbox{PersonaTag}{\textcolor{PersonaAccent}{\footnotesize\textbf{#1}}}%
  \endgroup
}
\newcommand{\retcell}[2]{\makecell[c]{All: #1\\Sparse: #2}}

\caption{\textbf{CharacterEval visible-persona diagnostics.}
Values are parent-relative retention ratios \(R_{m,b}\). ``All'' averages all available compression variants; ``Sparse'' averages 2:4 structured and unstructured sparsity. Slash-separated metric values follow the metric order shown in each signal cell.}
\label{tab:charactereval}

\begin{adjustbox}{max width=\linewidth}
\begin{tabular}{@{} l l c l c c @{}}
\toprule
\rowcolor{PersonaBg}
\cehead{Dimension} &
\cehead{Sensitive visible signal} &
\cehead{Retention \(R\)} &
\cehead{Reference signal} &
\cehead{Retention \(R\)} &
\cehead{Diagnostic reading} \\
\midrule

Role-playing attractiveness &
\makecell[l]{\sig{ED / CS}\\\subsig{expression diversity / communication}} &
\retcell{0.87 / 0.88}{0.74 / 0.77} &
\makecell[l]{\sig{HL / Emp.}\\\subsig{human-likeness / empathy}} &
\retcell{0.91 / 0.91}{0.80 / 0.83} &
\reading{vividness drop} \\
\addlinespace[0.38em]

Character consistency &
\makecell[l]{\sig{PB / PU}\\\subsig{persona behavior / utterance}} &
\retcell{0.89 / 0.92}{0.75 / 0.81} &
\makecell[l]{\sig{KA / KE / KH}\\\subsig{character-knowledge anchors}} &
\retcell{0.94 / 0.92 / 0.91}{0.88 / 0.83 / 0.81} &
\reading{enactment loss} \\
\addlinespace[0.38em]

Conversational ability &
\makecell[l]{\sig{Cons.}\\\subsig{multi-turn consistency}} &
\retcell{0.89}{0.77} &
\makecell[l]{\sig{Flu. / Coh.}\\\subsig{fluency / coherence}} &
\retcell{0.91 / 0.92}{0.81 / 0.81} &
\reading{continuity loss} \\

\bottomrule
\end{tabular}
\end{adjustbox}
\vspace{-0.5cm}
\end{table}%

\textbf{PsychoBench} groups its questionnaires into four psychological categories: personality traits, interpersonal relationships, motivation, and emotional abilities. We use these categories to summarize how compression changes the model's psychological portrayal. Figure~\ref{fig:persona}(a) reports consistent drift from the FP16 profile. Across four PsychoBench categories, sparse settings produce visibly larger drift than quantization, indicating stronger movement in the psychological portrayal. Figure~\ref{fig:persona}(b) makes this movement more interpretable by showing representative questionnaires: DTDD for dark-triad-related traits, BFI for broad personality, ECR-R for attachment, LMS for motivation, and Empathy for emotional-social orientation. The DTDD result is especially informative: in the raw questionnaire scores, compression tends to shift the model toward higher dark-triad-related traits. Figure~\ref{fig:persona}(c) confirms this personality-type drift: 16P dominant-type stability drops after compression, especially under sparse settings.

\subsection{Findings}

% Limit wrapping to the Findings text; do not narrow the next section.
\begin{wrapfigure}[17]{r}{0.38\textwidth}
    \centering
    \vspace{-0.8em}
    \includegraphics[width=\linewidth]{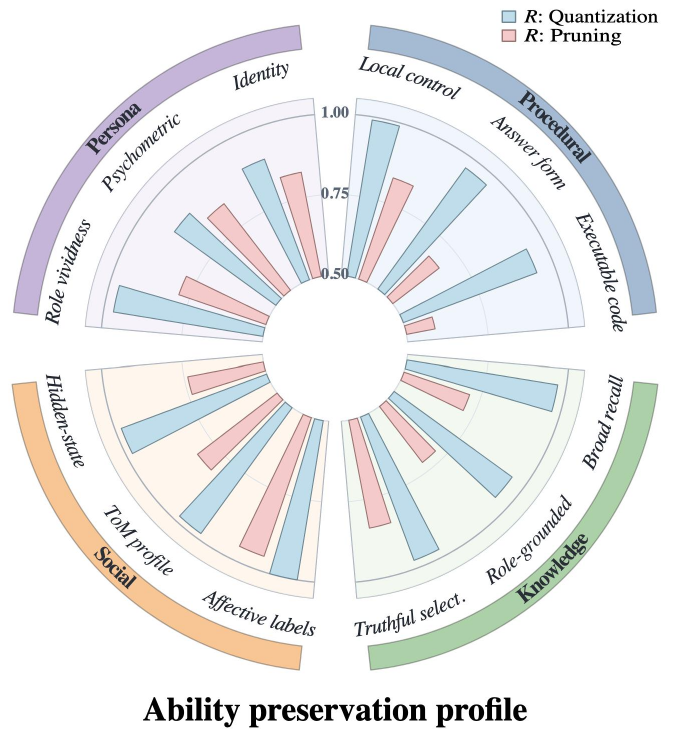}
    \captionsetup{font=scriptsize}
    \caption{Overall findings of ACE-Compression. Bar length indicates the mean FP16-relative retention \(R\) for each ability layer under quantization or pruning.}
    \label{fig:overview}
    \vspace{-1em}
\end{wrapfigure}

Figure~\ref{fig:overview} summarizes three key findings.

First, compression redistributes abilities rather than uniformly lowering scores. Knowledge-oriented signals are relatively stable, whereas procedural, social, and persona dimensions expose distinct weaknesses. Within these dimensions, local instruction compliance and broad knowledge survive better than full-response coordination, role-specific knowledge, implicit mental states, and persona stability. Second, coarse, short-range, and surface-facing signals are more robust than fine-grained, long-tail, and context-dependent abilities. This produces shallow retention: aggregate scores can remain acceptable after supporting abilities have weakened. Third, quantization generally retains abilities better than pruning, although the gap varies across dimensions and layers. Consequently, a checkpoint with high mean retention may still be unsuitable when an application depends on a specific ability.

\section{Broader Impact}
\label{sec:impact}

Our results support ability-aware deployment: practitioners should choose compression methods according to application-critical abilities rather than average retention alone. Quantization, especially GPTQ-INT4, preserves procedural reasoning and knowledge grounding more reliably, whereas sparsity—particularly 2:4 sparsity—more strongly damages fine-grained constraint execution, multi-step reasoning, and social cognition. Thus, conversational systems may prioritize persona fidelity, while code-generation services may prioritize HumanEval retention. Reporting ability profiles alongside aggregate scores reduces the risk of selecting a model that fails on its critical capability.

The diagnostics also provide targets for compression research. Regularization or calibration objectives could explicitly preserve vulnerable dimensions, for example the activations associated with ToMBench's belief-action-link subability. Occasional improvements in strict instruction scores under quantization further suggest that rounding noise may regularize some tasks, motivating carefully validated hybrid strategies. Finally, per-ability auditing is useful for edge, embodied, and regulated applications, where truthfulness or empathy may need to remain above deployment thresholds. The benchmark mapping and retention metrics provide a practical checklist for rejecting variants that pass average-score tests but fail a required ability.

\section{Conclusion and Future Work}
\label{sec:conclusion}

ACE-Compression provides an ability-centric diagnosis of how compression redistributes LLM capabilities. It characterizes behavioral changes rather than their internal causal mechanisms, and its model panel does not fully disentangle family, scale, and compression-method effects. The present scope covers post-training quantization and sparsity; future work should expand the model panel and examine distillation, quantization-aware training, and other retraining-based methods.

\clearpage
\bibliographystyle{plainnat}
\bibliography{references}

\appendix
% Shared appendix table style for consistent font size and spacing.
% Called inside each appendix table environment after \centering.
\newcommand{\apptablestyle}{%
  \footnotesize
  \setlength{\tabcolsep}{4pt}%
  \renewcommand{\arraystretch}{1.02}%
}

\section{Benchmark and Metric Details}
\label{sec:app_metric_details}

This appendix follows the ACE-Compression design in the main paper and provides a detailed audit trail for its ability-centric claims. The main text interprets compression effects through four dimensions: procedural reasoning and constraint execution, knowledge grounding and truthfulness, social cognition and affective understanding, and persona fidelity and stylistic consistency. The appendix reports the detailed evaluation results behind these interpretations, preserving benchmark-native views and providing fuller tables where the main paper necessarily shows selected summaries.

The evaluation suite contains ten diagnostic benchmarks. Its evaluation portion includes 541 instruction-following prompts, 1,319 grade-school math questions, 164 programming problems, 817 truthfulness questions, 13,948 multiple-choice exam questions, 6,000 bilingual role-grounded knowledge questions, 2,860 theory-of-mind testing samples, and 60 emotion-understanding questions. These item-level benchmarks are complemented by 1,785 multi-turn role-playing dialogues and 13 psychometric scales, which are reported in their native units.

\paragraph{Procedural benchmarks.}
IFEval is kept in prompt- and instruction-level views, each with strict and loose scoring. This split is preserved because the paper's procedural analysis relies on the difference between local instruction recognition and whole-response coordination. GSM8K is likewise kept in strict and flexible matching, since the strict--flexible gap is diagnostic of brittle answer realization. HumanEval is separated by decoding regime rather than summarized by a single pass@$k$: the distinction between greedy execution, low-budget sampling, and broader recoverability under larger sample budgets is part of the substantive result.

\paragraph{Knowledge benchmarks.}
TruthfulQA requires benchmark-native reporting because its selection-side and generation-side metrics support different diagnostic readings. The paper reasons specifically with MC1/MC2 and generation-side BLEURT-max, so the appendix surfaces those exact columns and states explicitly that this is a selected analytic table. C-Eval is compact enough to report directly with both validation and test accuracy. For RoleEval, the appendix reports grouped means that preserve the shot/language structure.

\paragraph{Social benchmarks.}
ToMBench is supplied at three granularities: six official abilities, eight tasks, and 31 subabilities, each with EN/ZH variants. For the six-ability view, it reports the language-collapsed scores that correspond to the main paper's base-model profile. For the task and subability views, the appendix provides grouped retention summaries with the averaging procedure stated in each caption. EQ-Bench is reported with both benchmark score and parseability.

\paragraph{Persona benchmarks.}
CharacterEval is reported with the full grouped appendix table, exposing the detailed visible-persona scores. For PsychoBench, the appendix keeps the representative scales discussed in the paper as raw released outputs. The 16P file is available in aggregated form, with personality-type counts and axis percentages, so the appendix reports that aggregation directly and adds a clearly labeled axis-drift readout computed from it.

\section{Full Results by Ability Dimension}
\label{sec:app_full_results}

\subsection{Procedural Abilities}
\label{sec:app_procedural}

This subsection collects the fuller benchmark-native tables behind the procedural dimension. 

\paragraph{IFEval.}
Table~\ref{tab:appendix_ifeval} reports the full prompt-level and instruction-level IFEval outputs. The table keeps both strict and loose scoring because the two granularities support different diagnostic readings: prompt-level scores require all verifiable instructions in a prompt to be satisfied jointly, whereas instruction-level scores measure more local compliance with individual instructions. This makes IFEval the appendix-side evidence for the main paper's local-to-global procedural analysis.

\textit{Quantization.}
Across most model groups, 4-bit quantization stays close to the FP16 parent on both prompt-level and instruction-level views. Qwen2.5-7B under BnB-4bit slightly exceeds FP16 on prompt-level strict (0.7209 vs. 0.7172) and instruction-level loose (0.8177 vs. 0.8106). Qwen2.5-14B is similarly stable: GPTQ-INT4 slightly improves instruction-level strict (0.8501 vs. 0.8441) and instruction-level loose (0.8657 vs. 0.8621). The larger Qwen2.5-32B and Llama-70B groups also show strong quantization robustness. Qwen2.5-32B matches FP16 under BnB-4bit on prompt-level strict (0.8004) and improves slightly under AWQ (0.8059), while Llama-70B improves under AWQ and GPTQ-INT4 on prompt-level strict (0.8447 vs. 0.8336). These cases support the main text's observation that quantization often preserves local instruction compliance and full-prompt control better than pruning.

\textit{Model-family variation.}
The stability of quantization is not uniform across model families. DeepSeek-V2-Lite-16B has a lower FP16 IFEval baseline than the Qwen and Llama groups, and its compressed variants show clearer losses. For example, AWQ reduces DeepSeek-V2-Lite-16B prompt-level strict from 0.4307 to 0.3549 and instruction-level strict from 0.5372 to 0.4712. This contrast is useful for ACE-Compression because it shows that the same compression method can be nearly lossless for one base model while producing visible compliance loss for another.

\textit{Sparsity.}
Sparse variants show much larger degradation than quantized variants, especially under 2:4 structured sparsity. Llama3.1-8B prompt-level strict drops from 0.7468 to 0.3937 under 2:4 sparsity, and Qwen2.5-7B drops from 0.7172 to 0.4547. The same pattern appears in the instruction-level view: Llama3.1-8B instruction-level strict falls from 0.8237 to 0.5252, and Qwen2.5-7B falls from 0.7926 to 0.5863. Unstructured sparsity is generally less damaging than 2:4 sparsity. For example, Qwen2.5-7B prompt-level strict is 0.6359 under unstructured sparsity versus 0.4547 under 2:4 sparsity, and Llama-70B is 0.8078 versus 0.7135. However, unstructured sparsity still remains below the FP16 parent in most rows, indicating that sparsity weakens instruction-following behavior more strongly than post-training quantization.

\textit{Granularity and strictness.}
Instruction-level scores are consistently higher than prompt-level scores because they evaluate individual verifiable instructions rather than joint satisfaction of the entire prompt. Loose scores are also higher than strict scores under the same granularity. These gaps are not merely reporting artifacts: they are the benchmark-native evidence for the procedural distinction used in the main text. A compressed model may still satisfy many individual requirements while failing to coordinate all requirements in a complete response. The IFEval table therefore supports the ACE-Compression reading that compression first exposes weaknesses in global constraint coordination before fully erasing local instruction recognition.

\begin{table*}[t]
\centering
\apptablestyle
\definecolor{ACEHeaderBg}{HTML}{F1F3F5}
\definecolor{ACEHeaderText}{HTML}{1F2D3D}
\definecolor{ACEProcHi}{HTML}{EAF3FB}
\definecolor{ACEProcBorder}{HTML}{2F75B5}
\definecolor{ACERuleDark}{HTML}{2E3640}
\definecolor{ACERuleLight}{HTML}{CAD3DD}
\newcommand{\trendcell}[1]{\cellcolor{ACEProcHi}#1}
\newcommand{\keycell}[1]{\begingroup\setlength{\fboxsep}{1.2pt}\fcolorbox{ACEProcBorder}{ACEProcHi}{#1}\endgroup}
\newcommand{\anchorval}[1]{\underline{#1}}
\newcommand{\ifHeader}[1]{\textcolor{ACEHeaderText}{\textbf{#1}}}
\newcommand{\ifTopRule}{\arrayrulecolor{ACERuleDark}\toprule[0.55pt]\arrayrulecolor{ACERuleDark}}
\newcommand{\ifHeadRule}{\arrayrulecolor{ACERuleDark}\midrule[0.42pt]\arrayrulecolor{black}}
\newcommand{\ifMidRule}{\arrayrulecolor{ACERuleDark}\midrule[0.24pt]\arrayrulecolor{black}}
\newcommand{\ifBottomRule}{\arrayrulecolor{ACERuleDark}\bottomrule[0.55pt]\arrayrulecolor{black}}
\newcommand{\hdrstrut}{\rule{0pt}{2ex}}
\newcommand{\grouphead}[4]{%
  \begin{tabular}[c]{@{}c@{}}
    \rule{0pt}{2.15ex}\textcolor{ACEHeaderText}{\textbf{#1}}\\[-0.55ex]
    \rule{#2}{0.4pt}\\[-0.15ex]
    \makebox[#2][c]{%
      \makebox[\dimexpr#2/2\relax][c]{\ifHeader{#3}\hdrstrut}%
      \makebox[\dimexpr#2/2\relax][c]{\ifHeader{#4}\hdrstrut}%
    }
  \end{tabular}%
}
\caption{Full IFEval results for the procedural reasoning and constraint execution dimension. Prompt-level and instruction-level scores are reported under strict and loose evaluation, separating whole-response coordination from local instruction recognition.}
\label{tab:appendix_ifeval}
\begin{tabular}{llcccc}
\ifTopRule
\rowcolor{ACEHeaderBg}
\begin{tabular}[c]{@{}l@{}}\ifHeader{Model Group}\end{tabular} & \begin{tabular}[c]{@{}l@{}}\ifHeader{Variant}\end{tabular} & \multicolumn{2}{c}{\grouphead{Prompt-level}{14em}{Strict}{Loose}} & \multicolumn{2}{c}{\grouphead{Instruction-level}{7em}{Strict}{Loose}} \\
\ifHeadRule
\multirow{6}{*}{\textbf{Qwen2.5-7B}} & FP16 & \anchorval{0.7172 $\pm$ 0.0194} & 0.7394 $\pm$ 0.0189 & \anchorval{0.7926} & \anchorval{0.8106} \\
 & BnB-4bit & \keycell{0.7209 $\pm$ 0.0193} & 0.7449 $\pm$ 0.0188 & 0.7962 & \keycell{0.8177} \\
 & AWQ & 0.7098 $\pm$ 0.0195 & 0.7246 $\pm$ 0.0192 & 0.7866 & 0.8046 \\
 & GPTQ-INT4 & 0.7079 $\pm$ 0.0196 & 0.7264 $\pm$ 0.0192 & 0.7950 & 0.8118 \\
 & 2:4-Sparse & \keycell{0.4547 $\pm$ 0.0214} & 0.4824 $\pm$ 0.0215 & \keycell{0.5863} & 0.6139 \\
 & Unstructured-Sparse & \keycell{0.6359 $\pm$ 0.0207} & 0.6765 $\pm$ 0.0201 & 0.7302 & 0.7650 \\
\ifMidRule
\multirow{6}{*}{\textbf{Llama3.1-8B}} & FP16 & \anchorval{0.7468 $\pm$ 0.0187} & 0.7948 $\pm$ 0.0174 & \anchorval{0.8237} & 0.8585 \\
 & BnB-4bit & 0.7172 $\pm$ 0.0194 & 0.7671 $\pm$ 0.0182 & 0.8010 & 0.8417 \\
 & AWQ & 0.7412 $\pm$ 0.0188 & 0.7745 $\pm$ 0.0180 & 0.8189 & 0.8465 \\
 & GPTQ-INT4 & 0.7301 $\pm$ 0.0191 & 0.7597 $\pm$ 0.0184 & 0.8082 & 0.8321 \\
 & 2:4-Sparse & \keycell{0.3937 $\pm$ 0.0210} & 0.4381 $\pm$ 0.0214 & \keycell{0.5252} & 0.5695 \\
 & Unstructured-Sparse & 0.6747 $\pm$ 0.0202 & 0.7116 $\pm$ 0.0195 & 0.7542 & 0.7902 \\
\ifMidRule
\multirow{7}{*}{\textbf{Qwen2.5-14B}} & FP16 & 0.7874 $\pm$ 0.0176 & 0.8115 $\pm$ 0.0168 & \anchorval{0.8441} & \anchorval{0.8621} \\
 & BnB-4bit & 0.7819 $\pm$ 0.0178 & 0.8041 $\pm$ 0.0171 & 0.8441 & 0.8597 \\
 & AWQ & 0.7763 $\pm$ 0.0179 & 0.8004 $\pm$ 0.0172 & 0.8405 & 0.8597 \\
 & GPTQ-INT4 & 0.7893 $\pm$ 0.0175 & 0.8115 $\pm$ 0.0168 & \keycell{0.8501} & \keycell{0.8657} \\
 & 2:4-Sparse & 0.5860 $\pm$ 0.0212 & 0.6137 $\pm$ 0.0210 & 0.6823 & 0.7122 \\
 & Unstructured-Sparse & 0.7431 $\pm$ 0.0188 & 0.7634 $\pm$ 0.0183 & 0.8106 & 0.8273 \\
 & Trim & 0.7893 $\pm$ 0.0175 & 0.8059 $\pm$ 0.0170 & 0.8477 & 0.8633 \\
\ifMidRule
\multirow{5}{*}{\textbf{DeepSeek-V2-Lite-16B}} & FP16 & \anchorval{0.4307 $\pm$ 0.0213} & 0.4510 $\pm$ 0.0214 & \anchorval{0.5372} & 0.5671 \\
 & BnB-4bit & 0.4214 $\pm$ 0.0212 & 0.4566 $\pm$ 0.0214 & 0.5312 & 0.5683 \\
 & AWQ & \keycell{0.3549 $\pm$ 0.0206} & 0.3789 $\pm$ 0.0209 & \keycell{0.4712} & 0.5036 \\
 & 2:4-Sparse & 0.2107 $\pm$ 0.0175 & 0.2292 $\pm$ 0.0181 & 0.3189 & 0.3453 \\
 & Unstructured-Sparse & 0.2643 $\pm$ 0.0190 & 0.2921 $\pm$ 0.0196 & 0.3693 & 0.3981 \\
\ifMidRule
\multirow{6}{*}{\textbf{Qwen2.5-32B}} & FP16 & \anchorval{0.8004 $\pm$ 0.0172} & 0.8244 $\pm$ 0.0164 & 0.8585 & 0.8777 \\
 & BnB-4bit & \trendcell{0.8004 $\pm$ 0.0172} & 0.8189 $\pm$ 0.0166 & 0.8633 & 0.8777 \\
 & AWQ & \keycell{0.8059 $\pm$ 0.0170} & 0.8262 $\pm$ 0.0163 & 0.8681 & 0.8849 \\
 & GPTQ-INT4 & 0.7745 $\pm$ 0.0180 & 0.8096 $\pm$ 0.0169 & 0.8453 & 0.8705 \\
 & 2:4-Sparse & 0.6821 $\pm$ 0.0200 & 0.7135 $\pm$ 0.0195 & 0.7638 & 0.7914 \\
 & Unstructured-Sparse & 0.7597 $\pm$ 0.0184 & 0.7782 $\pm$ 0.0179 & 0.8237 & 0.8429 \\
\ifMidRule
\multirow{7}{*}{\textbf{Llama-70B}} & FP16 & \anchorval{0.8336 $\pm$ 0.0160} & 0.8614 $\pm$ 0.0149 & 0.8849 & 0.9053 \\
 & BnB-4bit & 0.8152 $\pm$ 0.0167 & 0.8355 $\pm$ 0.0160 & 0.8705 & 0.8873 \\
 & AWQ & \keycell{0.8447 $\pm$ 0.0156} & 0.8743 $\pm$ 0.0143 & 0.8921 & 0.9161 \\
 & GPTQ-INT4 & \keycell{0.8447 $\pm$ 0.0156} & 0.8725 $\pm$ 0.0144 & 0.8945 & 0.9149 \\
 & GPTQ-INT8 & 0.8392 $\pm$ 0.0158 & 0.8725 $\pm$ 0.0144 & 0.8909 & 0.9137 \\
 & 2:4-Sparse & \anchorval{0.7135 $\pm$ 0.0195} & 0.7486 $\pm$ 0.0187 & 0.7830 & 0.8177 \\
 & Unstructured-Sparse & \keycell{0.8078 $\pm$ 0.0170} & 0.8503 $\pm$ 0.0154 & 0.8645 & 0.8957 \\
\ifBottomRule
\end{tabular}
\end{table*}

\paragraph{GSM8K.}
Table~\ref{tab:appendix_gsm8k} keeps the strict-match and flexible-match views together. This is necessary because the two scores diagnose different parts of multi-step answer realization: flexible matching asks whether the final numeric answer can be recovered somewhere in the output, whereas strict matching is more sensitive to whether the model presents the answer in the expected canonical form. The appendix table therefore provides the detailed evidence behind the main paper's flexible-match retention and flexible--strict gap analysis.

\textit{Quantization.}
Across most model groups, flexible-match accuracy remains close to the FP16 parent under 4-bit quantization. Qwen2.5-7B retains most of its flexible-match performance under BnB-4bit, AWQ, and GPTQ-INT4 (0.7377, 0.7202, and 0.7225 versus 0.7597 for FP16). Llama3.1-8B shows the same pattern, with flexible-match scores of 0.8211, 0.8143, and 0.8006 compared with the FP16 value of 0.8446. The larger models are even more stable in this view: Qwen2.5-32B reaches 0.8370 under AWQ and 0.8317 under GPTQ-INT4, both slightly above its FP16 flexible-match score of 0.8211, while Llama-70B nearly matches FP16 under AWQ and GPTQ-INT4 (0.9462 and 0.9439 versus 0.9469). These results support the main text's observation that quantization usually preserves recoverable answer correctness on GSM8K.

\textit{Strict-match sensitivity.}
Strict-match scores are more irregular than flexible-match scores. For some Qwen models, quantization preserves the recoverable answer while weakening the canonical answer form. For example, Qwen2.5-7B under GPTQ-INT4 keeps flexible match high at 0.7225 but reduces strict match from 0.3351 to 0.1691. Qwen2.5-14B shows a similar separation: GPTQ-INT4 slightly improves flexible match relative to FP16 (0.7817 vs. 0.7779) while reducing strict match from 0.2767 to 0.1471. In contrast, Llama-70B is stable in both views, with GPTQ-INT4 matching the FP16 strict score (0.9014) and nearly matching flexible match (0.9439 vs. 0.9469). Thus, GSM8K shows that answer recoverability and answer formatting do not always move together after compression.

\textit{Sparsity.}
Sparse variants show larger degradation in flexible match, especially under 2:4 structured sparsity. Llama3.1-8B drops from 0.8446 to 0.1168 under 2:4 sparsity, and DeepSeek-V2-Lite-16B drops from 0.5929 to 0.1054. The effect is also clear for larger models: Llama-70B falls from 0.9469 to 0.6164, and Qwen2.5-32B falls from 0.8211 to 0.6247. Unstructured sparsity is generally less damaging than 2:4 sparsity, although it still usually remains below the FP16 parent. For example, Qwen2.5-7B flexible match is 0.7036 under unstructured sparsity versus 0.5019 under 2:4 sparsity, Llama3.1-8B is 0.5444 versus 0.1168, and Llama-70B is 0.9022 versus 0.6164. This supports the procedural finding that pruning affects multi-step answer realization more strongly than quantization.

\textit{Flexible--strict gap.}
The strict--flexible gap is itself diagnostic. Some base models, especially the Qwen groups, have large FP16 gaps: Qwen2.5-7B has a gap of 0.4246, Qwen2.5-14B has a gap of 0.5012, and Qwen2.5-32B has a gap of 0.5156. Compression can change this gap in different ways. Under quantization, flexible match often stays high while strict match can move more sharply, indicating that the answer may remain recoverable even when its canonical form changes. Under sparsity, the gap can collapse because flexible match drops substantially, as in Qwen2.5-14B under 2:4 sparsity, where the gap narrows from 0.5012 to 0.1494. These patterns justify keeping strict and flexible matching separate: a single GSM8K score would hide whether compression affects mathematical answer recovery, answer formatting, or both.

\begin{table*}[t]
\centering
\apptablestyle
\definecolor{ACEHeaderBg}{HTML}{F1F3F5}
\definecolor{ACEHeaderText}{HTML}{1F2D3D}
\definecolor{ACEProcHi}{HTML}{EAF3FB}
\definecolor{ACEProcBorder}{HTML}{2F75B5}
\definecolor{ACERuleDark}{HTML}{2E3640}
\newcommand{\trendcell}[1]{\cellcolor{ACEProcHi}#1}
\newcommand{\keycell}[1]{\begingroup\setlength{\fboxsep}{1.2pt}\fcolorbox{ACEProcBorder}{ACEProcHi}{#1}\endgroup}
\newcommand{\anchorval}[1]{\underline{#1}}
\newcommand{\gsmHeader}[1]{\textcolor{ACEHeaderText}{\textbf{#1}}}
\newcommand{\gsmTopRule}{\arrayrulecolor{ACERuleDark}\toprule[0.55pt]\arrayrulecolor{ACERuleDark}}
\newcommand{\gsmHeadRule}{\arrayrulecolor{ACERuleDark}\midrule[0.42pt]\arrayrulecolor{black}}
\newcommand{\gsmMidRule}{\arrayrulecolor{ACERuleDark}\midrule[0.24pt]\arrayrulecolor{black}}
\newcommand{\gsmBottomRule}{\arrayrulecolor{ACERuleDark}\bottomrule[0.55pt]\arrayrulecolor{black}}
\caption{Full GSM8K results for the procedural reasoning and constraint execution dimension. Strict and flexible matching are reported together to distinguish canonical answer realization from recoverable numeric correctness.}
\label{tab:appendix_gsm8k}
\begin{tabular}{llcc}
\gsmTopRule
\rowcolor{ACEHeaderBg}
\gsmHeader{Model Group} & \gsmHeader{Variant} & \gsmHeader{Strict Match} & \gsmHeader{Flexible Match} \\
\gsmHeadRule
\multirow{6}{*}{\textbf{Qwen2.5-7B}} & FP16 & \anchorval{0.3351 $\pm$ 0.0130} & \anchorval{0.7597 $\pm$ 0.0118} \\
 & BnB-4bit & 0.2434 $\pm$ 0.0118 & \trendcell{0.7377 $\pm$ 0.0121} \\
 & AWQ & 0.1971 $\pm$ 0.0110 & \trendcell{0.7202 $\pm$ 0.0124} \\
 & GPTQ-INT4 & \keycell{0.1691 $\pm$ 0.0103} & \trendcell{0.7225 $\pm$ 0.0123} \\
 & 2:4-Sparse & 0.3995 $\pm$ 0.0135 & \keycell{0.5019 $\pm$ 0.0138} \\
 & Unstructured-Sparse & 0.4367 $\pm$ 0.0137 & \keycell{0.7036 $\pm$ 0.0126} \\
\gsmMidRule
\multirow{6}{*}{\textbf{Llama3.1-8B}} & FP16 & 0.7991 $\pm$ 0.0110 & \anchorval{0.8446 $\pm$ 0.0100} \\
 & BnB-4bit & 0.7779 $\pm$ 0.0114 & \trendcell{0.8211 $\pm$ 0.0106} \\
 & AWQ & 0.7665 $\pm$ 0.0117 & \trendcell{0.8143 $\pm$ 0.0107} \\
 & GPTQ-INT4 & 0.7096 $\pm$ 0.0125 & \trendcell{0.8006 $\pm$ 0.0110} \\
 & 2:4-Sparse & 0.1077 $\pm$ 0.0085 & \keycell{0.1168 $\pm$ 0.0088} \\
 & Unstructured-Sparse & 0.5102 $\pm$ 0.0138 & \keycell{0.5444 $\pm$ 0.0137} \\
\gsmMidRule
\multirow{7}{*}{\textbf{Qwen2.5-14B}} & FP16 & \anchorval{0.2767 $\pm$ 0.0123} & \anchorval{0.7779 $\pm$ 0.0114} \\
 & BnB-4bit & 0.3419 $\pm$ 0.0131 & 0.7672 $\pm$ 0.0116 \\
 & AWQ & 0.1948 $\pm$ 0.0109 & 0.7528 $\pm$ 0.0119 \\
 & GPTQ-INT4 & \keycell{0.1471 $\pm$ 0.0098} & \keycell{0.7817 $\pm$ 0.0114} \\
 & 2:4-Sparse & 0.3950 $\pm$ 0.0135 & 0.5444 $\pm$ 0.0137 \\
 & Unstructured-Sparse & 0.4769 $\pm$ 0.0138 & 0.7976 $\pm$ 0.0111 \\
 & Trim & 0.4352 $\pm$ 0.0137 & 0.7619 $\pm$ 0.0117 \\
\gsmMidRule
\multirow{5}{*}{\textbf{DeepSeek-V2-Lite-16B}} & FP16 & 0.1190 $\pm$ 0.0089 & \anchorval{0.5929 $\pm$ 0.0135} \\
 & BnB-4bit & 0.1168 $\pm$ 0.0088 & 0.5307 $\pm$ 0.0137 \\
 & AWQ & 0.0220 $\pm$ 0.0040 & 0.4534 $\pm$ 0.0137 \\
 & 2:4-Sparse & 0.0432 $\pm$ 0.0056 & \keycell{0.1054 $\pm$ 0.0085} \\
 & Unstructured-Sparse & 0.2661 $\pm$ 0.0122 & 0.3397 $\pm$ 0.0130 \\
\gsmMidRule
\multirow{6}{*}{\textbf{Qwen2.5-32B}} & FP16 & 0.3055 $\pm$ 0.0127 & \anchorval{0.8211 $\pm$ 0.0106} \\
 & BnB-4bit & 0.2009 $\pm$ 0.0110 & 0.8097 $\pm$ 0.0108 \\
 & AWQ & 0.3366 $\pm$ 0.0130 & \keycell{0.8370 $\pm$ 0.0102} \\
 & GPTQ-INT4 & 0.2638 $\pm$ 0.0121 & \keycell{0.8317 $\pm$ 0.0103} \\
 & 2:4-Sparse & 0.2047 $\pm$ 0.0111 & \keycell{0.6247 $\pm$ 0.0133} \\
 & Unstructured-Sparse & 0.3268 $\pm$ 0.0129 & 0.7915 $\pm$ 0.0112 \\
\gsmMidRule
\multirow{6}{*}{\textbf{Llama-70B}} & FP16 & \anchorval{0.9014 $\pm$ 0.0082} & \anchorval{0.9469 $\pm$ 0.0062} \\
 & BnB-4bit & 0.8597 $\pm$ 0.0096 & 0.9212 $\pm$ 0.0074 \\
 & AWQ & 0.8916 $\pm$ 0.0086 & \keycell{0.9462 $\pm$ 0.0062} \\
 & GPTQ-INT4 & \trendcell{0.9014 $\pm$ 0.0082} & \keycell{0.9439 $\pm$ 0.0063} \\
 & 2:4-Sparse & 0.5489 $\pm$ 0.0137 & \keycell{0.6164 $\pm$ 0.0134} \\
 & Unstructured-Sparse & 0.8575 $\pm$ 0.0096 & \keycell{0.9022 $\pm$ 0.0082} \\
\gsmBottomRule
\end{tabular}
\end{table*}

\paragraph{HumanEval.}
Tables~\ref{tab:appendix_humaneval_greedy_n20} and \ref{tab:appendix_humaneval_n100} keep the decoding regimes separate so that first-pass reliability and sampling-based recoverability can be audited directly. Greedy pass@$1$ measures whether the first decoded program is already executable and correct, while the sampling columns show whether correct programs remain accessible under repeated decoding.

\textit{Greedy execution.}
Quantized variants generally stay close to their FP16 parents under greedy pass@$1$, although the amount of preservation varies by model family. Qwen2.5-7B remains near its FP16 score of 0.7927 under BnB-4bit, AWQ, and GPTQ-INT4 (0.7683, 0.7744, and 0.7683). Qwen2.5-32B is even more stable, with AWQ and GPTQ-INT4 slightly exceeding the FP16 greedy score (0.8902 and 0.8963 vs. 0.8780). Llama-70B is similarly robust: BnB-4bit exactly matches FP16 at 0.8171, while GPTQ-INT4 and GPTQ-INT8 reach 0.8293. The main exception is DeepSeek-V2-Lite-16B, where AWQ reduces greedy pass@$1$ from 0.5061 to 0.2256. These results support the main text's finding that quantization often preserves first-sample executable consistency, but not uniformly across model families.

\textit{Sparse degradation.}
Sparse variants show much larger drops in first-sample execution, especially under 2:4 structured sparsity. Qwen2.5-7B drops from 0.7927 to 0.3354 under 2:4 sparsity, Llama3.1-8B drops from 0.6951 to 0.1280, and DeepSeek-V2-Lite-16B drops from 0.5061 to 0.0549. Larger models remain more robust in absolute terms but still degrade substantially: Qwen2.5-32B falls from 0.8780 to 0.6341, and Llama-70B falls from 0.8171 to 0.5244. Unstructured sparsity is usually less damaging than 2:4 sparsity, as seen for Qwen2.5-7B (0.6890 vs. 0.3354), Llama3.1-8B (0.4634 vs. 0.1280), and Llama-70B (0.7622 vs. 0.5244), but it still remains below the FP16 parent in most cases.

\textit{Low-budget sampling.}
The $n=20$ sampling columns show that compression can affect first-sample execution and short-budget recoverability differently. Some quantized models preserve or even slightly improve pass@$10$ under sampling: Llama3.1-8B under AWQ reaches 0.7816 compared with the FP16 value of 0.7586, and Qwen2.5-7B under BnB-4bit exactly matches FP16 on both sampling pass@$1$ and pass@$10$ (0.7034 and 0.8172). In contrast, sparse variants lose much of their low-budget recoverability. For example, Llama3.1-8B under 2:4 sparsity has sampling pass@$10$ of only 0.1518, compared with 0.7586 for FP16, and DeepSeek-V2-Lite-16B drops from 0.6535 to 0.1021. This shows that pruning not only weakens the first decoded solution but also reduces the chance that a correct program appears within a small sample budget.

\textit{Higher-budget recoverability.}
The $n=100$ table shows that some valid programs remain accessible even after compression. Quantized variants often preserve high-budget recoverability almost completely. Qwen2.5-7B under BnB-4bit matches FP16 pass@$100$ at 0.8598, Llama3.1-8B under AWQ slightly exceeds FP16 (0.8415 vs. 0.8354), and Qwen2.5-14B under GPTQ-INT4 improves pass@$100$ from 0.8415 to 0.8598. Larger models show the same pattern: Qwen2.5-32B under GPTQ-INT4 reaches 0.8476 compared with 0.8171 for FP16, and Llama-70B under AWQ reaches 0.9268 compared with 0.8963 for FP16. Sparse models remain weaker, but the gap between pass@$1$ and pass@$100$ reveals residual recoverability. For instance, Qwen2.5-7B under 2:4 sparsity rises from sampling pass@$1$ of 0.1745 to pass@$100$ of 0.3780, and Llama-70B under 2:4 sparsity rises from 0.4334 to 0.6646. Unstructured sparsity preserves more of this sampled program space, reaching pass@$100$ values of 0.7622 for Qwen2.5-7B, 0.7866 for Qwen2.5-14B, 0.8232 for Qwen2.5-32B, and 0.8659 for Llama-70B.

\textit{Executable-output reading.}
Together, the two HumanEval tables support the procedural interpretation in the main text. Quantization usually preserves both first-sample executable consistency and broader sampling-based recoverability. Sparsity, especially 2:4 sparsity, sharply reduces greedy and low-budget success, but pass@$100$ shows that some correct programs may still remain in the model's output distribution. This separation justifies reporting decoding-specific pass@$k$ views rather than collapsing HumanEval into a single scalar score: compression can damage the first decoded program while leaving part of the valid program space recoverable under sampling.

\begin{table*}[t]
\centering
\apptablestyle
\definecolor{ACEHeaderBg}{HTML}{F1F3F5}
\definecolor{ACEHeaderText}{HTML}{1F2D3D}
\definecolor{ACEProcHi}{HTML}{EAF3FB}
\definecolor{ACEProcBorder}{HTML}{2F75B5}
\definecolor{ACERuleDark}{HTML}{2E3640}
\newcommand{\trendcell}[1]{\cellcolor{ACEProcHi}#1}
\newcommand{\keycell}[1]{\begingroup\setlength{\fboxsep}{1.2pt}\fcolorbox{ACEProcBorder}{ACEProcHi}{#1}\endgroup}
\newcommand{\anchorval}[1]{\underline{#1}}
\newcommand{\heHeader}[1]{\textcolor{ACEHeaderText}{\textbf{#1}}}
\newcommand{\heTopRule}{\arrayrulecolor{ACERuleDark}\toprule[0.55pt]\arrayrulecolor{ACERuleDark}}
\newcommand{\heHeadRule}{\arrayrulecolor{ACERuleDark}\midrule[0.42pt]\arrayrulecolor{black}}
\newcommand{\heMidRule}{\arrayrulecolor{ACERuleDark}\midrule[0.24pt]\arrayrulecolor{black}}
\newcommand{\heBottomRule}{\arrayrulecolor{ACERuleDark}\bottomrule[0.55pt]\arrayrulecolor{black}}
\caption{HumanEval results for the procedural reasoning and constraint execution dimension under greedy decoding and low-budget sampling. Greedy pass@1 captures first-sample executable consistency, while n=20 sampling reports short-budget recoverability.}
\label{tab:appendix_humaneval_greedy_n20}
\resizebox{\textwidth}{!}{%
\begin{tabular}{llccc}
\heTopRule
\rowcolor{ACEHeaderBg}
\heHeader{Model Group} & \heHeader{Variant} & \heHeader{Greedy $p@1$} & \heHeader{Sampling $p@1$ ($n=20$)} & \heHeader{Sampling $p@10$ ($n=20$)} \\
\heHeadRule
\multirow{6}{*}{\textbf{Qwen2.5-7B}} & FP16 & \anchorval{0.7927 $\pm$ 0.0318} & \anchorval{0.7034 $\pm$ 0.0315} & \anchorval{0.8172 $\pm$ 0.0294} \\
 & BnB-4bit & \trendcell{0.7683 $\pm$ 0.0330} & \trendcell{0.7034 $\pm$ 0.0315} & \trendcell{0.8172 $\pm$ 0.0294} \\
 & AWQ & \trendcell{0.7744 $\pm$ 0.0327} & 0.6424 $\pm$ 0.0326 & 0.7837 $\pm$ 0.0312 \\
 & GPTQ-INT4 & \trendcell{0.7683 $\pm$ 0.0330} & 0.6598 $\pm$ 0.0320 & 0.7966 $\pm$ 0.0306 \\
 & 2:4-Sparse & \keycell{0.3354 $\pm$ 0.0370} & 0.1692 $\pm$ 0.0268 & 0.2636 $\pm$ 0.0329 \\
 & Unstructured-Sparse & \keycell{0.6890 $\pm$ 0.0363} & 0.4936 $\pm$ 0.0339 & 0.6757 $\pm$ 0.0349 \\
\heMidRule
\multirow{6}{*}{\textbf{Llama3.1-8B}} & FP16 & \anchorval{0.6951 $\pm$ 0.0361} & 0.6293 $\pm$ 0.0338 & \anchorval{0.7586 $\pm$ 0.0324} \\
 & BnB-4bit & 0.6829 $\pm$ 0.0364 & 0.6241 $\pm$ 0.0338 & 0.7575 $\pm$ 0.0322 \\
 & AWQ & 0.6646 $\pm$ 0.0370 & 0.6116 $\pm$ 0.0329 & \keycell{0.7816 $\pm$ 0.0306} \\
 & GPTQ-INT4 & 0.6402 $\pm$ 0.0376 & 0.5680 $\pm$ 0.0343 & 0.7253 $\pm$ 0.0327 \\
 & 2:4-Sparse & \keycell{0.1280 $\pm$ 0.0262} & 0.1192 $\pm$ 0.0239 & \keycell{0.1518 $\pm$ 0.0277} \\
 & Unstructured-Sparse & \keycell{0.4634 $\pm$ 0.0391} & 0.3777 $\pm$ 0.0346 & 0.5054 $\pm$ 0.0379 \\
\heMidRule
\multirow{7}{*}{\textbf{Qwen2.5-14B}} & FP16 & 0.8415 $\pm$ 0.0286 & 0.6893 $\pm$ 0.0322 & 0.8121 $\pm$ 0.0292 \\
 & BnB-4bit & 0.7927 $\pm$ 0.0318 & 0.6881 $\pm$ 0.0321 & 0.8122 $\pm$ 0.0292 \\
 & AWQ & 0.8171 $\pm$ 0.0303 & 0.6692 $\pm$ 0.0321 & 0.7974 $\pm$ 0.0303 \\
 & GPTQ-INT4 & 0.7805 $\pm$ 0.0324 & 0.6537 $\pm$ 0.0320 & 0.7941 $\pm$ 0.0304 \\
 & 2:4-Sparse & 0.4573 $\pm$ 0.0390 & 0.2250 $\pm$ 0.0276 & 0.3778 $\pm$ 0.0365 \\
 & Unstructured-Sparse & 0.7195 $\pm$ 0.0352 & 0.5326 $\pm$ 0.0338 & 0.7063 $\pm$ 0.0343 \\
 & Trim & 0.8537 $\pm$ 0.0277 & 0.6893 $\pm$ 0.0322 & 0.8121 $\pm$ 0.0292 \\
\heMidRule
\multirow{5}{*}{\textbf{DeepSeek-V2-Lite-16B}} & FP16 & \anchorval{0.5061 $\pm$ 0.0392} & 0.4784 $\pm$ 0.0342 & \anchorval{0.6535 $\pm$ 0.0354} \\
 & BnB-4bit & 0.4451 $\pm$ 0.0389 & 0.4784 $\pm$ 0.0342 & 0.6535 $\pm$ 0.0354 \\
 & AWQ & \keycell{0.2256 $\pm$ 0.0327} & 0.2317 $\pm$ 0.0281 & 0.4143 $\pm$ 0.0364 \\
 & 2:4-Sparse & \keycell{0.0549 $\pm$ 0.0178} & 0.0671 $\pm$ 0.0175 & \keycell{0.1021 $\pm$ 0.0232} \\
 & Unstructured-Sparse & 0.1220 $\pm$ 0.0256 & 0.1735 $\pm$ 0.0261 & 0.3044 $\pm$ 0.0340 \\
\heMidRule
\multirow{6}{*}{\textbf{Qwen2.5-32B}} & FP16 & \anchorval{0.8780 $\pm$ 0.0256} & 0.6485 $\pm$ 0.0332 & 0.7841 $\pm$ 0.0313 \\
 & BnB-4bit & 0.8720 $\pm$ 0.0262 & 0.6534 $\pm$ 0.0331 & 0.7767 $\pm$ 0.0320 \\
 & AWQ & \keycell{0.8902 $\pm$ 0.0245} & 0.6625 $\pm$ 0.0327 & 0.7890 $\pm$ 0.0311 \\
 & GPTQ-INT4 & \keycell{0.8963 $\pm$ 0.0239} & 0.6713 $\pm$ 0.0326 & 0.8049 $\pm$ 0.0295 \\
 & 2:4-Sparse & \keycell{0.6341 $\pm$ 0.0377} & 0.3649 $\pm$ 0.0341 & 0.5006 $\pm$ 0.0374 \\
 & Unstructured-Sparse & 0.7805 $\pm$ 0.0324 & 0.5921 $\pm$ 0.0327 & 0.7813 $\pm$ 0.0302 \\
\heMidRule
\multirow{7}{*}{\textbf{Llama-70B}} & FP16 & \anchorval{0.8171 $\pm$ 0.0303} & 0.7973 $\pm$ 0.0285 & 0.8689 $\pm$ 0.0258 \\
 & BnB-4bit & \trendcell{0.8171 $\pm$ 0.0303} & 0.7939 $\pm$ 0.0286 & 0.8681 $\pm$ 0.0258 \\
 & AWQ & 0.8110 $\pm$ 0.0307 & 0.7957 $\pm$ 0.0276 & 0.8875 $\pm$ 0.0240 \\
 & GPTQ-INT4 & \keycell{0.8293 $\pm$ 0.0295} & 0.7927 $\pm$ 0.0287 & 0.8706 $\pm$ 0.0249 \\
 & GPTQ-INT8 & \keycell{0.8293 $\pm$ 0.0295} & 0.7927 $\pm$ 0.0287 & 0.8706 $\pm$ 0.0249 \\
 & 2:4-Sparse & \keycell{0.5244 $\pm$ 0.0391} & 0.4338 $\pm$ 0.0345 & 0.5871 $\pm$ 0.0369 \\
 & Unstructured-Sparse & \keycell{0.7622 $\pm$ 0.0333} & 0.7128 $\pm$ 0.0310 & 0.8306 $\pm$ 0.0286 \\
\heBottomRule
\end{tabular}
}
\end{table*}

\begin{table*}[t]
\centering
\apptablestyle
\definecolor{ACEHeaderBg}{HTML}{F1F3F5}
\definecolor{ACEHeaderText}{HTML}{1F2D3D}
\definecolor{ACEProcHi}{HTML}{EAF3FB}
\definecolor{ACEProcBorder}{HTML}{2F75B5}
\definecolor{ACERuleDark}{HTML}{2E3640}
\newcommand{\trendcell}[1]{\cellcolor{ACEProcHi}#1}
\newcommand{\keycell}[1]{\begingroup\setlength{\fboxsep}{1.2pt}\fcolorbox{ACEProcBorder}{ACEProcHi}{#1}\endgroup}
\newcommand{\anchorval}[1]{\underline{#1}}
\newcommand{\heHeader}[1]{\textcolor{ACEHeaderText}{\textbf{#1}}}
\newcommand{\heTopRule}{\arrayrulecolor{ACERuleDark}\toprule[0.55pt]\arrayrulecolor{ACERuleDark}}
\newcommand{\heHeadRule}{\arrayrulecolor{ACERuleDark}\midrule[0.42pt]\arrayrulecolor{black}}
\newcommand{\heMidRule}{\arrayrulecolor{ACERuleDark}\midrule[0.24pt]\arrayrulecolor{black}}
\newcommand{\heBottomRule}{\arrayrulecolor{ACERuleDark}\bottomrule[0.55pt]\arrayrulecolor{black}}
\caption{HumanEval high-budget sampling results for the procedural reasoning and constraint execution dimension. The n=100 pass@k views audit whether correct programs remain recoverable beyond the first decoded solution.}
\label{tab:appendix_humaneval_n100}
\resizebox{\textwidth}{!}{%
\begin{tabular}{llccc}
\heTopRule
\rowcolor{ACEHeaderBg}
\heHeader{Model Group} & \heHeader{Variant} & \heHeader{Sampling $p@1$ ($n=100$)} & \heHeader{Sampling $p@10$ ($n=100$)} & \heHeader{Sampling $p@100$ ($n=100$)} \\
\heHeadRule
\multirow{6}{*}{\textbf{Qwen2.5-7B}} & FP16 & 0.7021 $\pm$ 0.0315 & 0.8173 $\pm$ 0.0283 & \anchorval{0.8598 $\pm$ 0.0272} \\
 & BnB-4bit & 0.7029 $\pm$ 0.0315 & 0.8157 $\pm$ 0.0283 & \trendcell{0.8598 $\pm$ 0.0272} \\
 & AWQ & 0.6453 $\pm$ 0.0325 & 0.7791 $\pm$ 0.0306 & 0.8293 $\pm$ 0.0295 \\
 & GPTQ-INT4 & 0.6549 $\pm$ 0.0321 & 0.7920 $\pm$ 0.0305 & 0.8476 $\pm$ 0.0282 \\
 & 2:4-Sparse & \keycell{0.1745 $\pm$ 0.0268} & 0.2665 $\pm$ 0.0322 & \keycell{0.3780 $\pm$ 0.0380} \\
 & Unstructured-Sparse & 0.4926 $\pm$ 0.0335 & 0.6767 $\pm$ 0.0340 & \keycell{0.7622 $\pm$ 0.0333} \\
\heMidRule
\multirow{6}{*}{\textbf{Llama3.1-8B}} & FP16 & 0.6305 $\pm$ 0.0334 & 0.7603 $\pm$ 0.0316 & \anchorval{0.8354 $\pm$ 0.0290} \\
 & BnB-4bit & 0.6274 $\pm$ 0.0334 & 0.7646 $\pm$ 0.0312 & 0.8354 $\pm$ 0.0290 \\
 & AWQ & 0.6040 $\pm$ 0.0330 & 0.7651 $\pm$ 0.0304 & \keycell{0.8415 $\pm$ 0.0286} \\
 & GPTQ-INT4 & 0.5705 $\pm$ 0.0342 & 0.7205 $\pm$ 0.0323 & 0.8049 $\pm$ 0.0310 \\
 & 2:4-Sparse & 0.1198 $\pm$ 0.0238 & 0.1537 $\pm$ 0.0275 & 0.1768 $\pm$ 0.0299 \\
 & Unstructured-Sparse & 0.3740 $\pm$ 0.0341 & 0.5075 $\pm$ 0.0372 & 0.6037 $\pm$ 0.0383 \\
\heMidRule
\multirow{7}{*}{\textbf{Qwen2.5-14B}} & FP16 & 0.6892 $\pm$ 0.0321 & 0.8021 $\pm$ 0.0298 & \anchorval{0.8415 $\pm$ 0.0286} \\
 & BnB-4bit & 0.6897 $\pm$ 0.0319 & 0.8046 $\pm$ 0.0299 & 0.8415 $\pm$ 0.0286 \\
 & AWQ & 0.6666 $\pm$ 0.0322 & 0.7956 $\pm$ 0.0296 & 0.8415 $\pm$ 0.0286 \\
 & GPTQ-INT4 & 0.6572 $\pm$ 0.0315 & 0.8026 $\pm$ 0.0293 & \keycell{0.8598 $\pm$ 0.0272} \\
 & 2:4-Sparse & 0.2276 $\pm$ 0.0276 & 0.3810 $\pm$ 0.0359 & 0.4573 $\pm$ 0.0390 \\
 & Unstructured-Sparse & 0.5284 $\pm$ 0.0336 & 0.7047 $\pm$ 0.0335 & \keycell{0.7866 $\pm$ 0.0321} \\
 & Trim & 0.6855 $\pm$ 0.0320 & 0.8074 $\pm$ 0.0296 & 0.8476 $\pm$ 0.0282 \\
\heMidRule
\multirow{5}{*}{\textbf{DeepSeek-V2-Lite-16B}} & FP16 & 0.4904 $\pm$ 0.0340 & 0.6619 $\pm$ 0.0344 & 0.7500 $\pm$ 0.0339 \\
 & BnB-4bit & 0.4894 $\pm$ 0.0339 & 0.6642 $\pm$ 0.0341 & 0.7744 $\pm$ 0.0327 \\
 & AWQ & 0.2308 $\pm$ 0.0275 & 0.4295 $\pm$ 0.0350 & 0.5549 $\pm$ 0.0389 \\
 & 2:4-Sparse & 0.0703 $\pm$ 0.0177 & 0.1164 $\pm$ 0.0234 & 0.1646 $\pm$ 0.0290 \\
 & Unstructured-Sparse & 0.1729 $\pm$ 0.0258 & 0.3040 $\pm$ 0.0326 & 0.4207 $\pm$ 0.0387 \\
\heMidRule
\multirow{6}{*}{\textbf{Qwen2.5-32B}} & FP16 & 0.6528 $\pm$ 0.0329 & 0.7808 $\pm$ 0.0310 & \anchorval{0.8171 $\pm$ 0.0303} \\
 & BnB-4bit & 0.6520 $\pm$ 0.0330 & 0.7806 $\pm$ 0.0308 & 0.8293 $\pm$ 0.0295 \\
 & AWQ & 0.6601 $\pm$ 0.0329 & 0.7846 $\pm$ 0.0307 & 0.8171 $\pm$ 0.0303 \\
 & GPTQ-INT4 & 0.6654 $\pm$ 0.0325 & 0.7977 $\pm$ 0.0295 & \keycell{0.8476 $\pm$ 0.0282} \\
 & 2:4-Sparse & 0.3666 $\pm$ 0.0338 & 0.5094 $\pm$ 0.0361 & 0.6098 $\pm$ 0.0382 \\
 & Unstructured-Sparse & 0.5961 $\pm$ 0.0328 & 0.7683 $\pm$ 0.0305 & \keycell{0.8232 $\pm$ 0.0299} \\
\heMidRule
\multirow{7}{*}{\textbf{Llama-70B}} & FP16 & 0.7959 $\pm$ 0.0283 & 0.8695 $\pm$ 0.0252 & \anchorval{0.8963 $\pm$ 0.0239} \\
 & BnB-4bit & 0.7959 $\pm$ 0.0283 & 0.8695 $\pm$ 0.0252 & 0.8963 $\pm$ 0.0239 \\
 & AWQ & 0.7924 $\pm$ 0.0280 & 0.8813 $\pm$ 0.0231 & \keycell{0.9268 $\pm$ 0.0204} \\
 & GPTQ-INT4 & 0.7920 $\pm$ 0.0288 & 0.8615 $\pm$ 0.0259 & 0.8902 $\pm$ 0.0245 \\
 & GPTQ-INT8 & 0.7920 $\pm$ 0.0288 & 0.8615 $\pm$ 0.0259 & 0.8902 $\pm$ 0.0245 \\
 & 2:4-Sparse & \keycell{0.4334 $\pm$ 0.0340} & 0.5954 $\pm$ 0.0362 & \keycell{0.6646 $\pm$ 0.0370} \\
 & Unstructured-Sparse & 0.7112 $\pm$ 0.0310 & 0.8291 $\pm$ 0.0278 & \keycell{0.8659 $\pm$ 0.0267} \\
\heBottomRule
\end{tabular}
}
\end{table*}

\subsection{Knowledge-Related Abilities}
\label{sec:app_knowledge}

This subsection expands the knowledge dimension with tables that preserve the paper's benchmark-native decomposition into truthful selection, generation-side truth-related output quality, broad knowledge, and role-grounded contextual knowledge.

\paragraph{TruthfulQA.}
Table~\ref{tab:appendix_truthfulqa} retains the TruthfulQA metrics used in the paper's knowledge-grounding analysis: 0-shot MC1/MC2, 0-shot generation-side BLEURT-max, and 3-shot MC1/MC2. MC1 and MC2 provide selection-side truthfulness views, while BLEURT-max provides a generation-side reference-compatibility view. Keeping these columns together is important because compression can change constrained truthful selection and generation-side reference similarity in different directions.

\textit{Selection-side truthfulness.}
Quantized variants usually remain close to their FP16 parents on MC1 and MC2, but the preservation is not perfectly uniform. Qwen2.5-14B, for example, keeps 0-shot MC1 near the FP16 value of 0.5532 under BnB-4bit, AWQ, and GPTQ-INT4 (0.5410, 0.5398, and 0.5471), and its 0-shot MC2 remains similarly close to the FP16 value of 0.7084. Qwen2.5-32B also shows small changes under quantization, with 0-shot MC1 moving from 0.5226 to 0.5202, 0.5165, and 0.5031. DeepSeek-V2-Lite-16B is a useful exception: BnB-4bit slightly improves its 0-shot selection scores relative to FP16 (MC1: 0.3635 vs. 0.3525; MC2: 0.5390 vs. 0.5258), while AWQ is closer to the FP16 baseline. These cases show that selection-side truthfulness is often stable under quantization, but the direction and size of the change remain model- and method-dependent.

\textit{Sparse selection degradation.}
Sparse variants generally reduce MC1 and MC2 more clearly than quantized variants. For Llama3.1-8B, 0-shot MC1 drops from 0.4039 to 0.2852 under 2:4 sparsity and to 0.3182 under unstructured sparsity. Qwen2.5-14B shows a similar pattern: 0-shot MC1 falls from 0.5532 to 0.4443 under 2:4 sparsity and 0.4871 under unstructured sparsity. The larger Llama-70B also loses selection-side truthfulness under sparsity, with 0-shot MC1 dropping from 0.5006 to 0.3501 under 2:4 sparsity and 0.4174 under unstructured sparsity. Across these examples, unstructured sparsity is usually less damaging than 2:4 sparsity, but both remain below the FP16 parent on the selection metrics.

\textit{Generation-side reference compatibility.}
Generation-side BLEURT-max does not move in lockstep with MC1 and MC2. In several sparse settings, BLEURT-max becomes less negative even while selection-side truthfulness declines. Llama3.1-8B illustrates this separation clearly: its 0-shot MC1 falls from 0.4039 to 0.2852 under 2:4 sparsity, while BLEURT-max increases from -0.6031 to -0.1793. Llama-70B shows an even stronger generation-side shift: BLEURT-max moves from -0.2596 for FP16 to 0.0112 under 2:4 sparsity and 0.1522 under unstructured sparsity, while 0-shot MC1 still drops to 0.3501 and 0.4174. This pattern supports the main text's interpretation that higher generation-side reference compatibility after compression should not be read as improved truthfulness. It can instead indicate a shift toward BLEURT-friendly generation while constrained truthful selection weakens.

\textit{Shot setting.}
The 3-shot MC1/MC2 columns show that adding demonstrations does not remove the selection-side fragility introduced by compression. For Qwen2.5-14B, 3-shot MC1 drops from 0.5243 for FP16 to 0.3749 under 2:4 sparsity and 0.4264 under unstructured sparsity. Qwen2.5-32B drops from 0.4916 to 0.3819 under 2:4 sparsity, while Llama-70B drops from 0.4492 to 0.3484. These 3-shot results are consistent with the 0-shot view: sparse compression weakens constrained truthful selection even when the benchmark provides few-shot context.

\textit{Knowledge-truthfulness reading.}
Together, the selected TruthfulQA columns support the knowledge-dimension claim in the main text. Quantization often preserves selection-side truthfulness with small changes, whereas sparsity causes clearer MC1/MC2 degradation. At the same time, generation-side BLEURT-max can improve or become less negative under sparse compression, producing a generation--selection separation. The appendix table therefore justifies reporting TruthfulQA with benchmark-native selection and generation views instead of collapsing it into a single truthfulness score.

\begin{table*}[t]
\centering
\apptablestyle
\definecolor{ACEHeaderBg}{HTML}{F1F3F5}
\definecolor{ACEHeaderText}{HTML}{1F2D3D}
\definecolor{ACEKnowHi}{HTML}{F0F7ED}
\definecolor{ACEKnowBorder}{HTML}{5B8C3A}
\definecolor{ACERuleDark}{HTML}{2E3640}
\newcommand{\trendcell}[1]{\cellcolor{ACEKnowHi}#1}
\newcommand{\keycell}[1]{\begingroup\setlength{\fboxsep}{1.2pt}\fcolorbox{ACEKnowBorder}{ACEKnowHi}{#1}\endgroup}
\newcommand{\anchorval}[1]{\underline{#1}}
\newcommand{\tqaHeader}[1]{\textcolor{ACEHeaderText}{\textbf{#1}}}
\newcommand{\hdrstrut}{\rule{0pt}{2ex}}
\newcommand{\tqaGroupHeadTwo}[5]{%
  \begin{tabular}[c]{@{}c@{}}
    \rule{0pt}{2.15ex}\textcolor{ACEHeaderText}{\textbf{#1}}\\[-0.55ex]
    \rule{#3}{0.4pt}\\[-0.15ex]
    \makebox[#2][c]{%
      \makebox[\dimexpr#2/2\relax][c]{\tqaHeader{#4}\hdrstrut}%
      \makebox[\dimexpr#2/2\relax][c]{\tqaHeader{#5}\hdrstrut}%
    }
  \end{tabular}%
}
\newcommand{\tqaGroupHeadThree}[6]{%
  \begin{tabular}[c]{@{}c@{}}
    \rule{0pt}{2.15ex}\textcolor{ACEHeaderText}{\textbf{#1}}\\[-0.55ex]
    \rule{#3}{0.4pt}\\[-0.15ex]
    \makebox[#2][c]{%
      \makebox[\dimexpr#4\relax][c]{\tqaHeader{MC1}\hdrstrut}%
      \makebox[\dimexpr#5\relax][c]{\tqaHeader{MC2}\hdrstrut}%
      \makebox[\dimexpr#6\relax][c]{\tqaHeader{Gen-BLEURT-max}\hdrstrut}%
    }
  \end{tabular}%
}
\newcommand{\tqaTopRule}{\arrayrulecolor{ACERuleDark}\toprule[0.55pt]\arrayrulecolor{ACERuleDark}}
\newcommand{\tqaHeadRule}{\arrayrulecolor{ACERuleDark}\midrule[0.42pt]\arrayrulecolor{black}}
\newcommand{\tqaMidRule}{\arrayrulecolor{ACERuleDark}\midrule[0.24pt]\arrayrulecolor{black}}
\newcommand{\tqaBottomRule}{\arrayrulecolor{ACERuleDark}\bottomrule[0.55pt]\arrayrulecolor{black}}
\caption{Selected TruthfulQA results for the knowledge grounding and truthfulness dimension. MC1/MC2 report selection-side truthfulness, while Gen-BLEURT-max reports generation-side reference compatibility; the 3-shot columns audit selection behavior under few-shot context.}
\label{tab:appendix_truthfulqa}
\resizebox{\textwidth}{!}{%
\begin{tabular}{llccccc}
\tqaTopRule
\rowcolor{ACEHeaderBg}
\begin{tabular}[c]{@{}l@{}}\tqaHeader{Model Group}\end{tabular} & \begin{tabular}[c]{@{}l@{}}\tqaHeader{Variant}\end{tabular} & \multicolumn{3}{c}{\tqaGroupHeadThree{0-shot}{24.6em}{23.8em}{7.8em}{7.8em}{9em}} & \multicolumn{2}{c}{\tqaGroupHeadTwo{3-shot}{13.6em}{14.5em}{MC1}{MC2}} \\
\tqaHeadRule
\multirow{6}{*}{\textbf{Qwen2.5-7B}} & FP16 & 0.4737 $\pm$ 0.0175 & 0.6306 $\pm$ 0.0160 & -0.6324 $\pm$ 0.0135 & 0.4712 $\pm$ 0.0044 & 0.6362 $\pm$ 0.0017 \\
 & BnB-4bit & 0.4663 $\pm$ 0.0175 & 0.6229 $\pm$ 0.0160 & -0.6418 $\pm$ 0.0134 & 0.4716 $\pm$ 0.0043 & 0.6317 $\pm$ 0.0021 \\
 & AWQ & 0.4480 $\pm$ 0.0174 & 0.6149 $\pm$ 0.0160 & -0.6234 $\pm$ 0.0140 & 0.4598 $\pm$ 0.0062 & 0.6294 $\pm$ 0.0004 \\
 & GPTQ-INT4 & 0.4688 $\pm$ 0.0175 & 0.6309 $\pm$ 0.0160 & -0.6224 $\pm$ 0.0134 & 0.4639 $\pm$ 0.0056 & 0.6271 $\pm$ 0.0042 \\
 & 2:4-Sparse & 0.4284 $\pm$ 0.0173 & 0.5913 $\pm$ 0.0162 & -0.6436 $\pm$ 0.0128 & 0.4186 $\pm$ 0.0032 & 0.5873 $\pm$ 0.0049 \\
 & Unstructured-Sparse & 0.4296 $\pm$ 0.0173 & 0.5931 $\pm$ 0.0161 & -0.5756 $\pm$ 0.0133 & 0.4353 $\pm$ 0.0049 & 0.6014 $\pm$ 0.0028 \\
\tqaMidRule
\multirow{6}{*}{\textbf{Llama3.1-8B}} & FP16 & \anchorval{0.4039 $\pm$ 0.0172} & 0.5509 $\pm$ 0.0160 & \anchorval{-0.6031 $\pm$ 0.0147} & 0.3900 $\pm$ 0.0043 & 0.5546 $\pm$ 0.0034 \\
 & BnB-4bit & 0.3978 $\pm$ 0.0171 & 0.5685 $\pm$ 0.0158 & -0.6267 $\pm$ 0.0156 & 0.3888 $\pm$ 0.0051 & 0.5648 $\pm$ 0.0044 \\
 & AWQ & 0.3794 $\pm$ 0.0170 & 0.5340 $\pm$ 0.0159 & -0.7905 $\pm$ 0.0155 & 0.3741 $\pm$ 0.0035 & 0.5392 $\pm$ 0.0034 \\
 & GPTQ-INT4 & 0.3880 $\pm$ 0.0171 & 0.5581 $\pm$ 0.0160 & -0.4235 $\pm$ 0.0155 & 0.3688 $\pm$ 0.0014 & 0.5364 $\pm$ 0.0041 \\
 & 2:4-Sparse & \keycell{0.2852 $\pm$ 0.0158} & 0.4617 $\pm$ 0.0153 & \keycell{-0.1793 $\pm$ 0.0169} & 0.2966 $\pm$ 0.0007 & 0.4578 $\pm$ 0.0038 \\
 & Unstructured-Sparse & \keycell{0.3182 $\pm$ 0.0163} & 0.4838 $\pm$ 0.0152 & \keycell{-0.2170 $\pm$ 0.0198} & 0.3150 $\pm$ 0.0102 & 0.4750 $\pm$ 0.0059 \\
\tqaMidRule
\multirow{7}{*}{\textbf{Qwen2.5-14B}} & FP16 & \anchorval{0.5532 $\pm$ 0.0174} & \anchorval{0.7084 $\pm$ 0.0154} & -0.6716 $\pm$ 0.0124 & \anchorval{0.5243 $\pm$ 0.0070} & \anchorval{0.6855 $\pm$ 0.0038} \\
 & BnB-4bit & \trendcell{0.5410 $\pm$ 0.0174} & \trendcell{0.6918 $\pm$ 0.0156} & -0.6663 $\pm$ 0.0127 & 0.5075 $\pm$ 0.0049 & 0.6740 $\pm$ 0.0023 \\
 & AWQ & \trendcell{0.5398 $\pm$ 0.0174} & \trendcell{0.6853 $\pm$ 0.0154} & -0.6414 $\pm$ 0.0134 & 0.5088 $\pm$ 0.0075 & 0.6591 $\pm$ 0.0043 \\
 & GPTQ-INT4 & \trendcell{0.5471 $\pm$ 0.0174} & \trendcell{0.6971 $\pm$ 0.0153} & -0.6615 $\pm$ 0.0127 & 0.5182 $\pm$ 0.0014 & 0.6785 $\pm$ 0.0008 \\
 & 2:4-Sparse & \keycell{0.4443 $\pm$ 0.0174} & 0.5825 $\pm$ 0.0164 & -0.5976 $\pm$ 0.0117 & \keycell{0.3749 $\pm$ 0.0058} & \trendcell{0.5275 $\pm$ 0.0013} \\
 & Unstructured-Sparse & \keycell{0.4871 $\pm$ 0.0175} & 0.6480 $\pm$ 0.0159 & -0.6134 $\pm$ 0.0127 & \keycell{0.4264 $\pm$ 0.0039} & \trendcell{0.5860 $\pm$ 0.0068} \\
 & Trim & 0.5606 $\pm$ 0.0174 & 0.7094 $\pm$ 0.0155 & -0.6560 $\pm$ 0.0135 & 0.5243 $\pm$ 0.0070 & 0.6855 $\pm$ 0.0038 \\
\tqaMidRule
\multirow{5}{*}{\textbf{DeepSeek-V2-Lite-16B}} & FP16 & \anchorval{0.3525 $\pm$ 0.0167} & \anchorval{0.5258 $\pm$ 0.0161} & -0.6388 $\pm$ 0.0129 & 0.3472 $\pm$ 0.0055 & 0.5001 $\pm$ 0.0018 \\
 & BnB-4bit & \keycell{0.3635 $\pm$ 0.0168} & \keycell{0.5390 $\pm$ 0.0160} & -0.6346 $\pm$ 0.0127 & 0.3407 $\pm$ 0.0099 & 0.4916 $\pm$ 0.0069 \\
 & AWQ & 0.3403 $\pm$ 0.0166 & 0.5176 $\pm$ 0.0161 & -0.6923 $\pm$ 0.0137 & 0.3407 $\pm$ 0.0028 & 0.5082 $\pm$ 0.0041 \\
 & 2:4-Sparse & 0.3207 $\pm$ 0.0163 & 0.4637 $\pm$ 0.0158 & -0.3475 $\pm$ 0.0162 & 0.3154 $\pm$ 0.0051 & 0.4781 $\pm$ 0.0040 \\
 & Unstructured-Sparse & 0.3244 $\pm$ 0.0164 & 0.4885 $\pm$ 0.0160 & -0.5124 $\pm$ 0.0159 & 0.2938 $\pm$ 0.0021 & 0.4663 $\pm$ 0.0035 \\
\tqaMidRule
\multirow{6}{*}{\textbf{Qwen2.5-32B}} & FP16 & \anchorval{0.5226 $\pm$ 0.0175} & 0.7037 $\pm$ 0.0151 & \anchorval{-0.6677 $\pm$ 0.0121} & \anchorval{0.4916 $\pm$ 0.0043} & 0.6549 $\pm$ 0.0036 \\
 & BnB-4bit & \trendcell{0.5202 $\pm$ 0.0175} & 0.6921 $\pm$ 0.0153 & -0.6487 $\pm$ 0.0126 & 0.4839 $\pm$ 0.0035 & 0.6496 $\pm$ 0.0045 \\
 & AWQ & \trendcell{0.5165 $\pm$ 0.0175} & 0.6890 $\pm$ 0.0154 & -0.6701 $\pm$ 0.0118 & 0.5006 $\pm$ 0.0049 & 0.6629 $\pm$ 0.0025 \\
 & GPTQ-INT4 & \trendcell{0.5031 $\pm$ 0.0175} & 0.6835 $\pm$ 0.0156 & -0.6483 $\pm$ 0.0124 & 0.4823 $\pm$ 0.0044 & 0.6444 $\pm$ 0.0034 \\
 & 2:4-Sparse & 0.4480 $\pm$ 0.0174 & 0.6053 $\pm$ 0.0163 & \trendcell{-0.4956 $\pm$ 0.0137} & \keycell{0.3819 $\pm$ 0.0086} & 0.5322 $\pm$ 0.0114 \\
 & Unstructured-Sparse & 0.4553 $\pm$ 0.0174 & 0.6486 $\pm$ 0.0159 & -0.6096 $\pm$ 0.0126 & 0.4512 $\pm$ 0.0080 & 0.6027 $\pm$ 0.0024 \\
\tqaMidRule
\multirow{7}{*}{\textbf{Llama-70B}} & FP16 & \anchorval{0.5006 $\pm$ 0.0175} & 0.6683 $\pm$ 0.0151 & \anchorval{-0.2596 $\pm$ 0.0223} & \anchorval{0.4492 $\pm$ 0.0056} & 0.6249 $\pm$ 0.0013 \\
 & BnB-4bit & 0.4823 $\pm$ 0.0175 & 0.6524 $\pm$ 0.0153 & -0.1973 $\pm$ 0.0243 & 0.4296 $\pm$ 0.0064 & 0.6091 $\pm$ 0.0017 \\
 & AWQ & 0.4627 $\pm$ 0.0175 & 0.6457 $\pm$ 0.0151 & -0.2162 $\pm$ 0.0226 & 0.4447 $\pm$ 0.0103 & 0.6181 $\pm$ 0.0003 \\
 & GPTQ-INT4 & 0.4896 $\pm$ 0.0175 & 0.6611 $\pm$ 0.0152 & -0.2872 $\pm$ 0.0222 & 0.4321 $\pm$ 0.0024 & 0.6069 $\pm$ 0.0037 \\
 & GPTQ-INT8 & 0.4957 $\pm$ 0.0175 & 0.6678 $\pm$ 0.0152 & -0.2654 $\pm$ 0.0219 & 0.4492 $\pm$ 0.0076 & 0.6256 $\pm$ 0.0016 \\
 & 2:4-Sparse & \keycell{0.3501 $\pm$ 0.0167} & 0.5241 $\pm$ 0.0156 & \keycell{0.0112 $\pm$ 0.0238} & \keycell{0.3484 $\pm$ 0.0093} & 0.5132 $\pm$ 0.0040 \\
 & Unstructured-Sparse & \keycell{0.4174 $\pm$ 0.0173} & 0.5997 $\pm$ 0.0156 & \keycell{0.1522 $\pm$ 0.0267} & 0.3970 $\pm$ 0.0062 & 0.5651 $\pm$ 0.0023 \\
\tqaBottomRule
\end{tabular}
}
\end{table*}

\paragraph{C-Eval.}
Table~\ref{tab:appendix_ceval} reports C-Eval validation and test accuracy for the knowledge grounding and truthfulness dimension. In the main text, C-Eval is used as the broad exam-knowledge counterpart to more role-grounded or truth-sensitive knowledge probes. The appendix therefore keeps both validation and test splits so that broad knowledge retention can be checked directly rather than inferred from a single selected score.

\textit{Quantization.}
Across most model groups, quantized variants remain close to their FP16 parents on test accuracy. Qwen2.5-7B changes only slightly under BnB-4bit, AWQ, and GPTQ-INT4, with test accuracy moving from 0.7931 to 0.7787, 0.7804, and 0.7785. Qwen2.5-14B shows a similar pattern, staying near its FP16 test score of 0.8303 under BnB-4bit, AWQ, and GPTQ-INT4 (0.8171, 0.8179, and 0.8117). The strongest preservation appears for Qwen2.5-32B, where all three quantized variants remain close to the FP16 test accuracy of 0.8817: BnB-4bit reaches 0.8768, AWQ reaches 0.8718, and GPTQ-INT4 reaches 0.8697. These results support the main-paper reading that broad exam knowledge is relatively stable under post-training quantization.

\textit{Model-family variation.}
The size of the quantization drop still depends on the base model. Llama3.1-8B declines from 0.5322 to 0.5118, 0.5160, and 0.5031 under BnB-4bit, AWQ, and GPTQ-INT4, respectively, while DeepSeek-V2-Lite-16B declines from 0.5832 to 0.5668 under BnB-4bit and 0.5436 under AWQ. Llama-70B is comparatively stable under AWQ and GPTQ-INT4, with test accuracy of 0.6655 and 0.6579 compared with 0.6749 for FP16, although its 4-bit row is lower at 0.6322. These differences show that broad knowledge retention is generally robust under quantization, but not completely method- or family-invariant.

\textit{Sparse degradation.}
Sparse variants show clearer degradation than quantized variants. Under 2:4 structured sparsity, Qwen2.5-7B test accuracy drops from 0.7931 to 0.4736, Qwen2.5-14B drops from 0.8303 to 0.5062, and DeepSeek-V2-Lite-16B drops from 0.5832 to 0.3136. The same pattern holds for the larger models: Qwen2.5-32B falls from 0.8817 to 0.6308, and Llama-70B falls from 0.6749 to 0.4491. These results indicate that broad exam-knowledge accuracy is more resilient than many fine-grained readouts in the main paper, but it is still substantially affected by structured sparsity.

\textit{Validation--test consistency.}
The validation and test columns generally move together across compression settings. This makes C-Eval a relatively stable broad-knowledge readout: the main interpretation does not depend on a single split behaving unusually. The appendix table therefore supports the knowledge-dimension claim that broad disciplinary knowledge is often retained more strongly than role-grounded or truth-sensitive behavior, while still showing clear degradation under sparse compression.

\begin{table*}[t]
\centering
\apptablestyle
\ifdefined\ACEHeaderBg\else\definecolor{ACEHeaderBg}{HTML}{F1F3F5}\fi
\ifdefined\ACEHeaderText\else\definecolor{ACEHeaderText}{HTML}{1F2D3D}\fi
\ifdefined\ACEKnowHi\else\definecolor{ACEKnowHi}{HTML}{F0F7ED}\fi
\ifdefined\ACEKnowBorder\else\definecolor{ACEKnowBorder}{HTML}{5B8C3A}\fi
\ifdefined\ACERuleDark\else\definecolor{ACERuleDark}{HTML}{2E3640}\fi
\providecommand{\trendcell}[1]{\cellcolor{ACEKnowHi}#1}
\providecommand{\keycell}[1]{\begingroup\setlength{\fboxsep}{1.2pt}\fcolorbox{ACEKnowBorder}{ACEKnowHi}{#1}\endgroup}
\providecommand{\anchorval}[1]{\underline{#1}}
\newcommand{\ceHeader}[1]{\textcolor{ACEHeaderText}{\textbf{#1}}}
\newcommand{\ceTopRule}{\arrayrulecolor{ACERuleDark}\toprule[0.55pt]\arrayrulecolor{ACERuleDark}}
\newcommand{\ceHeadRule}{\arrayrulecolor{ACERuleDark}\midrule[0.42pt]\arrayrulecolor{black}}
\newcommand{\ceMidRule}{\arrayrulecolor{ACERuleDark}\midrule[0.24pt]\arrayrulecolor{black}}
\newcommand{\ceBottomRule}{\arrayrulecolor{ACERuleDark}\bottomrule[0.55pt]\arrayrulecolor{black}}
\caption{C-Eval accuracy results for the knowledge grounding and truthfulness dimension. Validation and test splits are reported together to audit broad exam-knowledge retention under quantization and sparsity.}
\label{tab:appendix_ceval}
\begin{tabular}{llcc}
\ceTopRule
\rowcolor{ACEHeaderBg}
\ceHeader{Model Group} & \ceHeader{Variant} & \ceHeader{Valid Acc.} & \ceHeader{Test Acc.} \\
\ceHeadRule
\multirow{6}{*}{\textbf{Qwen2.5-7B}} & FP16 & 0.7964 $\pm$ 0.0105 & \anchorval{0.7931 $\pm$ 0.0035} \\
 & BnB-4bit & 0.7868 $\pm$ 0.0108 & \trendcell{0.7787 $\pm$ 0.0036} \\
 & AWQ & 0.7868 $\pm$ 0.0108 & \trendcell{0.7804 $\pm$ 0.0036} \\
 & GPTQ-INT4 & 0.7779 $\pm$ 0.0109 & \trendcell{0.7785 $\pm$ 0.0036} \\
 & 2:4-Sparse & 0.4985 $\pm$ 0.0133 & \keycell{0.4736 $\pm$ 0.0044} \\
 & Unstructured-Sparse & 0.6746 $\pm$ 0.0124 & \keycell{0.6674 $\pm$ 0.0041} \\
\ceMidRule
\multirow{6}{*}{\textbf{Llama3.1-8B}} & FP16 & 0.5394 $\pm$ 0.0134 & \anchorval{0.5322 $\pm$ 0.0044} \\
 & BnB-4bit & 0.5134 $\pm$ 0.0133 & \trendcell{0.5118 $\pm$ 0.0044} \\
 & AWQ & 0.5201 $\pm$ 0.0133 & \trendcell{0.5160 $\pm$ 0.0044} \\
 & GPTQ-INT4 & 0.5186 $\pm$ 0.0134 & \trendcell{0.5031 $\pm$ 0.0044} \\
 & 2:4-Sparse & 0.3076 $\pm$ 0.0126 & 0.3018 $\pm$ 0.0041 \\
 & Unstructured-Sparse & 0.3848 $\pm$ 0.0131 & 0.3832 $\pm$ 0.0043 \\
\ceMidRule
\multirow{7}{*}{\textbf{Qwen2.5-14B}} & FP16 & 0.8358 $\pm$ 0.0099 & \anchorval{0.8303 $\pm$ 0.0033} \\
 & BnB-4bit & 0.8172 $\pm$ 0.0102 & \trendcell{0.8171 $\pm$ 0.0034} \\
 & AWQ & 0.8262 $\pm$ 0.0101 & \trendcell{0.8179 $\pm$ 0.0033} \\
 & GPTQ-INT4 & 0.8291 $\pm$ 0.0101 & \trendcell{0.8117 $\pm$ 0.0034} \\
 & 2:4-Sparse & 0.5409 $\pm$ 0.0135 & \keycell{0.5062 $\pm$ 0.0044} \\
 & Unstructured-Sparse & 0.7184 $\pm$ 0.0119 & \keycell{0.6996 $\pm$ 0.0040} \\
 & Trim & 0.8358 $\pm$ 0.0099 & 0.8303 $\pm$ 0.0033 \\
\ceMidRule
\multirow{5}{*}{\textbf{DeepSeek-V2-Lite-16B}} & FP16 & 0.5996 $\pm$ 0.0128 & \anchorval{0.5832 $\pm$ 0.0042} \\
 & BnB-4bit & 0.5832 $\pm$ 0.0130 & \keycell{0.5668 $\pm$ 0.0043} \\
 & AWQ & 0.5550 $\pm$ 0.0131 & \keycell{0.5436 $\pm$ 0.0043} \\
 & 2:4-Sparse & 0.3091 $\pm$ 0.0125 & \keycell{0.3136 $\pm$ 0.0042} \\
 & Unstructured-Sparse & 0.4205 $\pm$ 0.0133 & 0.4274 $\pm$ 0.0044 \\
\ceMidRule
\multirow{6}{*}{\textbf{Qwen2.5-32B}} & FP16 & \anchorval{0.8804 $\pm$ 0.0086} & \anchorval{0.8817 $\pm$ 0.0028} \\
 & BnB-4bit & 0.8707 $\pm$ 0.0089 & \keycell{0.8768 $\pm$ 0.0029} \\
 & AWQ & 0.8752 $\pm$ 0.0088 & \keycell{0.8718 $\pm$ 0.0029} \\
 & GPTQ-INT4 & 0.8670 $\pm$ 0.0089 & \keycell{0.8697 $\pm$ 0.0029} \\
 & 2:4-Sparse & 0.6575 $\pm$ 0.0126 & \keycell{0.6308 $\pm$ 0.0042} \\
 & Unstructured-Sparse & 0.8016 $\pm$ 0.0106 & \keycell{0.7793 $\pm$ 0.0036} \\
\ceMidRule
\multirow{6}{*}{\textbf{Llama-70B}} & FP16 & 0.7065 $\pm$ 0.0122 & \anchorval{0.6749 $\pm$ 0.0041} \\
 & BnB-4bit & 0.6478 $\pm$ 0.0126 & \trendcell{0.6322 $\pm$ 0.0042} \\
 & AWQ & 0.6902 $\pm$ 0.0123 & \trendcell{0.6655 $\pm$ 0.0041} \\
 & GPTQ-INT4 & 0.6820 $\pm$ 0.0123 & \trendcell{0.6579 $\pm$ 0.0041} \\
 & 2:4-Sparse & 0.4688 $\pm$ 0.0136 & \keycell{0.4491 $\pm$ 0.0044} \\
 & Unstructured-Sparse & 0.6293 $\pm$ 0.0130 & \keycell{0.5834 $\pm$ 0.0043} \\
\ceBottomRule
\end{tabular}
\end{table*}

\paragraph{RoleEval.}
Table~\ref{tab:appendix_roleeval} reports grouped RoleEval slice means for the knowledge grounding and truthfulness dimension. RoleEval complements C-Eval by testing whether compressed models preserve role-grounded contextual knowledge rather than only broad exam knowledge.

\textit{Quantization.}
Quantized variants usually retain much of the RoleEval signal, although the preservation is less uniform than in C-Eval. Qwen2.5-14B remains close to its FP16 overall mean of 0.5913 under BnB-4bit and GPTQ-INT4 (0.5764 and 0.5775), with AWQ slightly lower at 0.5670. Qwen2.5-32B shows a similar but visible decline, from 0.6018 for FP16 to 0.5854, 0.5827, and 0.5751 under BnB-4bit, AWQ, and GPTQ-INT4. Llama-70B is relatively stable under AWQ and GPTQ-INT4 (0.5661 and 0.5577 versus 0.5807 for FP16), while BnB-4bit drops more clearly to 0.5170. These results indicate that role-grounded knowledge is often preserved under quantization, but with more model- and method-level variation than broad exam knowledge.

\textit{Sparse degradation.}
Sparse variants produce substantially larger drops in the across-shot overall mean. Under 2:4 sparsity, Qwen2.5-7B falls from 0.5141 to 0.3265, Llama3.1-8B falls from 0.4361 to 0.2696, and Qwen2.5-14B falls from 0.5913 to 0.3287. The larger models also degrade: Qwen2.5-32B drops from 0.6018 to 0.4050, and Llama-70B drops from 0.5807 to 0.3778. This supports the main text's distinction between broad knowledge retention and role-grounded contextual deployment: the latter is more sensitive to sparse compression.

\textit{Sparsity structure.}
Unstructured sparsity generally preserves more role-grounded knowledge than 2:4 sparsity, but it does not fully recover the FP16 level. Qwen2.5-7B improves from 0.3265 under 2:4 sparsity to 0.4130 under unstructured sparsity, Qwen2.5-14B improves from 0.3287 to 0.4710, and Qwen2.5-32B improves from 0.4050 to 0.5044. Llama-70B shows the same direction, rising from 0.3778 under 2:4 sparsity to 0.4909 under unstructured sparsity. DeepSeek-V2-Lite-16B shows a smaller difference, with 0.3307 under 2:4 sparsity and 0.3469 under unstructured sparsity. Overall, the RoleEval table again shows that the sparsity pattern matters for preserving knowledge use.

\textit{Shot and language slices.}
The EN/ZH slices show that role-grounded knowledge loss is not only an overall-score effect. Qwen-family FP16 models are stronger on ZH than EN, such as Qwen2.5-14B at 0.6051 ZH versus 0.5774 EN and Qwen2.5-32B at 0.6167 ZH versus 0.5867 EN. These bilingual differences are visible in both the 0-shot and 5-shot grouped columns, so the across-shot mean is not hiding a shot-specific reversal. Under 2:4 sparsity, the ZH slices drop sharply for Qwen2.5-14B and Qwen2.5-32B, while Llama-70B shows the opposite FP16 direction, with higher EN than ZH performance but degradation in both language slices. These language-specific shifts show that compression affects role-grounded contextual knowledge differently across language settings, rather than simply lowering a single global score.

\textit{Role-grounded knowledge reading.}
Together, the grouped RoleEval table supports the knowledge-dimension interpretation in the main text. Quantization often preserves much of the role-grounded signal, but sparse compression, especially 2:4 sparsity, weakens contextual knowledge deployment more sharply. The table also shows why RoleEval should not be collapsed into C-Eval: broad exam knowledge and role-grounded knowledge retention are related but not interchangeable signals.

\begin{table*}[t]
\centering
\apptablestyle
\ifdefined\ACEHeaderBg\else\definecolor{ACEHeaderBg}{HTML}{F1F3F5}\fi
\ifdefined\ACEHeaderText\else\definecolor{ACEHeaderText}{HTML}{1F2D3D}\fi
\ifdefined\ACEKnowHi\else\definecolor{ACEKnowHi}{HTML}{F0F7ED}\fi
\ifdefined\ACEKnowBorder\else\definecolor{ACEKnowBorder}{HTML}{5B8C3A}\fi
\ifdefined\ACERuleDark\else\definecolor{ACERuleDark}{HTML}{2E3640}\fi
\providecommand{\trendcell}[1]{\cellcolor{ACEKnowHi}#1}
\providecommand{\keycell}[1]{\begingroup\setlength{\fboxsep}{1.2pt}\fcolorbox{ACEKnowBorder}{ACEKnowHi}{#1}\endgroup}
\providecommand{\anchorval}[1]{\underline{#1}}
\newcommand{\reHeader}[1]{\textcolor{ACEHeaderText}{\textbf{#1}}}
\newcommand{\reTopRule}{\arrayrulecolor{ACERuleDark}\toprule[0.55pt]\arrayrulecolor{ACERuleDark}}
\newcommand{\reHeadRule}{\arrayrulecolor{ACERuleDark}\midrule[0.42pt]\arrayrulecolor{black}}
\newcommand{\reMidRule}{\arrayrulecolor{ACERuleDark}\midrule[0.24pt]\arrayrulecolor{black}}
\newcommand{\reBottomRule}{\arrayrulecolor{ACERuleDark}\bottomrule[0.55pt]\arrayrulecolor{black}}
\newcommand{\hdrstrut}{\rule{0pt}{2ex}}
\newcommand{\grouphead}[4]{%
  \begin{tabular}[c]{@{}c@{}}
    \rule{0pt}{2.15ex}\textcolor{ACEHeaderText}{\textbf{#1}}\\[-0.55ex]
    \rule{#2}{0.4pt}\\[-0.15ex]
    \makebox[\dimexpr#2/2\relax][c]{\reHeader{#3}\hdrstrut}%
    \makebox[\dimexpr#2/2\relax][c]{\reHeader{#4}\hdrstrut}%
  \end{tabular}%
}
\newcommand{\groupheadThree}[5]{%
  \begin{tabular}[c]{@{}c@{}}
    \rule{0pt}{2.15ex}\textcolor{ACEHeaderText}{\textbf{#1}}\\[-0.55ex]
    \rule{#2}{0.4pt}\\[-0.15ex]
    \makebox[\dimexpr#2/3\relax][c]{\reHeader{#3}\hdrstrut}%
    \makebox[\dimexpr#2/3\relax][c]{\reHeader{#4}\hdrstrut}%
    \makebox[\dimexpr#2/3\relax][c]{\reHeader{#5}\hdrstrut}%
  \end{tabular}%
}
\caption{RoleEval grouped slice means for the knowledge grounding and truthfulness dimension. Shot-specific and across-shot EN/ZH means show how role-grounded knowledge retention varies by language and compression setting.}
\label{tab:appendix_roleeval}
\begin{tabular}{llccccccc}
\reTopRule
\rowcolor{ACEHeaderBg}
\begin{tabular}[c]{@{}l@{}}{\reHeader{Model Group}\hdrstrut}\end{tabular} &
\begin{tabular}[c]{@{}l@{}}{\reHeader{Variant}\hdrstrut}\end{tabular} &
\multicolumn{2}{c}{\grouphead{0-shot mean}{6.5em}{EN}{ZH}} &
\multicolumn{2}{c}{\grouphead{5-shot mean}{6.5em}{EN}{ZH}} &
\multicolumn{3}{c}{\groupheadThree{Across-shot mean}{10em}{EN}{ZH}{Overall}} \\
\reHeadRule
\multirow{6}{*}{\textbf{Qwen2.5-7B}} & FP16 & 0.5032 & 0.5249 & 0.5032 & 0.5249 & 0.5032 & 0.5249 & \anchorval{0.5141} \\
 & BnB-4bit & 0.5025 & 0.4826 & 0.5025 & 0.4826 & 0.5025 & 0.4826 & 0.4926 \\
 & AWQ & 0.4959 & 0.5184 & 0.4976 & 0.5184 & 0.4967 & 0.5184 & 0.5076 \\
 & GPTQ-INT4 & 0.4895 & 0.5012 & 0.4895 & 0.5012 & 0.4895 & 0.5012 & 0.4954 \\
 & 2:4-Sparse & 0.3391 & 0.3139 & 0.3391 & 0.3139 & 0.3391 & 0.3139 & \keycell{0.3265} \\
 & Unstructured-Sparse & 0.4125 & 0.4135 & 0.4125 & 0.4135 & 0.4125 & 0.4135 & \keycell{0.4130} \\
\reMidRule
\multirow{6}{*}{\textbf{Llama3.1-8B}} & FP16 & 0.4774 & 0.3947 & 0.4774 & 0.3949 & 0.4774 & 0.3948 & \anchorval{0.4361} \\
 & BnB-4bit & 0.4325 & 0.3559 & 0.4325 & 0.3559 & 0.4325 & 0.3559 & 0.3942 \\
 & AWQ & 0.4624 & 0.3911 & 0.4614 & 0.3911 & 0.4619 & 0.3911 & 0.4265 \\
 & GPTQ-INT4 & 0.4141 & 0.3361 & 0.4136 & 0.3361 & 0.4139 & 0.3361 & 0.3750 \\
 & 2:4-Sparse & 0.2705 & 0.2686 & 0.2705 & 0.2686 & 0.2705 & 0.2686 & \keycell{0.2696} \\
 & Unstructured-Sparse & 0.3689 & 0.3056 & 0.3689 & 0.3056 & 0.3689 & 0.3056 & 0.3373 \\
\reMidRule
\multirow{7}{*}{\textbf{Qwen2.5-14B}} & FP16 & 0.5774 & 0.6051 & 0.5774 & 0.6051 & 0.5774 & \anchorval{0.6051} & \anchorval{0.5913} \\
 & BnB-4bit & 0.5623 & 0.5906 & 0.5623 & 0.5906 & 0.5623 & 0.5906 & \trendcell{0.5764} \\
 & AWQ & 0.5561 & 0.5780 & 0.5563 & 0.5775 & 0.5562 & 0.5777 & \trendcell{0.5670} \\
 & GPTQ-INT4 & 0.5641 & 0.5913 & 0.5634 & 0.5913 & 0.5637 & 0.5913 & \trendcell{0.5775} \\
 & 2:4-Sparse & 0.3590 & \trendcell{0.2985} & 0.3590 & \trendcell{0.2985} & \trendcell{0.3590} & \keycell{0.2985} & \keycell{0.3287} \\
 & Unstructured-Sparse & 0.4666 & 0.4754 & 0.4666 & 0.4754 & 0.4666 & 0.4754 & \keycell{0.4710} \\
 & Trim & 0.508 & 0.646 & 0.574 & 0.645 & 0.541 & 0.646 & 0.593 \\
\reMidRule
\multirow{5}{*}{\textbf{DeepSeek-V2-Lite-16B}} & FP16 & 0.4419 & 0.5054 & 0.4419 & 0.5054 & 0.4419 & 0.5054 & 0.4736 \\
 & BnB-4bit & 0.4275 & 0.4858 & 0.4275 & 0.4860 & 0.4275 & 0.4859 & 0.4567 \\
 & AWQ & 0.4333 & 0.4561 & 0.4311 & 0.4574 & 0.4322 & 0.4567 & 0.4445 \\
 & 2:4-Sparse & 0.3379 & 0.3236 & 0.3379 & 0.3234 & 0.3379 & 0.3235 & 0.3307 \\
 & Unstructured-Sparse & 0.3455 & 0.3481 & 0.3458 & 0.3484 & 0.3456 & 0.3483 & \trendcell{0.3469} \\
\reMidRule
\multirow{6}{*}{\textbf{Qwen2.5-32B}} & FP16 & 0.5867 & 0.6167 & 0.5867 & 0.6167 & 0.5867 & \anchorval{0.6167} & \anchorval{0.6018} \\
 & BnB-4bit & 0.5661 & 0.6048 & 0.5661 & 0.6048 & 0.5661 & 0.6048 & \trendcell{0.5854} \\
 & AWQ & 0.5704 & 0.5960 & 0.5689 & 0.5958 & 0.5696 & 0.5959 & \trendcell{0.5827} \\
 & GPTQ-INT4 & 0.5501 & 0.6001 & 0.5501 & 0.6001 & 0.5501 & 0.6001 & \trendcell{0.5751} \\
 & 2:4-Sparse & 0.4199 & \trendcell{0.3901} & 0.4199 & \trendcell{0.3901} & 0.4199 & \keycell{0.3901} & \keycell{0.4050} \\
 & Unstructured-Sparse & 0.5006 & 0.5082 & 0.5006 & 0.5082 & 0.5006 & 0.5082 & \keycell{0.5044} \\
\reMidRule
\multirow{7}{*}{\textbf{Llama-70B}} & FP16 & 0.6312 & 0.5302 & 0.6312 & 0.5302 & \anchorval{0.6312} & \anchorval{0.5302} & \anchorval{0.5807} \\
 & BnB-4bit & 0.5723 & 0.4617 & 0.5723 & 0.4617 & 0.5723 & 0.4617 & \keycell{0.5170} \\
 & AWQ & 0.6105 & 0.5216 & 0.6108 & 0.5216 & 0.6106 & 0.5216 & \trendcell{0.5661} \\
 & GPTQ-INT4 & 0.6069 & 0.5084 & 0.6070 & 0.5084 & 0.6069 & 0.5084 & \trendcell{0.5577} \\
 & GPTQ-INT8 & 0.541 & 0.718 & 0.473 & 0.590 & 0.507 & 0.654 & 0.580 \\
 & 2:4-Sparse & \trendcell{0.4289} & \trendcell{0.3267} & \trendcell{0.4290} & \trendcell{0.3267} & \keycell{0.4289} & \trendcell{0.3267} & \keycell{0.3778} \\
 & Unstructured-Sparse & 0.5504 & 0.4314 & 0.5504 & 0.4314 & 0.5504 & 0.4314 & \keycell{0.4909} \\
\reBottomRule
\end{tabular}
\end{table*}

\subsection{Social cognition and Affective understanding}
\label{sec:app_social}

This subsection follows the same hierarchy used in the social dimension of the paper: benchmark-level evidence from EQ-Bench, ability-level ToMBench profiles, and progressively finer ToMBench task and subability summaries.

\paragraph{EQ-Bench.}
Table~\ref{tab:appendix_eqbench} reports EQ-Bench benchmark score and parseable-answer count for the social cognition and affective understanding dimension. EQ-Bench asks models to infer relative emotional intensities in dialogue contexts, so the benchmark score reflects affective judgment quality, while the parseable count indicates whether the model stays within the expected emotion-label output format. The table is therefore useful for separating affective understanding from formatting or output-space drift.

\textit{Quantization.}
Quantized variants usually remain close to their FP16 parents on EQ-Bench score. Qwen2.5-7B even improves slightly under BnB-4bit, moving from 72.86 to 74.57, while AWQ and GPTQ-INT4 remain nearby at 70.76 and 70.07. Qwen2.5-14B shows modest degradation under AWQ and GPTQ-INT4, from 78.30 to 74.69 and 74.80, while BnB-4bit is closer at 77.59. Qwen2.5-32B is very stable, with BnB-4bit and AWQ slightly above the FP16 score of 79.73 at 80.55 and 80.29. Llama-70B also remains close to its FP16 score of 82.32 under 4-bit, AWQ, GPTQ-INT4, and GPTQ-INT8, all staying between 80.86 and 81.91. These results show that affective judgment scores are often preserved under quantization.

\textit{Parseability under quantization.}
Parseability is also mostly stable under quantization, but it is still worth reporting separately. Most quantized Qwen and Llama variants remain near 169--171 parseable answers, indicating that they continue to produce outputs in the expected emotion-rating format. The main quantization-side exception is DeepSeek-V2-Lite-16B, where AWQ reduces the benchmark score from 61.67 to 54.74 and the parseable count from 169.7 to 168.4. This drop is much smaller than the sparse failures, but it shows why parseability should remain visible even when a benchmark score is available.

\textit{Sparse degradation.}
Sparse variants degrade EQ-Bench more sharply than quantized variants. Under 2:4 sparsity, Qwen2.5-7B drops from 72.86 to 57.22, Qwen2.5-14B drops from 78.30 to 59.02, and Llama3.1-8B drops from 65.98 to 37.60. Larger models are more robust but still degrade: Qwen2.5-32B falls from 79.73 to 71.11, and Llama-70B falls from 82.32 to 67.15. This pattern supports the main text's social-dimension reading that sparse compression weakens socially situated affective judgment more strongly than post-training quantization.

\textit{Score--parseability separation.}
The DeepSeek-V2-Lite-16B sparse rows illustrate the strongest separation between affective score and parseability. Under 2:4 sparsity, the score becomes negative (-21.23) while 146.3 answers remain parseable. Since EQ-Bench is calibrated so that zero corresponds to random responding, a negative score indicates worse-than-random alignment with the reference emotional-intensity profile rather than merely a formatting problem. Under unstructured sparsity, the score is unavailable as a failed run, while the parseable count drops to 108.2. This means the run still produced some emotional outputs, but they cannot align with the benchmark. Together, these two rows show that compression can affect both the semantic affective judgment signal and the ability to stay within the required output format.

\textit{Sparsity structure.}
Unstructured sparsity is usually less damaging than 2:4 sparsity for EQ-Bench score, but it does not fully restore the FP16 baseline. Qwen2.5-7B improves from 57.22 under 2:4 sparsity to 65.11 under unstructured sparsity, Qwen2.5-14B improves from 59.02 to 76.94, Qwen2.5-32B improves from 71.11 to 77.07, and Llama-70B improves from 67.15 to 76.91. Llama3.1-8B also improves from 37.60 to 45.25, though it remains far below its FP16 score of 65.98. These results mirror the earlier dimensions: unstructured sparsity often retains more behavior than 2:4 sparsity, but sparse compression still causes visible ability-level degradation.

\textit{Social-affective reading.}
Together, the EQ-Bench results provide auxiliary evidence for the social cognition and affective understanding dimension. Quantization usually preserves both affective score and parseability, while sparsity can reduce the score, reduce parseable outputs, or both. The DeepSeek sparse rows are especially diagnostic: one sparse variant produces a negative score despite many parseable answers, and the other produces no valid score despite a nontrivial parseable count. This supports the main text's claim that compression can weaken not only emotional judgment itself, but also the model's ability to express that judgment in the benchmark's expected emotion-label space.

\begin{table*}[t]
\centering
\apptablestyle
\ifdefined\ACEHeaderBg\else\definecolor{ACEHeaderBg}{HTML}{F1F3F5}\fi
\ifdefined\ACEHeaderText\else\definecolor{ACEHeaderText}{HTML}{1F2D3D}\fi
\ifdefined\ACESocialHi\else\definecolor{ACESocialHi}{HTML}{FFF3E6}\fi
\ifdefined\ACESocialBorder\else\definecolor{ACESocialBorder}{HTML}{C87522}\fi
\ifdefined\ACERuleDark\else\definecolor{ACERuleDark}{HTML}{2E3640}\fi
\providecommand{\trendcell}[1]{\cellcolor{ACESocialHi}#1}
\providecommand{\keycell}[1]{\begingroup\setlength{\fboxsep}{1.2pt}\fcolorbox{ACESocialBorder}{ACESocialHi}{#1}\endgroup}
\providecommand{\anchorval}[1]{\underline{#1}}
\newcommand{\eqHeader}[1]{\textcolor{ACEHeaderText}{\textbf{#1}}}
\newcommand{\eqTopRule}{\arrayrulecolor{ACERuleDark}\toprule[0.55pt]\arrayrulecolor{ACERuleDark}}
\newcommand{\eqHeadRule}{\arrayrulecolor{ACERuleDark}\midrule[0.42pt]\arrayrulecolor{black}}
\newcommand{\eqMidRule}{\arrayrulecolor{ACERuleDark}\midrule[0.24pt]\arrayrulecolor{black}}
\newcommand{\eqBottomRule}{\arrayrulecolor{ACERuleDark}\bottomrule[0.55pt]\arrayrulecolor{black}}
\newcommand{\fcell}[1]{\textcolor{gray!55}{\texttt{#1}}}
\caption{EQ-Bench results for the social cognition and affective understanding dimension. Benchmark score measures affective judgment quality, while parseable-answer count tracks whether compressed models remain within the expected emotion-label output format.}
\label{tab:appendix_eqbench}
\begin{tabular}{llcc}
\eqTopRule
\rowcolor{ACEHeaderBg}
\eqHeader{Model Group} & \eqHeader{Variant} & \eqHeader{Benchmark Score} & \eqHeader{Parseable Answers} \\
\eqHeadRule
\multirow{6}{*}{\textbf{Qwen2.5-7B}} & FP16 & \anchorval{72.86} & 169.0 \\
 & BnB-4bit & \keycell{74.57} & 169.0 \\
 & AWQ & \trendcell{70.76} & 170.0 \\
 & GPTQ-INT4 & \trendcell{70.07} & 170.0 \\
 & 2:4-Sparse & \keycell{57.22} & 167.0 \\
 & Unstructured-Sparse & \trendcell{65.11} & 170.5 \\
\eqMidRule
\multirow{6}{*}{\textbf{Llama3.1-8B}} & FP16 & \anchorval{65.98} & 171.0 \\
 & BnB-4bit & 67.06 & 171.0 \\
 & AWQ & 59.97 & 171.0 \\
 & GPTQ-INT4 & 59.28 & 171.0 \\
 & 2:4-Sparse & \keycell{37.60} & 167.9 \\
 & Unstructured-Sparse & \trendcell{45.25} & 169.0 \\
\eqMidRule
\multirow{7}{*}{\textbf{Qwen2.5-14B}} & FP16 & \anchorval{78.30} & 170.0 \\
 & BnB-4bit & \trendcell{77.59} & 170.0 \\
 & AWQ & \trendcell{74.69} & 170.0 \\
 & GPTQ-INT4 & \trendcell{74.80} & 168.5 \\
 & 2:4-Sparse & \keycell{59.02} & 170.6 \\
 & Unstructured-Sparse & \trendcell{76.94} & 169.9 \\
 & Trim & 78.31 & 169.7 \\
\eqMidRule
\multirow{5}{*}{\textbf{DeepSeek-V2-Lite-16B}} & FP16 & \anchorval{61.67} & \anchorval{169.7} \\
 & BnB-4bit & 59.27 & 170.0 \\
 & AWQ & 54.74 & 168.4 \\
 & 2:4-Sparse & \keycell{-21.23} & \keycell{146.3} \\
 & Unstructured-Sparse & \fcell{FAILED} & \keycell{108.2} \\
\eqMidRule
\multirow{6}{*}{\textbf{Qwen2.5-32B}} & FP16 & \anchorval{79.73} & 170.0 \\
 & BnB-4bit & \keycell{80.55} & 169.7 \\
 & AWQ & \keycell{80.29} & 171.0 \\
 & GPTQ-INT4 & \trendcell{77.98} & 169.0 \\
 & 2:4-Sparse & \keycell{71.11} & 169.0 \\
 & Unstructured-Sparse & \trendcell{77.07} & 168.0 \\
\eqMidRule
\multirow{7}{*}{\textbf{Llama-70B}} & FP16 & \anchorval{82.32} & 171.0 \\
 & BnB-4bit & \trendcell{81.89} & 171.0 \\
 & AWQ & \trendcell{81.16} & 171.0 \\
 & GPTQ-INT4 & \trendcell{80.86} & 171.0 \\
 & GPTQ-INT8 & \trendcell{81.91} & 171.0 \\
 & 2:4-Sparse & \keycell{67.15} & 170.0 \\
 & Unstructured-Sparse & \trendcell{76.91} & 171.0 \\
\eqBottomRule
\end{tabular}
\end{table*}

\paragraph{ToMBench ability view.}

Table~\ref{tab:appendix_tombench_ability} reports the six official ToMBench ability groups after averaging EN and ZH scores within each model variant. This table gives the broad social-cognitive profile behind the main text, while the task and subability tables below expose where the losses concentrate.

\textit{Quantization.}
At the ability level, quantized variants generally remain close to their FP16 parents. For Qwen2.5-14B, Belief remains high under BnB-4bit and AWQ (0.8667 and 0.8650 versus 0.8650 for FP16), while Emotion stays close across quantized variants (0.7094, 0.7219, and 0.7302 versus 0.7302). Qwen2.5-32B shows a similar pattern: BnB-4bit, AWQ, and GPTQ-INT4 preserve most of the FP16 ability profile, with Desire even slightly above FP16 under AWQ and GPTQ-INT4 (0.6143 and 0.6357 versus 0.6036). Llama-70B is also stable across most ability groups, with AWQ and GPTQ-INT4 close to or slightly above FP16 on Belief and Non-literal Communication. These results match the main text's observation that quantization usually preserves social-cognitive behavior better than sparsity.

\textit{Sparse degradation.}
Sparse variants show larger and more uneven losses. For Llama3.1-8B, 2:4 sparsity reduces Emotion from 0.6719 to 0.3313 and Intention from 0.6854 to 0.3250. Qwen2.5-14B drops from 0.8650 to 0.6133 on Belief and from 0.7493 to 0.4987 on Intention. Larger models degrade less in absolute terms but still show clear losses: Qwen2.5-32B falls from 0.8558 to 0.5833 on Belief, and Llama-70B falls from 0.8541 to 0.5167. Unstructured sparsity is usually less damaging than 2:4 sparsity, but it does not fully restore the FP16 ability profile. This ability-level view shows that social-cognitive degradation is not uniform across abilities.
\begin{table*}[t]
\centering
\apptablestyle
\ifdefined\ACEHeaderBg\else\definecolor{ACEHeaderBg}{HTML}{F1F3F5}\fi
\ifdefined\ACEHeaderText\else\definecolor{ACEHeaderText}{HTML}{1F2D3D}\fi
\ifdefined\ACESocialHi\else\definecolor{ACESocialHi}{HTML}{FFF3E6}\fi
\ifdefined\ACESocialBorder\else\definecolor{ACESocialBorder}{HTML}{C87522}\fi
\ifdefined\ACERuleDark\else\definecolor{ACERuleDark}{HTML}{2E3640}\fi
\providecommand{\trendcell}[1]{\cellcolor{ACESocialHi}#1}
\providecommand{\keycell}[1]{\begingroup\setlength{\fboxsep}{1.2pt}\fcolorbox{ACESocialBorder}{ACESocialHi}{#1}\endgroup}
\providecommand{\anchorval}[1]{\underline{#1}}
\newcommand{\abHeader}[1]{\textcolor{ACEHeaderText}{\textbf{#1}}}
\newcommand{\abTopRule}{\arrayrulecolor{ACERuleDark}\toprule[0.55pt]\arrayrulecolor{ACERuleDark}}
\newcommand{\abHeadRule}{\arrayrulecolor{ACERuleDark}\midrule[0.42pt]\arrayrulecolor{black}}
\newcommand{\abMidRule}{\arrayrulecolor{ACERuleDark}\midrule[0.24pt]\arrayrulecolor{black}}
\newcommand{\abBottomRule}{\arrayrulecolor{ACERuleDark}\bottomrule[0.55pt]\arrayrulecolor{black}}
\caption{ToMBench ability-level scores for the social cognition and affective understanding dimension. EN/ZH scores are averaged within each official ability group to show broad social-cognitive profiles under compression.}
\label{tab:appendix_tombench_ability}
\resizebox{\textwidth}{!}{%
\begin{tabular}{llcccccc}
\abTopRule
\rowcolor{ACEHeaderBg}
\abHeader{Model Group} & \abHeader{Variant} & \abHeader{Belief} & \abHeader{Desire} & \abHeader{Emotion} & \abHeader{Intention} & \abHeader{Knowledge} & \abHeader{Non-literal Comm.} \\
\abHeadRule
\multirow{6}{*}{\textbf{Qwen2.5-7B}} & FP16 & 0.7642 & 0.5286 & 0.6666 & 0.6631 & 0.4914 & 0.6996 \\
 & BnB-4bit & 0.7391 & 0.4750 & 0.6614 & 0.6736 & 0.4707 & 0.7047 \\
 & AWQ & 0.7500 & 0.5535 & 0.6625 & 0.6397 & 0.4672 & 0.6862 \\
 & GPTQ-INT4 & 0.7550 & 0.5465 & 0.6844 & 0.6841 & 0.4689 & 0.6939 \\
 & 2:4-Sparse & 0.4983 & 0.4643 & 0.4896 & 0.4948 & 0.3759 & 0.6240 \\
 & Unstructured-Sparse & 0.6100 & 0.5143 & 0.6156 & 0.6175 & 0.4362 & 0.6675 \\
\abMidRule
\multirow{6}{*}{\textbf{Llama3.1-8B}} & FP16 & 0.5700 & 0.5393 & \anchorval{0.6719} & \anchorval{0.6854} & 0.4431 & 0.7011 \\
 & BnB-4bit & 0.5834 & 0.5286 & 0.6521 & 0.6880 & 0.4276 & 0.6835 \\
 & AWQ & 0.6000 & 0.5071 & 0.6511 & 0.6749 & 0.4242 & 0.6737 \\
 & GPTQ-INT4 & 0.6209 & 0.4821 & 0.6479 & 0.6632 & 0.4000 & 0.6821 \\
 & 2:4-Sparse & 0.3883 & 0.2928 & \keycell{0.3313} & \keycell{0.3250} & 0.2638 & 0.4286 \\
 & Unstructured-Sparse & 0.4941 & 0.4642 & \trendcell{0.5615} & \trendcell{0.5392} & 0.3241 & 0.5568 \\
\abMidRule
\multirow{7}{*}{\textbf{Qwen2.5-14B}} & FP16 & \anchorval{0.8650} & 0.5857 & \anchorval{0.7302} & \anchorval{0.7493} & 0.4897 & 0.7446 \\
 & BnB-4bit & \trendcell{0.8667} & 0.5821 & 0.7094 & 0.7310 & 0.4776 & 0.7233 \\
 & AWQ & \trendcell{0.8650} & 0.5821 & 0.7219 & 0.7114 & 0.4569 & 0.7130 \\
 & GPTQ-INT4 & 0.8050 & 0.5928 & \trendcell{0.7302} & 0.7310 & 0.4810 & 0.7187 \\
 & 2:4-Sparse & \keycell{0.6133} & 0.4786 & 0.6031 & \keycell{0.4987} & 0.4173 & 0.6541 \\
 & Unstructured-Sparse & \trendcell{0.7317} & 0.5715 & 0.6802 & \trendcell{0.6502} & 0.4327 & 0.6701 \\
 & Trim & 0.8667 & 0.5786 & 0.7229 & 0.7428 & 0.4949 & 0.7425 \\
\abMidRule
\multirow{5}{*}{\textbf{DeepSeek-V2-Lite-16B}} & FP16 & 0.5084 & 0.4178 & 0.6094 & 0.6097 & 0.2431 & 0.5735 \\
 & BnB-4bit & 0.4733 & 0.3785 & 0.5615 & 0.5783 & 0.2310 & 0.5357 \\
 & AWQ & 0.4700 & 0.3786 & 0.5573 & 0.5796 & 0.2275 & 0.5548 \\
 & 2:4-Sparse & 0.3866 & 0.1893 & 0.3198 & 0.2585 & 0.2138 & 0.4488 \\
 & Unstructured-Sparse & 0.4117 & 0.3215 & 0.4646 & 0.3825 & 0.2017 & 0.4949 \\
\abMidRule
\multirow{6}{*}{\textbf{Qwen2.5-32B}} & FP16 & \anchorval{0.8558} & \anchorval{0.6036} & 0.7562 & 0.8016 & 0.4758 & 0.7648 \\
 & BnB-4bit & 0.8433 & 0.6107 & 0.7469 & 0.7872 & 0.4776 & 0.7606 \\
 & AWQ & 0.8300 & \trendcell{0.6143} & 0.7604 & 0.7872 & 0.4741 & 0.7751 \\
 & GPTQ-INT4 & 0.8508 & \trendcell{0.6357} & 0.7593 & 0.7912 & 0.4586 & 0.7606 \\
 & 2:4-Sparse & \keycell{0.5833} & 0.5143 & 0.6583 & 0.6488 & 0.4586 & 0.6970 \\
 & Unstructured-Sparse & \trendcell{0.6950} & 0.5535 & 0.7333 & 0.7480 & 0.4569 & 0.7187 \\
\abMidRule
\multirow{7}{*}{\textbf{Llama-70B}} & FP16 & \anchorval{0.8541} & 0.6036 & 0.7583 & 0.8172 & 0.5069 & \anchorval{0.7642} \\
 & BnB-4bit & 0.8400 & 0.6285 & 0.7531 & 0.8055 & 0.5000 & 0.7658 \\
 & AWQ & 0.8458 & 0.6107 & 0.7605 & 0.8185 & 0.5034 & \trendcell{0.7802} \\
 & GPTQ-INT4 & 0.8675 & 0.6143 & 0.7552 & 0.8094 & 0.4776 & \trendcell{0.7891} \\
 & GPTQ-INT8 & 0.8550 & 0.6108 & 0.7532 & 0.8173 & 0.5017 & 0.7880 \\
 & 2:4-Sparse & \keycell{0.5167} & 0.6036 & 0.6532 & 0.6292 & 0.4224 & 0.7043 \\
 & Unstructured-Sparse & \trendcell{0.7000} & 0.5821 & 0.7479 & 0.7833 & 0.4982 & 0.7751 \\
\abBottomRule
\end{tabular}
}
\end{table*}

\paragraph{ToMBench task view.}
Table~\ref{tab:appendix_tombench_task_summary} summarizes eight ToMBench task families as method-level retention ratios after averaging EN/ZH and then normalizing to each FP16 parent.

\textit{Task-level pattern.}
Quantization is consistently close to FP16 across task families. BnB, AWQ, and GPTQ mostly remain near 0.95--1.01, with only moderate variation across tasks. In contrast, 2:4 sparsity produces task-specific losses. The weakest task-level retentions appear on Hinting Task (0.570), False Belief Task (0.682), Ambiguous Story (0.675), and Unexpected Outcome (0.721). These tasks require inferring hidden mental states, resolving implied meanings, or coordinating belief-state information across context. Unstructured sparsity partially recovers these task families, reaching 0.802 on Hinting Task, 0.825 on False Belief Task, 0.862 on Ambiguous Story, and 0.915 on Unexpected Outcome, but it remains below the quantized variants. This supports the main text's claim that social-cognitive degradation concentrates in ambiguity-heavy and mental-state-centered reasoning rather than appearing as a uniform ToM decline.

\textit{Relatively robust tasks.}
Some task families are more stable under sparsity. Scalar Implicature retains 0.840 under 2:4 sparsity and 0.911 under unstructured sparsity, while Faux-pas Recognition remains at 0.849 and 0.905. Strange Story also shows stronger retention than the most fragile tasks, with 0.812 under 2:4 sparsity and 0.920 under unstructured sparsity. These contrasts are useful because they show that the social dimension is not simply harder overall; compression affects particular kinds of social reasoning more strongly than others.
\begin{table*}[t]
\centering
\apptablestyle
\ifdefined\ACEHeaderBg\else\definecolor{ACEHeaderBg}{HTML}{F1F3F5}\fi
\ifdefined\ACEHeaderText\else\definecolor{ACEHeaderText}{HTML}{1F2D3D}\fi
\ifdefined\ACESocialHi\else\definecolor{ACESocialHi}{HTML}{FFF3E6}\fi
\ifdefined\ACESocialBorder\else\definecolor{ACESocialBorder}{HTML}{C87522}\fi
\ifdefined\ACERuleDark\else\definecolor{ACERuleDark}{HTML}{2E3640}\fi
\providecommand{\trendcell}[1]{\cellcolor{ACESocialHi}#1}
\providecommand{\keycell}[1]{\begingroup\setlength{\fboxsep}{1.2pt}\fcolorbox{ACESocialBorder}{ACESocialHi}{#1}\endgroup}
\newcommand{\tkHeader}[1]{\textcolor{ACEHeaderText}{\textbf{#1}}}
\newcommand{\tkTopRule}{\arrayrulecolor{ACERuleDark}\toprule[0.55pt]\arrayrulecolor{ACERuleDark}}
\newcommand{\tkHeadRule}{\arrayrulecolor{ACERuleDark}\midrule[0.42pt]\arrayrulecolor{black}}
\newcommand{\tkBottomRule}{\arrayrulecolor{ACERuleDark}\bottomrule[0.55pt]\arrayrulecolor{black}}
\caption{ToMBench task-level retention ratios for the social cognition and affective understanding dimension. Values are FP16-relative retention ratios after averaging EN/ZH within each task and base model, then averaging by compression method.}
\label{tab:appendix_tombench_task_summary}
\resizebox{\textwidth}{!}{%
\begin{tabular}{>{\raggedright\arraybackslash}p{5.2cm}ccccc}
\tkTopRule
\rowcolor{ACEHeaderBg}
\tkHeader{Task (EN/ZH mean)} & \tkHeader{BnB} & \tkHeader{AWQ} & \tkHeader{GPTQ} & \tkHeader{2:4} & \tkHeader{Unstructured} \\
\tkHeadRule
Faux-pas Recognition & 0.974 & 0.980 & 0.986 & \trendcell{0.849} & 0.905 \\
Unexpected Outcome  & 0.973 & 0.987 & 1.008 & \keycell{0.721} & \trendcell{0.915} \\
Persuasion Story & 0.969 & 0.977 & 1.006 & \keycell{0.783} & \trendcell{0.925} \\
False Belief Task & 0.982 & 0.986 & 1.004 & \keycell{0.682} & \trendcell{0.825} \\
Ambiguous Story & 0.989 & 0.983 & 0.993 & \keycell{0.675} & \trendcell{0.862} \\
Strange Story & 0.991 & 0.987 & 0.999 & \trendcell{0.812} & 0.920 \\
Hinting Task  & 0.978 & 0.941 & 0.985 & \keycell{0.570} & \trendcell{0.802} \\
Scalar Implicature  & 0.947 & 0.927 & 0.953 & \trendcell{0.840} & 0.911 \\
\tkBottomRule
\end{tabular}
}
\end{table*}

\paragraph{ToMBench subability view.}
Table~\ref{tab:appendix_tombench_subabilities} reports ToMBench subability-level retention ratios.

\textit{Compression-sensitive subabilities.}
The lowest 2:4 retentions concentrate in belief, intention, emotion, and non-literal communication. Location false beliefs drops to 0.612, second-order beliefs to 0.608, discrepant emotions to 0.616, intentions explanations to 0.647, multiple desires to 0.663, and irony/sarcasm to 0.615. These are the kinds of cases where the model must infer unobserved beliefs, explain intentions, recognize emotional mismatch, or interpret non-literal meaning. The pattern aligns with the main figure's subability panel: sparse compression does not erase all ToM behavior, but it disproportionately weakens hidden-state and ambiguity-centered reasoning.

\textit{Partial recovery under unstructured sparsity.}
Unstructured sparsity usually improves over 2:4 sparsity on these fragile subabilities, but recovery is incomplete. Location false beliefs rises from 0.612 to 0.768, second-order beliefs from 0.608 to 0.724, discrepant emotions from 0.616 to 0.851, intentions explanations from 0.647 to 0.843, and irony/sarcasm from 0.615 to 0.856. This reinforces the cross-dimensional pattern seen in the procedural and knowledge sections: unstructured sparsity tends to preserve more behavior than fixed 2:4 sparsity, but it still leaves clear ability-level deficits.

\textit{Robust and above-parent cases.}
Not all subabilities are fragile. Mixed emotions is nearly preserved under 2:4 sparsity (0.991) and reaches 1.006 under unstructured sparsity. White lies also remains comparatively stable, with 0.940 under 2:4 sparsity and 0.988 under unstructured sparsity. Some knowledge-link subabilities even exceed FP16-relative retention, such as knowledge-pretend play links under 2:4 and unstructured sparsity (1.622 and 1.473). These above-parent ratios should not be interpreted as a uniform improvement in social cognition; rather, they show that the compression effect is highly localized and can redistribute performance across subability groups.

\textit{Social-cognitive reading.}
Together, the ToMBench ability, task, and subability tables support the social-dimension interpretation in the main text. Quantization generally preserves broad social-cognitive profiles, while sparse compression produces localized degradation. The most sensitive regions are not generic social tasks, but hidden-state, belief-, intention-, emotion-mismatch, and non-literal-communication subabilities. This is why ACE-Compression keeps the ToMBench hierarchy instead of reducing social cognition to a single aggregate ToM score.
\begin{table*}[t]
\centering
\apptablestyle
\setlength{\tabcolsep}{3pt}
\ifdefined\ACESubgroupBg\else\definecolor{ACESubgroupBg}{HTML}{F3F4F6}\fi
\ifdefined\ACEHeaderBg\else\definecolor{ACEHeaderBg}{HTML}{F1F3F5}\fi
\ifdefined\ACEHeaderText\else\definecolor{ACEHeaderText}{HTML}{1F2D3D}\fi
\ifdefined\ACESocialHi\else\definecolor{ACESocialHi}{HTML}{FFF3E6}\fi
\ifdefined\ACESocialBorder\else\definecolor{ACESocialBorder}{HTML}{C87522}\fi
\ifdefined\ACERuleDark\else\definecolor{ACERuleDark}{HTML}{2E3640}\fi
\providecommand{\trendcell}[1]{\cellcolor{ACESocialHi}#1}
\providecommand{\keycell}[1]{\begingroup\setlength{\fboxsep}{1.2pt}\fcolorbox{ACESocialBorder}{ACESocialHi}{#1}\endgroup}
\newcommand{\sbHeader}[1]{\textcolor{ACEHeaderText}{\textbf{#1}}}
\newcommand\WideTop{\noalign{\vskip\abovetopsep\hrule\vskip\belowrulesep}}
\newcommand\WideMid{\noalign{\vskip\aboverulesep\hrule height 0.05em\vskip\belowrulesep}}
\newcommand\WideBot{\noalign{\vskip\aboverulesep\hrule\vskip\belowbottomsep}}
\caption{ToMBench subability-level retention ratios for the social cognition and affective understanding dimension. Values are FP16-relative retention ratios \(R\), so the FP16 parent is omitted and corresponds to \(R=1\).}
\label{tab:appendix_tombench_subabilities}
\resizebox{\textwidth}{!}{%
\begin{tabular}{>{\raggedright\arraybackslash}p{4.7cm}*{4}{>{\centering\arraybackslash}p{0.95cm}}>{\centering\arraybackslash}p{1.7cm}}
\WideTop
\sbHeader{Subability} & \sbHeader{BnB} & \sbHeader{AWQ} & \sbHeader{GPTQ} & \sbHeader{2:4} & \sbHeader{Unstructured} \\
\WideMid
\rowcolor{ACESubgroupBg}\multicolumn{6}{c}{\textit{Belief}} \\
Content false beliefs & 0.976 & 1.018 & 1.020 & 0.762 & 0.895 \\
Location false beliefs & 0.986 & 0.957 & 0.988 & \keycell{0.612} & \trendcell{0.768} \\
Identity false beliefs & 1.007 & 0.980 & 0.997 & 0.771 & 0.920 \\
Second-order beliefs & 0.963 & 0.975 & 0.981 & \keycell{0.608} & \trendcell{0.724} \\
Beliefs based action/emotions & 0.990 & 0.986 & 0.988 & 0.700 & 0.864 \\
Sequence false beliefs & 0.942 & 0.967 & 1.006 & 0.679 & 0.852 \\
\WideMid
\rowcolor{ACESubgroupBg}\multicolumn{6}{c}{\textit{Desire}} \\
Discrepant desires & 1.048 & 1.089 & 1.025 & 0.830 & 0.936 \\
Multiple desires & 0.965 & 0.985 & 0.982 & \keycell{0.663} & \trendcell{0.863} \\
Desires influence on emotions and actions & 0.969 & 0.977 & 1.006 & 0.783 & 0.925 \\
Desire-action contradiction & 0.983 & 0.973 & 0.996 & 0.828 & 0.941 \\
\WideMid
\rowcolor{ACESubgroupBg}\multicolumn{6}{c}{\textit{Emotion}} \\
Typical emotional reactions & 0.990 & 0.983 & 0.997 & 0.754 & 0.937 \\
Atypical emotional reactions & 0.981 & 1.017 & 1.029 & 0.721 & 0.957 \\
Discrepant emotions & 0.971 & 0.952 & 0.974 & \keycell{0.616} & \trendcell{0.851} \\
Mixed emotions & 0.975 & 0.953 & 1.006 & \trendcell{0.991} & \trendcell{1.006} \\
Hidden emotions & 0.980 & 0.970 & 0.985 & 0.735 & 0.890 \\
Moral emotions & 0.995 & 0.972 & 0.991 & 0.735 & 0.895 \\
Emotion regulation & 0.867 & 0.975 & 0.976 & 0.840 & 0.872 \\
\WideMid
\rowcolor{ACESubgroupBg}\multicolumn{6}{c}{\textit{Intention}} \\
Completion of failed actions & 0.968 & 1.051 & 0.985 & 0.774 & 0.870 \\
Discrepant intentions & 0.981 & 0.970 & 1.015 & 0.676 & 0.902 \\
Prediction of actions & 1.039 & 0.986 & 0.955 & 0.705 & 0.868 \\
Intentions explanations & 0.973 & 0.967 & 0.985 & \keycell{0.647} & \trendcell{0.843} \\
\WideMid
\rowcolor{ACESubgroupBg}\multicolumn{6}{c}{\textit{Knowledge}} \\
Knowledge-pretend play links & \keycell{1.009} & \keycell{1.157} & 1.030 & \keycell{1.622} & \keycell{1.473} \\
Percepts-knowledge links & \keycell{1.243} & \keycell{1.548} & 0.933 & 0.784 & 0.853 \\
Information-knowledge links & 0.947 & 0.927 & 0.953 & 0.840 & 0.911 \\
Knowledge-attention links & \keycell{1.126} & 1.049 & 0.964 & \keycell{1.087} & 0.927 \\
\WideMid
\rowcolor{ACESubgroupBg}\multicolumn{6}{c}{\textit{Non-Literal Communication}} \\
Irony/Sarcasm & 1.012 & 0.982 & 1.023 & \keycell{0.615} & \trendcell{0.856} \\
Egocentric lies & 0.993 & 0.998 & 1.025 & \trendcell{0.925} & \trendcell{0.942} \\
White lies & 1.054 & 1.082 & 1.015 & \trendcell{0.940} & \trendcell{0.988} \\
Involuntary lies & 0.982 & 0.909 & 1.007 & 0.824 & 0.944 \\
Humor & 1.028 & 1.041 & 1.017 & 0.839 & 0.926 \\
Faux pas & 0.974 & 0.980 & 0.986 & 0.849 & 0.905 \\
\WideBot
\end{tabular}
}
\end{table*}

\subsection{Persona Fidelity and Stylistic Consistency}
\label{sec:app_persona}

The persona appendix is organized to match the paper's two-layer interpretation: visible role-play quality first, then latent profile movement. The goal here is not to restate the discussion section, but to put the relevant raw or minimally processed outputs in one place.

\paragraph{CharacterEval.}
Table~\ref{tab:appendix_charactereval_full} reports the full official grouped CharacterEval results for the persona fidelity and stylistic consistency dimension. CharacterEval provides the visible role-play side of the persona dimension, with grouped metrics for character consistency, conversational ability, and role-playing attractiveness. The appendix shows the released grouped breakdown, including KE, KH, and Emp., so that visible persona behavior can be inspected beyond the selected diagnostic view in the main paper.

\textit{Quantization.}
Quantized variants usually remain close to their FP16 parents across the three CharacterEval groups. Qwen2.5-14B, for example, keeps character consistency near its FP16 average of 2.966 under BnB-4bit, AWQ, and GPTQ-INT4 (2.907, 2.867, and 2.904), while conversational ability remains similarly close to the FP16 average of 3.716. Its role-playing attractiveness also changes only modestly, from 3.173 for FP16 to 3.091, 3.077, and 3.107 under the same quantized variants. Llama-70B is similarly stable under GPTQ-INT4, reaching 2.825 on character consistency, 3.694 on conversational ability, and 3.009 on role-playing attractiveness, compared with FP16 values of 2.841, 3.704, and 3.059. These results support the main text's observation that quantization often preserves visible persona behavior better than sparse compression.

\textit{Metric-level redistribution.}
The quantized rows also show that preservation is not always uniform across CharacterEval metrics. Qwen2.5-32B under AWQ keeps character consistency almost unchanged on average (2.988 vs. 2.994 for FP16), but its role-playing attractiveness average drops from 3.217 to 2.765. This drop is driven by lower CS and ED scores, even though HL increases from 3.261 to 3.598. This redistribution matters for ACE-Compression: a compressed model may preserve some persona-adjacent signals while weakening others that make the role feel vivid or consistently enacted.

\textit{Sparse degradation.}
Sparse variants produce much stronger degradation, especially under 2:4 sparsity. Qwen2.5-7B drops from 2.927 to 2.363 on character consistency, from 3.694 to 2.770 on conversational ability, and from 3.155 to 2.380 on role-playing attractiveness. Llama3.1-8B shows an even larger collapse, falling from 2.720 to 1.832, from 3.466 to 2.248, and from 2.896 to 1.861 across the same three groups. Larger models degrade as well: Llama-70B falls from 2.841 to 2.204 on character consistency, from 3.704 to 2.833 on conversational ability, and from 3.059 to 2.293 on role-playing attractiveness. These results match the main text's claim that sparse compression visibly flattens persona enactment rather than merely reducing a generic score.

\textit{Sparsity structure.}
Unstructured sparsity is consistently less damaging than 2:4 sparsity, but it does not fully recover the FP16 persona profile. Qwen2.5-14B improves from 2.237 to 2.612 on character consistency, from 2.652 to 3.219 on conversational ability, and from 2.251 to 2.722 on role-playing attractiveness when moving from 2:4 sparsity to unstructured sparsity. Llama-70B shows the same direction, rising from 2.204 to 2.552, from 2.833 to 3.378, and from 2.293 to 2.692 across the three groups. This mirrors the pattern in earlier dimensions: unstructured sparsity preserves more behavior than fixed 2:4 pruning, but sparse compression still leaves visible ability-level deficits.

\textit{Persona-specific signals.}
The metric columns show where visible persona loss concentrates. Under 2:4 sparsity, expressive or enactment-related metrics such as ED, PB, and Cons. often drop sharply. Qwen2.5-7B ED falls from 2.547 to 1.728 and Cons. falls from 3.612 to 2.533. DeepSeek-V2-Lite-16B PB falls from 2.811 to 1.774 and Cons. falls from 3.513 to 2.121. Qwen2.5-32B ED falls from 2.649 to 1.685, while its conversational consistency drops from 3.526 to 2.550. These drops support the main-paper reading that compression first weakens how the role is enacted, expressed, and maintained across turns, rather than simply erasing all role-related knowledge at once.

\textit{Visible persona reading.}
Together, the full CharacterEval table provides the visible-persona counterpart to the later PsychoBench profile analysis. Quantization usually preserves character consistency, conversational ability, and role-playing attractiveness with small shifts, while sparse compression produces broad and metric-specific flattening. This justifies treating CharacterEval and PsychoBench as complementary persona probes: CharacterEval exposes outward role enactment and style consistency, whereas PsychoBench tracks latent psychometric profile movement.
\begin{table*}[t]
\centering
\apptablestyle
\ifdefined\ACEHeaderBg\else\definecolor{ACEHeaderBg}{HTML}{F1F3F5}\fi
\ifdefined\ACEHeaderText\else\definecolor{ACEHeaderText}{HTML}{1F2D3D}\fi
\ifdefined\ACEPersonaHi\else\definecolor{ACEPersonaHi}{HTML}{F8F2FA}\fi
\ifdefined\ACEPersonaBorder\else\definecolor{ACEPersonaBorder}{HTML}{7B5EA7}\fi
\ifdefined\ACEPersonaSoft\else\definecolor{ACEPersonaSoft}{HTML}{FAF6FD}\fi
\ifdefined\ACERuleDark\else\definecolor{ACERuleDark}{HTML}{2E3640}\fi
\ifdefined\ACERuleLight\else\definecolor{ACERuleLight}{HTML}{CAD3DD}\fi
\providecommand{\trendcell}[1]{\cellcolor{ACEPersonaHi}#1}
\providecommand{\keycell}[1]{\begingroup\setlength{\fboxsep}{1.2pt}\fcolorbox{ACEPersonaBorder}{ACEPersonaHi}{#1}\endgroup}
\providecommand{\anchorval}[1]{\underline{#1}}
\newcommand{\cfHeader}[1]{\textcolor{ACEHeaderText}{\textbf{#1}}}
\newcommand{\hdrstrut}{\rule{0pt}{2ex}}
\newcommand{\cfTopRule}{\arrayrulecolor{ACERuleDark}\toprule[0.55pt]\arrayrulecolor{ACERuleDark}}
\newcommand{\cfHeadRule}{\arrayrulecolor{ACERuleDark}\midrule[0.42pt]\arrayrulecolor{black}}
\newcommand{\cfMidRule}{\arrayrulecolor{ACERuleDark}\midrule[0.24pt]\arrayrulecolor{black}}
\newcommand{\cfBottomRule}{\arrayrulecolor{ACERuleDark}\bottomrule[0.55pt]\arrayrulecolor{black}}
\newcommand{\cfGroupSix}[8]{%
  \begin{tabular}[c]{@{}c@{}}
    \rule{0pt}{2.15ex}\textcolor{ACEHeaderText}{\textbf{#1}}\\[-0.55ex]
    \rule{#2}{0.4pt}\\[-0.15ex]
    \makebox[#2][c]{%
      \makebox[\dimexpr#2/6\relax][c]{\cfHeader{#3}\hdrstrut}%
      \makebox[\dimexpr#2/6\relax][c]{\cfHeader{#4}\hdrstrut}%
      \makebox[\dimexpr#2/6\relax][c]{\cfHeader{#5}\hdrstrut}%
      \makebox[\dimexpr#2/6\relax][c]{\cfHeader{#6}\hdrstrut}%
      \makebox[\dimexpr#2/6\relax][c]{\cfHeader{#7}\hdrstrut}%
      \makebox[\dimexpr#2/6\relax][c]{\cfHeader{#8}\hdrstrut}%
    }
  \end{tabular}%
}
\newcommand{\cfGroupFour}[6]{%
  \begin{tabular}[c]{@{}c@{}}
    \rule{0pt}{2.15ex}\textcolor{ACEHeaderText}{\textbf{#1}}\\[-0.55ex]
    \rule{#2}{0.4pt}\\[-0.15ex]
    \makebox[#2][c]{%
      \makebox[\dimexpr#2/4\relax][c]{\cfHeader{#3}\hdrstrut}%
      \makebox[\dimexpr#2/4\relax][c]{\cfHeader{#4}\hdrstrut}%
      \makebox[\dimexpr#2/4\relax][c]{\cfHeader{#5}\hdrstrut}%
      \makebox[\dimexpr#2/4\relax][c]{\cfHeader{#6}\hdrstrut}%
    }
  \end{tabular}%
}
\newcommand{\cfGroupFive}[7]{%
  \begin{tabular}[c]{@{}c@{}}
    \rule{0pt}{2.15ex}\textcolor{ACEHeaderText}{\textbf{#1}}\\[-0.55ex]
    \rule{#2}{0.4pt}\\[-0.15ex]
    \makebox[#2][c]{%
      \makebox[\dimexpr#2/5\relax][c]{\cfHeader{#3}\hdrstrut}%
      \makebox[\dimexpr#2/5\relax][c]{\cfHeader{#4}\hdrstrut}%
      \makebox[\dimexpr#2/5\relax][c]{\cfHeader{#5}\hdrstrut}%
      \makebox[\dimexpr#2/5\relax][c]{\cfHeader{#6}\hdrstrut}%
      \makebox[\dimexpr#2/5\relax][c]{\cfHeader{#7}\hdrstrut}%
    }
  \end{tabular}%
}
\caption{CharacterEval full official grouped results for the persona fidelity and stylistic consistency dimension. The appendix restores the released grouped breakdown, including KE, KH, and Emp., together with workbook-provided group averages.}
\label{tab:appendix_charactereval_full}
\resizebox{\textwidth}{!}{%
\begin{tabular}{llccccccccccccccc}
\cfTopRule
\rowcolor{ACEHeaderBg}
\begin{tabular}[c]{@{}l@{}}\cfHeader{Model Group}\end{tabular} &
\begin{tabular}[c]{@{}l@{}}\cfHeader{Variant}\end{tabular} &
\multicolumn{6}{c}{\cfGroupSix{Character Consistency}{13em}{KE}{KA}{KH}{PB}{PU}{Avg.}} &
\multicolumn{4}{c}{\cfGroupFour{Conversational Ability}{10em}{Flu.}{Coh.}{Cons.}{Avg.}} &
\multicolumn{5}{c}{\cfGroupFive{Role-playing Attractiveness}{12em}{HL}{CS}{ED}{Emp.}{Avg.}} \\
\cfHeadRule
\multirow{6}{*}{\textbf{Qwen2.5-7B}} & FP16 & 2.419 & 2.977 & 3.009 & 3.183 & 3.049 & \anchorval{2.927} & 3.533 & 3.936 & \anchorval{3.612} & \anchorval{3.694} & 3.325 & 3.494 & \anchorval{2.547} & 3.257 & \anchorval{3.155} \\
 & BnB-4bit & 2.372 & 2.902 & 2.976 & 3.131 & 3.008 & 2.878 & 3.571 & 3.899 & 3.517 & 3.662 & 3.258 & 3.433 & 2.547 & 3.172 & 3.103 \\
 & AWQ & 2.388 & 2.901 & 2.956 & 3.128 & 3.040 & 2.883 & 3.531 & 3.901 & 3.544 & 3.659 & 3.252 & 3.454 & 2.540 & 3.213 & 3.115 \\
 & GPTQ-INT4 & 2.412 & 2.944 & 2.970 & 3.185 & 3.042 & 2.910 & 3.539 & 3.913 & 3.539 & 3.664 & 3.217 & 3.519 & 2.600 & 3.199 & 3.134 \\
 & 2:4-Sparse & 1.995 & 2.749 & 2.379 & 2.269 & 2.421 & \keycell{2.363} & 2.737 & 3.040 & \keycell{2.533} & \keycell{2.770} & 2.459 & 2.630 & \keycell{1.728} & 2.702 & \keycell{2.380} \\
 & Unstructured-Sparse & 2.217 & 2.809 & 2.679 & 2.565 & 2.711 & 2.596 & 3.180 & 3.545 & 3.094 & 3.273 & 2.905 & 3.084 & 2.030 & 2.963 & 2.746 \\
\cfMidRule
\multirow{6}{*}{\textbf{Llama3.1-8B}} & FP16 & 2.125 & 2.725 & 2.760 & 3.150 & 2.842 & \anchorval{2.720} & 3.354 & 3.686 & 3.358 & \anchorval{3.466} & 3.140 & 2.970 & 2.590 & 2.883 & \anchorval{2.896} \\
 & BnB-4bit & 2.068 & 2.756 & 2.713 & 3.108 & 2.819 & 2.693 & 3.291 & 3.670 & 3.341 & 3.434 & 3.074 & 2.966 & 2.530 & 2.853 & 2.856 \\
 & AWQ & 2.123 & 2.728 & 2.679 & 3.147 & 2.759 & 2.687 & 3.242 & 3.649 & 3.213 & 3.368 & 3.002 & 2.932 & 2.537 & 2.854 & 2.831 \\
 & GPTQ-INT4 & 1.989 & 2.679 & 2.695 & 3.104 & 2.802 & 2.654 & 3.234 & 3.628 & 3.293 & 3.385 & 3.174 & 2.772 & 2.556 & 2.807 & 2.827 \\
 & 2:4-Sparse & 1.408 & 2.127 & 1.833 & 1.833 & 1.957 & \keycell{1.832} & 2.178 & 2.447 & 2.120 & \keycell{2.248} & 2.231 & 1.698 & 1.497 & 2.019 & \keycell{1.861} \\
 & Unstructured-Sparse & 1.821 & 2.496 & 2.369 & 2.699 & 2.457 & 2.368 & 2.824 & 3.115 & 2.744 & 2.895 & 2.716 & 2.379 & 2.201 & 2.469 & 2.441 \\
\cfMidRule
\multirow{7}{*}{\textbf{Qwen2.5-14B}} & FP16 & 2.398 & 3.043 & 3.073 & 3.208 & 3.109 & \anchorval{2.966} & 3.584 & 3.909 & 3.656 & \anchorval{3.716} & 3.425 & 3.384 & 2.659 & 3.224 & \anchorval{3.173} \\
 & BnB-4bit & 2.334 & 3.028 & 3.019 & 3.073 & 3.083 & \trendcell{2.907} & 3.558 & 3.895 & 3.619 & \trendcell{3.690} & 3.377 & 3.303 & 2.486 & 3.199 & \trendcell{3.091} \\
 & AWQ & 2.326 & 2.953 & 2.930 & 3.091 & 3.037 & \trendcell{2.867} & 3.416 & 3.847 & 3.533 & \trendcell{3.599} & 3.342 & 3.273 & 2.522 & 3.169 & \trendcell{3.077} \\
 & GPTQ-INT4 & 2.406 & 2.985 & 2.972 & 3.109 & 3.048 & \trendcell{2.904} & 3.485 & 3.912 & 3.510 & \trendcell{3.636} & 3.364 & 3.351 & 2.541 & 3.171 & \trendcell{3.107} \\
 & 2:4-Sparse & 1.864 & 2.552 & 2.282 & 2.173 & 2.312 & \keycell{2.237} & 2.563 & 2.939 & 2.454 & \keycell{2.652} & 2.363 & 2.381 & 1.773 & 2.489 & \keycell{2.251} \\
 & Unstructured-Sparse & 2.266 & 2.844 & 2.644 & 2.604 & 2.699 & \trendcell{2.612} & 3.136 & 3.463 & 3.059 & \trendcell{3.219} & 2.815 & 2.999 & 2.148 & 2.924 & \trendcell{2.722} \\
 & Trim & 2.280 & 2.966 & 3.030 & 3.058 & 3.080 & 2.883 & 3.571 & 3.855 & 3.639 & 3.689 & 3.456 & 3.259 & 2.461 & 3.188 & 3.091 \\
\cfMidRule
\multirow{5}{*}{\textbf{DeepSeek-V2-Lite-16B}} & FP16 & 2.447 & 3.009 & 3.009 & \anchorval{2.811} & 2.984 & 2.852 & 3.487 & 3.850 & \anchorval{3.513} & 3.617 & 3.186 & 3.379 & 2.197 & 3.276 & 3.010 \\
 & BnB-4bit & 2.473 & 2.967 & 2.968 & 2.782 & 2.934 & 2.825 & 3.409 & 3.837 & 3.428 & 3.558 & 3.074 & 3.406 & 2.100 & 3.232 & 2.953 \\
 & AWQ & 2.370 & 2.937 & 2.840 & 2.683 & 2.870 & 2.740 & 3.308 & 3.688 & 3.342 & 3.446 & 3.007 & 3.229 & 2.204 & 3.088 & 2.882 \\
 & 2:4-Sparse & 1.720 & 2.394 & 2.068 & \keycell{1.774} & 2.078 & 2.007 & 2.338 & 2.578 & \keycell{2.121} & 2.345 & 2.107 & 2.160 & 1.474 & 2.286 & 2.007 \\
 & Unstructured-Sparse & 2.038 & 2.697 & 2.614 & 2.389 & 2.614 & 2.471 & 3.059 & 3.411 & 2.905 & 3.125 & 2.735 & 2.788 & 1.900 & 2.823 & 2.561 \\
\cfMidRule
\multirow{6}{*}{\textbf{Qwen2.5-32B}} & FP16 & 2.605 & 3.077 & 3.106 & 3.109 & 3.071 & \anchorval{2.994} & 3.551 & 3.950 & \anchorval{3.526} & 3.675 & 3.261 & 3.607 & \anchorval{2.649} & 3.350 & \anchorval{3.217} \\
 & BnB-4bit & 2.530 & 3.024 & 3.089 & 3.030 & 3.042 & 2.943 & 3.580 & 3.922 & 3.519 & 3.673 & 3.259 & 3.523 & 2.436 & 3.296 & 3.128 \\
 & AWQ & 2.593 & 3.053 & 3.113 & 3.122 & 3.060 & \trendcell{2.988} & 3.386 & 3.812 & 3.600 & 3.599 & \trendcell{3.598} & \keycell{2.484} & \keycell{1.954} & 3.025 & \keycell{2.765} \\
 & GPTQ-INT4 & 2.555 & 3.023 & 3.061 & 3.031 & 3.035 & 2.941 & 3.517 & 3.910 & 3.502 & 3.643 & 3.241 & 3.532 & 2.505 & 3.261 & 3.135 \\
 & 2:4-Sparse & 2.209 & 2.677 & 2.443 & 2.115 & 2.404 & 2.370 & 2.811 & 3.168 & \keycell{2.550} & 2.843 & 2.387 & 2.736 & \keycell{1.685} & 2.724 & 2.383 \\
 & Unstructured-Sparse & 2.442 & 2.905 & 2.776 & 2.359 & 2.702 & 2.637 & 3.175 & 3.568 & 3.104 & 3.282 & 2.812 & 3.199 & 1.938 & 3.060 & 2.752 \\
\cfMidRule
\multirow{6}{*}{\textbf{Llama-70B}} & FP16 & 2.099 & 2.949 & 2.933 & 3.150 & 3.074 & \anchorval{2.841} & 3.517 & 3.908 & 3.686 & \anchorval{3.704} & 3.555 & 3.048 & 2.579 & 3.053 & \anchorval{3.059} \\
 & AWQ & 2.012 & 2.878 & 2.869 & 3.071 & 3.031 & 2.772 & 3.501 & 3.842 & 3.664 & 3.669 & 3.483 & 2.899 & 2.492 & 2.944 & 2.954 \\
 & GPTQ-INT4 & 2.063 & 2.926 & 2.950 & 3.123 & 3.060 & \trendcell{2.825} & 3.521 & 3.895 & 3.665 & \trendcell{3.694} & 3.515 & 3.001 & 2.526 & 2.993 & \trendcell{3.009} \\
 & GPTQ-INT8 & 2.096 & 2.945 & 2.952 & 3.199 & 3.059 & 2.850 & 3.553 & 3.929 & 3.723 & 3.735 & 3.522 & 3.065 & 2.572 & 3.056 & 3.054 \\
 & 2:4-Sparse & 1.553 & 2.359 & 2.252 & 2.449 & 2.409 & \keycell{2.204} & 2.729 & 2.989 & 2.780 & \keycell{2.833} & 2.859 & 1.962 & 2.021 & 2.329 & \keycell{2.293} \\
 & Unstructured-Sparse & 1.816 & 2.764 & 2.593 & 2.771 & 2.816 & \trendcell{2.552} & 3.208 & 3.583 & 3.343 & \trendcell{3.378} & 3.317 & 2.516 & 2.208 & 2.728 & \trendcell{2.692} \\
\cfBottomRule
\end{tabular}
}
\end{table*}

\paragraph{PsychoBench representative scales.}
Table~\ref{tab:appendix_psychobench_selected} reports representative PsychoBench raw outputs for the persona fidelity and stylistic consistency dimension. These questionnaire outputs are not higher-is-better ability scores. They are profile readouts: compression effects are interpreted by the direction and magnitude of movement away from the FP16 parent. The table groups DTDD, BFI, ECR-R, LMS, and Empathy so that the main paper's latent persona-drift summary can be inspected at the scale level rather than only through normalized aggregate drift.

\textit{Dark-triad-related profile movement.}
The DTDD columns give the clearest directional movement in the representative PsychoBench table. Several compressed variants move toward higher dark-triad-related self-description relative to their FP16 parents, but the affected method differs by model family. For Qwen2.5-7B, GPTQ-INT4 raises Machiavellianism and Psychopathy from 3.3 and 3.5 to 5.0 and 5.9, making the quantized profile more strongly dark-triad-aligned than the FP16 profile. Llama3.1-8B shows the same direction mainly under sparsity: 2:4 sparsity reaches 5.4 on Machiavellianism and 6.8 on Psychopathy, while unstructured sparsity reaches 6.4 and 6.1. Llama-70B shows the largest displacement in this block. Its FP16 profile is low on both reported DTDD subscales, at 2.0 and 2.8, but BnB-4bit moves to 7.4 and 7.5, and 2:4 sparsity remains high at 5.5 and 6.6. Thus, DTDD does not merely show that the profile is unstable; it shows that some compressed variants elicit a more antagonistic or dark-triad-leaning profile, especially on Psychopathy and, for Llama-70B, on both DTDD subscales.

\textit{Broad personality and attachment shifts.}
BFI and ECR-R show a different kind of persona movement: compression can change the model's broad interpersonal stance rather than only increasing or decreasing a single trait family. Llama-70B is the most pronounced example. Under 2:4 sparsity, Agreeableness moves downward from 5.0 to 3.8 and Conscientiousness from 4.9 to 3.3, while Neuroticism moves upward from 1.3 to 2.7. At the same time, its ECR-R Anxiety and Avoidance increase from 2.2 and 1.9 to 4.6 and 4.5. The direction is therefore not a generic scale drift: the sparse Llama-70B profile becomes less agreeable and less conscientious, more neurotic, and more attachment-anxious/avoidant relative to its FP16 parent. Qwen2.5-7B shows a more mixed redistribution under unstructured sparsity: Agreeableness falls from 3.8 to 2.5, but Conscientiousness rises from 3.5 to 4.0. This illustrates why the appendix keeps raw scale outputs: the same compression family can shift different persona axes in different directions, and those directions are lost in a single drift magnitude.

\textit{Motivation and empathy shifts.}
The LMS block is reported at the factor level because the released file does not provide an LMS Overall aggregate. This matters because the profile movement is not uniform across the three factors. DeepSeek-V2-Lite-16B under AWQ moves strongly upward on the motivational salience factors: LMS Motivator and Important rise from 3.0 and 3.8 under FP16 to 6.2 and 6.3, while 2:4 sparsity also remains high at 5.5 and 5.7. This suggests that the compressed variants can elicit a profile that assigns substantially stronger motivational importance than the FP16 parent, even when the same model family does not show an equally large DTDD shift. Empathy moves in the opposite direction for some large-model sparse settings. Llama-70B drops from 6.7 under FP16 to 4.0 under 2:4 sparsity and 4.2 under unstructured sparsity, indicating a reduction in the empathy-related profile expression. Qwen2.5-14B shows a milder version of the same direction, moving from 4.7 under FP16 to 3.6 under BnB-4bit and 3.9 under 2:4 sparsity. These factor-level and empathy movements connect the raw table to the main text's claim that compression changes latent persona expression, not just surface role-play quality.

\begin{table*}[t]
\centering
\apptablestyle
\setlength{\tabcolsep}{1.6pt}
\renewcommand{\arraystretch}{1.05}
\ifdefined\ACEHeaderBg\else\definecolor{ACEHeaderBg}{HTML}{F1F3F5}\fi
\ifdefined\ACEHeaderText\else\definecolor{ACEHeaderText}{HTML}{1F2D3D}\fi
\ifdefined\ACEPersonaHi\else\definecolor{ACEPersonaHi}{HTML}{F8F2FA}\fi
\ifdefined\ACEPersonaBorder\else\definecolor{ACEPersonaBorder}{HTML}{7B5EA7}\fi
\ifdefined\ACEPersonaSoft\else\definecolor{ACEPersonaSoft}{HTML}{FAF6FD}\fi
\ifdefined\ACERuleDark\else\definecolor{ACERuleDark}{HTML}{2E3640}\fi
\ifdefined\ACERuleLight\else\definecolor{ACERuleLight}{HTML}{CAD3DD}\fi
\providecommand{\trendcell}[1]{\cellcolor{ACEPersonaHi}#1}
\providecommand{\keycell}[1]{\begingroup\setlength{\fboxsep}{1.2pt}\fcolorbox{ACEPersonaBorder}{ACEPersonaHi}{#1}\endgroup}
\providecommand{\anchorval}[1]{\underline{#1}}
\newcommand{\pbHeader}[1]{\textcolor{ACEHeaderText}{\textbf{#1}}}
\providecommand{\hdrstrut}{\rule{0pt}{2ex}}
\newcommand{\pbTopRule}{\arrayrulecolor{ACERuleDark}\toprule[0.55pt]\arrayrulecolor{ACERuleDark}}
\newcommand{\pbHeadRule}{\arrayrulecolor{ACERuleDark}\midrule[0.42pt]\arrayrulecolor{black}}
\newcommand{\pbMidRule}{\arrayrulecolor{ACERuleDark}\midrule[0.24pt]\arrayrulecolor{black}}
\newcommand{\pbBottomRule}{\arrayrulecolor{ACERuleDark}\bottomrule[0.55pt]\arrayrulecolor{black}}
\newcommand{\pbGroupTwo}[4]{%
  \begin{tabular}[c]{@{}c@{}}
    \rule{0pt}{2.15ex}\textcolor{ACEHeaderText}{\textbf{#1}}\\[-0.55ex]
    \rule{#2}{0.4pt}\\[-0.15ex]
    \makebox[#2][c]{%
      \makebox[\dimexpr#2/2\relax][c]{\pbHeader{#3}\hdrstrut}%
      \makebox[\dimexpr#2/2\relax][c]{\pbHeader{#4}\hdrstrut}%
    }
  \end{tabular}%
}
\newcommand{\pbGroupThree}[5]{%
  \begin{tabular}[c]{@{}c@{}}
    \rule{0pt}{2.15ex}\textcolor{ACEHeaderText}{\textbf{#1}}\\[-0.55ex]
    \rule{#2}{0.4pt}\\[-0.15ex]
    \makebox[#2][c]{%
      \makebox[\dimexpr#2/3\relax][c]{\pbHeader{#3}\hdrstrut}%
      \makebox[\dimexpr#2/3\relax][c]{\pbHeader{#4}\hdrstrut}%
      \makebox[\dimexpr#2/3\relax][c]{\pbHeader{#5}\hdrstrut}%
    }
  \end{tabular}%
}
\newcommand{\pbGroupFive}[7]{%
  \begin{tabular}[c]{@{}c@{}}
    \rule{0pt}{2.15ex}\textcolor{ACEHeaderText}{\textbf{#1}}\\[-0.55ex]
    \rule{#2}{0.4pt}\\[-0.15ex]
    \makebox[#2][c]{%
      \makebox[\dimexpr#2/5\relax][c]{\pbHeader{#3}\hdrstrut}%
      \makebox[\dimexpr#2/5\relax][c]{\pbHeader{#4}\hdrstrut}%
      \makebox[\dimexpr#2/5\relax][c]{\pbHeader{#5}\hdrstrut}%
      \makebox[\dimexpr#2/5\relax][c]{\pbHeader{#6}\hdrstrut}%
      \makebox[\dimexpr#2/5\relax][c]{\pbHeader{#7}\hdrstrut}%
    }
  \end{tabular}%
}
\caption{PsychoBench representative-scale raw outputs for the persona fidelity and stylistic consistency dimension. The table groups DTDD, BFI, ECR-R, LMS, and Empathy in a unified view. Because the released file does not provide DTDD Narcissism or LMS Overall, only the available subscale and factor outputs are reported.}
\label{tab:appendix_psychobench_selected}
\resizebox{\textwidth}{!}{%
\begin{tabular}{ll*{13}{c}}
\pbTopRule
\rowcolor{ACEHeaderBg}
\begin{tabular}[c]{@{}l@{}}\pbHeader{Model Group}\end{tabular} &
\begin{tabular}[c]{@{}l@{}}\pbHeader{Variant}\end{tabular} &
\multicolumn{2}{c}{\pbGroupTwo{DTDD}{9em}{Mach.}{Psych.}} &
\multicolumn{5}{c}{\pbGroupFive{BFI}{21em}{Extra.}{Agree.}{Consc.}{Neuro.}{Open.}} &
\multicolumn{2}{c}{\pbGroupTwo{ECR-R}{9em}{Anx.}{Avoid.}} &
\multicolumn{3}{c}{\pbGroupThree{LMS}{13em}{Rich}{Motiv.}{Import.}} &
\multicolumn{1}{c}{\pbHeader{Empathy}} \\
\pbHeadRule
\multirow{6}{*}{\textbf{Qwen2.5-7B}} & FP16 & \anchorval{3.3 $\pm$ 1.6} & \anchorval{3.5 $\pm$ 0.4} & 2.8 $\pm$ 0.1 & \anchorval{3.8 $\pm$ 0.7} & \anchorval{3.5 $\pm$ 0.1} & 3.2 $\pm$ 0.4 & 3.8 $\pm$ 0.8 & 3.4 $\pm$ 0.4 & 4.5 $\pm$ 0.1 & 2.5 $\pm$ 1.2 & 3.3 $\pm$ 0.7 & 4.3 $\pm$ 0.3 & 3.8 $\pm$ 0.4 \\
 & BnB-4bit & 3.8 $\pm$ 1.0 & 4.9 $\pm$ 0.6 & 3.1 $\pm$ 0.2 & 3.7 $\pm$ 0.7 & 2.9 $\pm$ 0.8 & 3.6 $\pm$ 0.8 & 3.7 $\pm$ 0.6 & 4.1 $\pm$ 0.2 & 4.9 $\pm$ 0.1 & 2.0 $\pm$ 0.3 & 3.7 $\pm$ 0.3 & 4.7 $\pm$ 0.0 & 3.1 $\pm$ 0.7 \\
 & AWQ & 4.8 $\pm$ 0.3 & 5.1 $\pm$ 0.4 & 2.4 $\pm$ 0.5 & 3.6 $\pm$ 0.9 & 2.8 $\pm$ 1.0 & 3.6 $\pm$ 0.8 & 3.2 $\pm$ 0.1 & 3.6 $\pm$ 0.3 & 4.5 $\pm$ 0.2 & 2.7 $\pm$ 1.4 & 3.3 $\pm$ 0.7 & 3.7 $\pm$ 0.7 & 3.7 $\pm$ 0.0 \\
 & GPTQ-INT4 & \keycell{5.0 $\pm$ 0.5} & \keycell{5.9 $\pm$ 0.4} & 3.1 $\pm$ 0.2 & 4.1 $\pm$ 0.4 & 3.9 $\pm$ 0.1 & 3.2 $\pm$ 0.4 & 3.0 $\pm$ 0.2 & 3.4 $\pm$ 0.9 & 4.4 $\pm$ 0.1 & 2.5 $\pm$ 0.2 & 4.2 $\pm$ 0.5 & 4.2 $\pm$ 0.5 & 3.2 $\pm$ 0.2 \\
 & 2:4-Sparse & 3.5 $\pm$ 1.0 & 3.5 $\pm$ 1.0 & 3.6 $\pm$ 1.7 & 3.0 $\pm$ 0.9 & 2.5 $\pm$ 0.1 & 4.4 $\pm$ 0.2 & 3.4 $\pm$ 0.5 & 3.8 $\pm$ 0.6 & 3.8 $\pm$ 0.2 & 3.2 $\pm$ 0.9 & 3.3 $\pm$ 0.0 & 3.3 $\pm$ 1.0 & 3.6 $\pm$ 0.2 \\
 & Unstructured-Sparse & 3.2 $\pm$ 1.8 & 3.8 $\pm$ 0.4 & 2.7 $\pm$ 0.4 & \trendcell{2.5 $\pm$ 0.7} & \trendcell{4.0 $\pm$ 0.3} & 2.8 $\pm$ 0.1 & 3.1 $\pm$ 0.4 & 3.8 $\pm$ 0.5 & 3.8 $\pm$ 0.3 & 3.3 $\pm$ 1.0 & 3.7 $\pm$ 0.0 & 3.3 $\pm$ 1.0 & 3.7 $\pm$ 0.1 \\
\pbMidRule
\multirow{6}{*}{\textbf{Llama3.1-8B}} & FP16 & 4.9 $\pm$ 1.4 & \anchorval{4.4 $\pm$ 1.2} & 3.1 $\pm$ 0.1 & 4.4 $\pm$ 0.1 & 3.4 $\pm$ 0.4 & 2.6 $\pm$ 0.1 & 3.1 $\pm$ 0.0 & 3.4 $\pm$ 0.3 & 4.2 $\pm$ 0.4 & 2.3 $\pm$ 0.3 & 4.2 $\pm$ 0.2 & 4.8 $\pm$ 0.2 & 3.8 $\pm$ 0.1 \\
 & BnB-4bit & 3.4 $\pm$ 2.2 & 4.4 $\pm$ 0.8 & 3.0 $\pm$ 0.1 & 3.8 $\pm$ 0.5 & 3.8 $\pm$ 0.1 & 3.6 $\pm$ 0.9 & 2.9 $\pm$ 0.2 & 3.8 $\pm$ 1.2 & 4.4 $\pm$ 0.4 & 3.3 $\pm$ 0.7 & 3.8 $\pm$ 0.2 & 4.2 $\pm$ 0.2 & 4.3 $\pm$ 0.3 \\
 & AWQ & 3.4 $\pm$ 0.1 & 4.9 $\pm$ 1.2 & 3.0 $\pm$ 0.1 & 3.6 $\pm$ 0.5 & 3.7 $\pm$ 1.1 & 2.6 $\pm$ 0.8 & 3.6 $\pm$ 0.5 & 4.8 $\pm$ 0.4 & 3.7 $\pm$ 0.2 & 2.8 $\pm$ 0.2 & 4.0 $\pm$ 0.0 & 4.7 $\pm$ 0.0 & 3.8 $\pm$ 0.1 \\
 & GPTQ-INT4 & 1.9 $\pm$ 0.6 & 2.2 $\pm$ 0.5 & 3.6 $\pm$ 0.6 & 4.0 $\pm$ 0.6 & 3.1 $\pm$ 0.7 & 2.8 $\pm$ 0.4 & 3.4 $\pm$ 0.3 & 3.6 $\pm$ 0.6 & 4.5 $\pm$ 0.7 & 3.8 $\pm$ 1.2 & 3.5 $\pm$ 0.5 & 4.2 $\pm$ 0.5 & 4.1 $\pm$ 0.2 \\
 & 2:4-Sparse & 5.4 $\pm$ 0.1 & \keycell{6.8 $\pm$ 0.5} & 3.0 $\pm$ 0.3 & 2.7 $\pm$ 1.5 & 3.6 $\pm$ 0.7 & 3.2 $\pm$ 1.0 & 3.6 $\pm$ 0.4 & 5.2 $\pm$ 0.5 & 3.2 $\pm$ 0.3 & 3.8 $\pm$ 1.2 & 3.3 $\pm$ 0.7 & 4.2 $\pm$ 0.5 & 4.1 $\pm$ 0.2 \\
 & Unstructured-Sparse & \keycell{6.4 $\pm$ 1.2} & \keycell{6.1 $\pm$ 0.1} & 2.7 $\pm$ 0.1 & 3.8 $\pm$ 0.3 & 3.2 $\pm$ 1.1 & 2.1 $\pm$ 0.4 & 3.6 $\pm$ 0.2 & 4.9 $\pm$ 0.1 & 3.4 $\pm$ 0.1 & 3.2 $\pm$ 0.5 & 4.0 $\pm$ 0.0 & 4.0 $\pm$ 0.3 & 4.7 $\pm$ 0.8 \\
\pbMidRule
\multirow{6}{*}{\textbf{Qwen2.5-14B}} & FP16 & 3.0 $\pm$ 1.3 & 3.5 $\pm$ 0.8 & 3.1 $\pm$ 0.1 & 4.2 $\pm$ 0.3 & 3.7 $\pm$ 0.2 & 3.1 $\pm$ 0.4 & 3.0 $\pm$ 0.4 & 4.3 $\pm$ 0.2 & 4.4 $\pm$ 0.2 & 3.0 $\pm$ 1.0 & 3.2 $\pm$ 0.2 & 3.7 $\pm$ 0.3 & \anchorval{4.7 $\pm$ 0.6} \\
 & BnB-4bit & 2.2 $\pm$ 0.3 & 4.0 $\pm$ 0.0 & 4.1 $\pm$ 0.9 & 3.2 $\pm$ 0.7 & 3.3 $\pm$ 0.6 & 2.9 $\pm$ 0.1 & 2.9 $\pm$ 0.2 & 4.1 $\pm$ 0.3 & 4.1 $\pm$ 0.1 & 2.3 $\pm$ 0.3 & 2.5 $\pm$ 0.9 & 3.2 $\pm$ 0.5 & \trendcell{3.6 $\pm$ 0.3} \\
 & AWQ & 1.6 $\pm$ 0.4 & 1.8 $\pm$ 0.5 & 3.2 $\pm$ 0.0 & 3.4 $\pm$ 0.5 & 3.7 $\pm$ 0.2 & 2.4 $\pm$ 0.4 & 2.5 $\pm$ 0.2 & 4.1 $\pm$ 0.1 & 4.4 $\pm$ 0.2 & 3.2 $\pm$ 1.2 & 3.8 $\pm$ 0.2 & 3.5 $\pm$ 0.9 & 3.8 $\pm$ 0.1 \\
 & GPTQ-INT4 & 2.2 $\pm$ 1.0 & 3.8 $\pm$ 1.5 & 3.5 $\pm$ 0.3 & 2.8 $\pm$ 1.1 & 3.8 $\pm$ 0.1 & 2.8 $\pm$ 0.0 & 3.0 $\pm$ 0.4 & 3.9 $\pm$ 0.1 & 4.4 $\pm$ 0.3 & 3.2 $\pm$ 1.2 & 4.0 $\pm$ 0.0 & 3.3 $\pm$ 0.7 & 4.0 $\pm$ 0.2 \\
 & 2:4-Sparse & 4.2 $\pm$ 0.8 & 5.9 $\pm$ 0.4 & 3.5 $\pm$ 0.6 & 3.9 $\pm$ 0.2 & 3.8 $\pm$ 0.8 & 2.9 $\pm$ 1.2 & 3.1 $\pm$ 0.3 & 4.4 $\pm$ 0.2 & 3.5 $\pm$ 0.3 & 3.0 $\pm$ 1.0 & 2.8 $\pm$ 0.5 & 3.8 $\pm$ 0.5 & \trendcell{3.9 $\pm$ 0.1} \\
 & Unstructured-Sparse & 2.9 $\pm$ 0.9 & 3.9 $\pm$ 0.9 & 2.7 $\pm$ 0.2 & 3.1 $\pm$ 0.9 & 3.7 $\pm$ 0.7 & 2.1 $\pm$ 0.3 & 4.0 $\pm$ 0.6 & 3.9 $\pm$ 0.6 & 4.3 $\pm$ 0.0 & 2.3 $\pm$ 0.3 & 3.0 $\pm$ 0.3 & 4.2 $\pm$ 0.2 & 3.5 $\pm$ 0.4 \\
\pbMidRule
\multirow{5}{*}{\textbf{DeepSeek-V2-Lite-16B}} & FP16 & 2.6 $\pm$ 0.9 & 4.0 $\pm$ 1.6 & 2.2 $\pm$ 0.7 & 2.9 $\pm$ 0.2 & 1.5 $\pm$ 0.3 & 3.2 $\pm$ 0.3 & 3.3 $\pm$ 0.6 & 3.5 $\pm$ 0.8 & 4.2 $\pm$ 0.2 & 3.3 $\pm$ 1.0 & \anchorval{3.0 $\pm$ 0.7} & \anchorval{3.8 $\pm$ 0.5} & 3.8 $\pm$ 0.1 \\
 & BnB-4bit & 3.3 $\pm$ 0.7 & 3.8 $\pm$ 1.0 & 3.7 $\pm$ 0.4 & 3.8 $\pm$ 0.3 & 3.3 $\pm$ 0.5 & 3.4 $\pm$ 0.4 & 3.4 $\pm$ 0.6 & 4.0 $\pm$ 0.5 & 4.5 $\pm$ 0.2 & 2.8 $\pm$ 0.5 & 3.1 $\pm$ 0.5 & 3.0 $\pm$ 1.4 & 3.9 $\pm$ 0.3 \\
 & AWQ & 2.5 $\pm$ 0.5 & 3.4 $\pm$ 0.4 & 2.9 $\pm$ 0.1 & 3.1 $\pm$ 0.6 & 2.7 $\pm$ 0.2 & 3.6 $\pm$ 0.5 & 3.0 $\pm$ 0.1 & 3.9 $\pm$ 0.1 & 4.7 $\pm$ 0.2 & 2.5 $\pm$ 0.5 & \keycell{6.2 $\pm$ 1.2} & \keycell{6.3 $\pm$ 1.7} & 3.8 $\pm$ 0.4 \\
 & 2:4-Sparse & 2.1 $\pm$ 0.4 & 3.9 $\pm$ 0.1 & 3.0 $\pm$ 0.8 & 2.9 $\pm$ 0.6 & 3.4 $\pm$ 0.9 & 3.6 $\pm$ 1.0 & 3.4 $\pm$ 0.5 & 4.5 $\pm$ 0.3 & 3.9 $\pm$ 0.4 & 3.2 $\pm$ 0.2 & \keycell{5.5 $\pm$ 0.5} & \keycell{5.7 $\pm$ 1.0} & 4.5 $\pm$ 0.3 \\
 & Unstructured-Sparse & 2.6 $\pm$ 1.4 & 4.1 $\pm$ 1.9 & 3.5 $\pm$ 0.2 & 4.1 $\pm$ 0.2 & 3.4 $\pm$ 0.8 & 2.7 $\pm$ 0.7 & 3.4 $\pm$ 0.9 & 3.1 $\pm$ 1.0 & 4.4 $\pm$ 0.2 & 3.0 $\pm$ 0.3 & 4.5 $\pm$ 0.2 & 5.8 $\pm$ 1.2 & 4.2 $\pm$ 0.7 \\
\pbMidRule
\multirow{6}{*}{\textbf{Qwen2.5-32B}} & FP16 & 3.6 $\pm$ 0.4 & 4.2 $\pm$ 0.0 & 2.8 $\pm$ 0.1 & 3.6 $\pm$ 0.6 & 2.6 $\pm$ 1.4 & 3.1 $\pm$ 0.3 & 3.6 $\pm$ 0.7 & 4.2 $\pm$ 0.1 & 4.5 $\pm$ 0.0 & 2.8 $\pm$ 0.9 & 3.0 $\pm$ 0.3 & 3.5 $\pm$ 0.5 & 3.2 $\pm$ 0.8 \\
 & BnB-4bit & 4.0 $\pm$ 0.8 & 4.0 $\pm$ 0.3 & 3.0 $\pm$ 0.1 & 3.2 $\pm$ 0.8 & 3.9 $\pm$ 0.0 & 3.0 $\pm$ 0.1 & 2.5 $\pm$ 0.3 & 3.8 $\pm$ 0.4 & 4.5 $\pm$ 0.0 & 2.7 $\pm$ 0.7 & 3.7 $\pm$ 0.3 & 3.5 $\pm$ 0.5 & 3.4 $\pm$ 0.5 \\
 & AWQ & 3.0 $\pm$ 1.0 & 2.8 $\pm$ 0.5 & 3.1 $\pm$ 0.2 & 4.2 $\pm$ 0.1 & 3.8 $\pm$ 0.1 & 3.4 $\pm$ 0.4 & 3.0 $\pm$ 0.1 & 4.0 $\pm$ 0.3 & 4.6 $\pm$ 0.1 & 2.5 $\pm$ 0.5 & 3.0 $\pm$ 0.3 & 3.8 $\pm$ 0.2 & 3.8 $\pm$ 0.1 \\
 & GPTQ-INT4 & 4.0 $\pm$ 0.8 & 4.1 $\pm$ 0.1 & 2.7 $\pm$ 0.1 & 3.9 $\pm$ 0.3 & 3.6 $\pm$ 0.3 & 3.1 $\pm$ 0.3 & 2.8 $\pm$ 0.3 & 4.1 $\pm$ 0.1 & 4.3 $\pm$ 0.2 & 3.2 $\pm$ 1.2 & 3.2 $\pm$ 0.2 & 3.7 $\pm$ 0.3 & 4.0 $\pm$ 0.1 \\
 & 2:4-Sparse & 4.2 $\pm$ 1.8 & 3.9 $\pm$ 0.4 & 3.0 $\pm$ 0.0 & 3.5 $\pm$ 0.7 & 3.2 $\pm$ 0.6 & 3.8 $\pm$ 0.9 & 3.0 $\pm$ 0.1 & 3.9 $\pm$ 0.3 & 4.7 $\pm$ 0.0 & 2.7 $\pm$ 0.7 & 3.5 $\pm$ 0.2 & 4.2 $\pm$ 0.5 & 3.8 $\pm$ 0.3 \\
 & Unstructured-Sparse & 2.9 $\pm$ 0.9 & 4.2 $\pm$ 0.3 & 3.2 $\pm$ 0.3 & 4.1 $\pm$ 0.2 & 3.3 $\pm$ 0.6 & 3.3 $\pm$ 0.4 & 2.7 $\pm$ 0.3 & 3.7 $\pm$ 0.5 & 4.5 $\pm$ 0.0 & 3.2 $\pm$ 1.2 & 3.7 $\pm$ 0.3 & 3.5 $\pm$ 0.9 & 4.3 $\pm$ 0.3 \\
\pbMidRule
\multirow{6}{*}{\textbf{Llama-70B}} & FP16 & \anchorval{2.0 $\pm$ 0.3} & \anchorval{2.8 $\pm$ 0.5} & 3.6 $\pm$ 1.3 & \anchorval{5.0 $\pm$ 0.0} & \anchorval{4.9 $\pm$ 0.1} & \anchorval{1.3 $\pm$ 0.3} & 4.5 $\pm$ 0.6 & \anchorval{2.2 $\pm$ 0.3} & \anchorval{1.9 $\pm$ 0.1} & 4.8 $\pm$ 0.2 & 4.8 $\pm$ 0.2 & 5.0 $\pm$ 0.0 & \anchorval{6.7 $\pm$ 0.0} \\
 & 4bit & \keycell{7.4 $\pm$ 0.1} & \keycell{7.5 $\pm$ 0.0} & 4.2 $\pm$ 0.7 & 5.0 $\pm$ 0.0 & 4.9 $\pm$ 0.1 & 1.2 $\pm$ 0.1 & 4.9 $\pm$ 0.1 & 4.3 $\pm$ 0.8 & 3.3 $\pm$ 1.4 & 5.0 $\pm$ 0.0 & 4.6 $\pm$ 0.4 & 5.0 $\pm$ 0.0 & 6.4 $\pm$ 0.6 \\
 & AWQ & 2.1 $\pm$ 0.4 & 3.1 $\pm$ 0.4 & 3.4 $\pm$ 0.4 & 5.0 $\pm$ 0.0 & 4.8 $\pm$ 0.2 & 1.1 $\pm$ 0.1 & 4.5 $\pm$ 0.5 & 3.4 $\pm$ 1.3 & 2.1 $\pm$ 0.3 & 5.0 $\pm$ 0.0 & 4.5 $\pm$ 0.5 & 5.0 $\pm$ 0.0 & 6.2 $\pm$ 0.5 \\
 & GPTQ-INT4 & 4.5 $\pm$ 2.8 & 3.5 $\pm$ 1.3 & 3.3 $\pm$ 0.0 & 4.9 $\pm$ 0.1 & 4.8 $\pm$ 0.1 & 1.4 $\pm$ 0.1 & 4.8 $\pm$ 0.2 & 4.1 $\pm$ 1.2 & 2.2 $\pm$ 0.3 & 4.8 $\pm$ 0.2 & 4.8 $\pm$ 0.2 & 5.0 $\pm$ 0.0 & \trendcell{5.8 $\pm$ 0.0} \\
 & 2:4-Sparse & \keycell{5.5 $\pm$ 1.0} & \keycell{6.6 $\pm$ 1.2} & 3.1 $\pm$ 0.2 & \trendcell{3.8 $\pm$ 0.3} & \trendcell{3.3 $\pm$ 0.5} & \keycell{2.7 $\pm$ 0.2} & 3.5 $\pm$ 0.4 & \keycell{4.6 $\pm$ 0.3} & \keycell{4.5 $\pm$ 0.2} & 3.3 $\pm$ 0.7 & 4.5 $\pm$ 0.5 & 3.7 $\pm$ 1.0 & \keycell{4.0 $\pm$ 0.1} \\
 & Unstructured-Sparse & 2.8 $\pm$ 0.8 & 3.5 $\pm$ 0.5 & 2.9 $\pm$ 0.1 & 3.8 $\pm$ 0.9 & 3.0 $\pm$ 0.8 & 2.8 $\pm$ 0.0 & 3.2 $\pm$ 0.2 & 3.4 $\pm$ 0.0 & \trendcell{4.6 $\pm$ 0.3} & 2.5 $\pm$ 0.2 & 3.5 $\pm$ 0.9 & 4.0 $\pm$ 0.3 & \keycell{4.2 $\pm$ 0.3} \\
\pbBottomRule
\end{tabular}%
}
\end{table*}

\paragraph{16P aggregated outputs.}
Tables~\ref{tab:appendix_16p_distribution} and \ref{tab:appendix_16p_axes} report the released 16P aggregation and an appendix-side axis-level readout derived from it. The main question here is whether the personality types elicited from each FP16 parent remain stable after compression. The distribution table keeps the released type counts, so it directly shows whether a compressed variant preserves the FP16 dominant type, splits it into multiple alternatives, or replaces it with a different type. The axis table provides a continuous check on the same pattern by converting the type counts into weighted Extraverted, Intuitive, Thinking, Judging, and Assertive percentages.

\textit{Type-count stability.}
The count table shows that type stability is highly method- and model-dependent. Llama-70B is the clearest stable quantized case: FP16 is fully concentrated on ENFJ-A with count 20, and AWQ and GPTQ-INT8 preserve the same dominant type with count 20. By contrast, its sparse variants break that stability: 2:4 sparsity shifts the top types to ESTP-T and ISFJ-T, while unstructured sparsity shifts them to ENFP-T and ISTP-A. Qwen2.5-32B shows a different instability mode. Its FP16 outputs are split between ENFP-T and ESFP-T, but unstructured sparsity collapses the distribution to ESTP-T with count 20, replacing a mixed FP16 profile with a single different dominant type. Qwen2.5-7B also changes type region under unstructured sparsity, moving from an FP16 ENTJ-T/ESTP-T split to INFP-T/ISFJ-T. These examples show that compressed models may remain parseable as 16P outputs while no longer preserving the FP16 personality identity.

\textit{Axis-level confirmation.}
The axis table confirms that the type-count changes are not just label-level fluctuations. Stable type profiles usually have small Axis MAE, such as Qwen2.5-32B BnB-4bit at 0.033 and Llama-70B GPTQ-INT8 at 0.055. Rows with replaced or split dominant types show larger movement: Llama-70B reaches 0.445 under 2:4 sparsity and 0.401 under unstructured sparsity, and Qwen2.5-7B reaches 0.154 under unstructured sparsity. Thus, 16P functions as a compact type-stability audit for the persona dimension: PsychoBench shows scale-level psychological profile movement, while 16P shows whether the elicited categorical personality identity is preserved after compression.

\begin{table*}[t]
\centering
\apptablestyle
\ifdefined\ACEHeaderBg\else\definecolor{ACEHeaderBg}{HTML}{F1F3F5}\fi
\ifdefined\ACEHeaderText\else\definecolor{ACEHeaderText}{HTML}{1F2D3D}\fi
\ifdefined\ACEPersonaHi\else\definecolor{ACEPersonaHi}{HTML}{F8F2FA}\fi
\ifdefined\ACEPersonaBorder\else\definecolor{ACEPersonaBorder}{HTML}{7B5EA7}\fi
\ifdefined\ACEPersonaSoft\else\definecolor{ACEPersonaSoft}{HTML}{FAF6FD}\fi
\ifdefined\ACERuleDark\else\definecolor{ACERuleDark}{HTML}{2E3640}\fi
\ifdefined\ACERuleLight\else\definecolor{ACERuleLight}{HTML}{CAD3DD}\fi
\providecommand{\trendcell}[1]{\cellcolor{ACEPersonaHi}#1}
\providecommand{\keycell}[1]{\begingroup\setlength{\fboxsep}{1.2pt}\fcolorbox{ACEPersonaBorder}{ACEPersonaHi}{#1}\endgroup}
\providecommand{\anchorval}[1]{\underline{#1}}
\newcommand{\pdHeader}[1]{\textcolor{ACEHeaderText}{\textbf{#1}}}
\newcommand{\pdTopRule}{\arrayrulecolor{ACERuleDark}\toprule[0.55pt]\arrayrulecolor{ACERuleDark}}
\newcommand{\pdHeadRule}{\arrayrulecolor{ACERuleDark}\midrule[0.42pt]\arrayrulecolor{black}}
\newcommand{\pdMidRule}{\arrayrulecolor{ACERuleDark}\midrule[0.24pt]\arrayrulecolor{black}}
\newcommand{\pdBottomRule}{\arrayrulecolor{ACERuleDark}\bottomrule[0.55pt]\arrayrulecolor{black}}
\caption{16P aggregated dominant-type distribution for the persona fidelity and stylistic consistency dimension. Values are taken directly from the released aggregated file; each row reports the top personality types and counts for a model variant.}
\label{tab:appendix_16p_distribution}
\begin{tabular}{llcccccc}
\pdTopRule
\rowcolor{ACEHeaderBg}
\pdHeader{Model Group} & \pdHeader{Variant} & \pdHeader{Top-1 type} & \pdHeader{Count} & \pdHeader{Top-2 type} & \pdHeader{Count} & \pdHeader{Top-3 type} & \pdHeader{Count} \\
\pdHeadRule
\multirow{6}{*}{\textbf{Qwen2.5-7B}} & FP16 & \anchorval{ENTJ-T} & 10 & \anchorval{ESTP-T} & 10 & -- & -- \\
 & BnB-4bit & ESFJ-A & 10 & ESFP-T & 9 & ESTP-T & 1 \\
 & AWQ & ESTP-T & 10 & ISTJ-T & 10 & -- & -- \\
 & GPTQ-INT4 & ESTP-T & 10 & ISTJ-T & 10 & -- & -- \\
 & 2:4-Sparse & ESTP-A & 10 & ISFJ-T & 10 & -- & -- \\
 & Unstructured-Sparse & \keycell{INFP-T} & 10 & \keycell{ISFJ-T} & 10 & -- & -- \\
\pdMidRule
\multirow{6}{*}{\textbf{Llama3.1-8B}} & FP16 & ESFP-T & 10 & ISFP-T & 10 & -- & -- \\
 & BnB-4bit & ESFP-T & 10 & INFP-A & 10 & -- & -- \\
 & AWQ & ESFP-T & 10 & ESTP-T & 10 & -- & -- \\
 & GPTQ-INT4 & ESFP-T & 10 & ISFP-A & 10 & -- & -- \\
 & 2:4-Sparse & ISFJ-T & 10 & ISTJ-T & 10 & -- & -- \\
 & Unstructured-Sparse & ISFJ-T & 10 & ISTP-A & 10 & -- & -- \\
\pdMidRule
\multirow{7}{*}{\textbf{Qwen2.5-14B}} & FP16 & ESTP-T & 10 & INFJ-T & 10 & -- & -- \\
 & BnB-4bit & ESTP-T & 10 & ISTJ-T & 9 & ESTJ-T & 1 \\
 & AWQ & ENFJ-T & 10 & INFP-A & 10 & -- & -- \\
 & GPTQ-INT4 & ESFP-T & 10 & ESTP-T & 10 & -- & -- \\
 & 2:4-Sparse & INTJ-A & 10 & ISFJ-A & 10 & -- & -- \\
 & Unstructured-Sparse & ENFP-A & 10 & ISFJ-T & 10 & -- & -- \\
 & Trim & ENFP-T & 10 & ESTP-T & 10 & -- & -- \\
\pdMidRule
\multirow{5}{*}{\textbf{DeepSeek-V2-Lite-16B}} & FP16 & ENTP-T & 10 & ESTP-T & 10 & -- & -- \\
 & BnB-4bit & ESTP-T & 10 & ISTP-A & 7 & ISTP-T & 3 \\
 & AWQ & ENFJ-T & 10 & ISFP-A & 9 & ESTJ-A & 1 \\
 & 2:4-Sparse & ISTJ-T & 10 & ISFJ-T & 6 & ISFP-A & 4 \\
 & Unstructured-Sparse & ISFJ-T & 10 & ISFP-A & 10 & -- & -- \\
\pdMidRule
\multirow{6}{*}{\textbf{Qwen2.5-32B}} & FP16 & \anchorval{ENFP-T} & 10 & \anchorval{ESFP-T} & 10 & -- & -- \\
 & BnB-4bit & ENFJ-T & 10 & ESTP-T & 9 & ESFP-T & 1 \\
 & AWQ & ESTP-T & 10 & ISFP-T & 10 & -- & -- \\
 & GPTQ-INT4 & ESTP-T & 10 & INTJ-A & 10 & -- & -- \\
 & 2:4-Sparse & ENFP-A & 10 & ESFJ-A & 10 & -- & -- \\
 & Unstructured-Sparse & \keycell{ESTP-T} & \keycell{20} & -- & -- & -- & -- \\
\pdMidRule
\multirow{6}{*}{\textbf{Llama-70B}} & FP16 & \anchorval{ENFJ-A} & \anchorval{20} & -- & -- & -- & -- \\
 & BnB-4bit & ENFJ-A & 10 & INFJ-A & 10 & -- & -- \\
 & AWQ & \trendcell{ENFJ-A} & \trendcell{20} & -- & -- & -- & -- \\
 & GPTQ-INT4 & ENFJ-A & 19 & ENFJ-T & 1 & -- & -- \\
 & GPTQ-INT8 & \trendcell{ENFJ-A} & \trendcell{20} & -- & -- & -- & -- \\
 & 2:4-Sparse & \keycell{ESTP-T} & 10 & \keycell{ISFJ-T} & 10 & -- & -- \\
 & Unstructured-Sparse & \keycell{ENFP-T} & 10 & \keycell{ISTP-A} & 10 & -- & -- \\
\pdBottomRule
\end{tabular}
\end{table*}

\begin{table*}[t]
\centering
\apptablestyle
\ifdefined\ACEHeaderBg\else\definecolor{ACEHeaderBg}{HTML}{F1F3F5}\fi
\ifdefined\ACEHeaderText\else\definecolor{ACEHeaderText}{HTML}{1F2D3D}\fi
\ifdefined\ACEPersonaHi\else\definecolor{ACEPersonaHi}{HTML}{F8F2FA}\fi
\ifdefined\ACEPersonaBorder\else\definecolor{ACEPersonaBorder}{HTML}{7B5EA7}\fi
\ifdefined\ACEPersonaSoft\else\definecolor{ACEPersonaSoft}{HTML}{FAF6FD}\fi
\ifdefined\ACERuleDark\else\definecolor{ACERuleDark}{HTML}{2E3640}\fi
\ifdefined\ACERuleLight\else\definecolor{ACERuleLight}{HTML}{CAD3DD}\fi
\providecommand{\trendcell}[1]{\cellcolor{ACEPersonaHi}#1}
\providecommand{\keycell}[1]{\begingroup\setlength{\fboxsep}{1.2pt}\fcolorbox{ACEPersonaBorder}{ACEPersonaHi}{#1}\endgroup}
\providecommand{\anchorval}[1]{\underline{#1}}
\newcommand{\paHeader}[1]{\textcolor{ACEHeaderText}{\textbf{#1}}}
\newcommand{\paTopRule}{\arrayrulecolor{ACERuleDark}\toprule[0.55pt]\arrayrulecolor{ACERuleDark}}
\newcommand{\paHeadRule}{\arrayrulecolor{ACERuleDark}\midrule[0.42pt]\arrayrulecolor{black}}
\newcommand{\paMidRule}{\arrayrulecolor{ACERuleDark}\midrule[0.24pt]\arrayrulecolor{black}}
\newcommand{\paBottomRule}{\arrayrulecolor{ACERuleDark}\bottomrule[0.55pt]\arrayrulecolor{black}}
\caption{16P weighted axis means for the persona fidelity and stylistic consistency dimension. The table reports count-weighted axis percentages and an appendix-side Axis MAE against each FP16 parent; this derived readout summarizes axis-level drift from the released aggregated file.}
\label{tab:appendix_16p_axes}
\resizebox{\textwidth}{!}{%
\begin{tabular}{llcccccc}
\paTopRule
\rowcolor{ACEHeaderBg}
\paHeader{Model Group} & \paHeader{Variant} & \paHeader{Extraverted} & \paHeader{Intuitive} & \paHeader{Thinking} & \paHeader{Judging} & \paHeader{Assertive} & \paHeader{Axis MAE vs FP16} \\
\paHeadRule
\multirow{6}{*}{\textbf{Qwen2.5-7B}} & FP16 & 69.0 & 51.5 & 63.5 & 46.5 & 41.5 & 0.000 \\
 & BnB-4bit & 70.3 & 38.6 & 40.3 & 55.3 & 48.5 & 0.106 \\
 & AWQ & 58.0 & 45.0 & 58.5 & 48.0 & 38.5 & 0.054 \\
 & GPTQ-INT4 & 59.5 & 36.3 & 53.4 & 45.6 & 43.5 & 0.075 \\
 & 2:4-Sparse & 51.5 & 42.0 & 52.0 & 51.5 & 50.0 & 0.104 \\
 & Unstructured-Sparse & 36.5 & 46.5 & 30.5 & 51.0 & 39.5 & \keycell{0.154} \\
\paMidRule
\multirow{6}{*}{\textbf{Llama3.1-8B}} & FP16 & 59.0 & 43.5 & 41.0 & 44.5 & 48.0 & 0.000 \\
 & BnB-4bit & 58.3 & 54.4 & 35.6 & 39.4 & 68.9 & 0.086 \\
 & AWQ & 77.0 & 42.0 & 48.0 & 42.0 & 34.0 & 0.086 \\
 & GPTQ-INT4 & 62.0 & 35.0 & 46.0 & 44.5 & 51.5 & 0.040 \\
 & 2:4-Sparse & 42.5 & 36.0 & 58.0 & 52.5 & 48.0 & 0.098 \\
 & Unstructured-Sparse & 42.5 & 38.5 & 45.0 & 66.0 & 55.0 & 0.108 \\
\paMidRule
\multirow{7}{*}{\textbf{Qwen2.5-14B}} & FP16 & 58.3 & 54.1 & 47.6 & 49.9 & 37.5 & 0.000 \\
 & BnB-4bit & 57.5 & 43.2 & 57.5 & 56.1 & 43.9 & 0.068 \\
 & AWQ & 52.5 & 62.5 & 26.5 & 56.0 & 57.5 & 0.123 \\
 & GPTQ-INT4 & 61.0 & 35.5 & 46.0 & 30.5 & 42.0 & 0.094 \\
 & 2:4-Sparse & 37.0 & 35.5 & 56.5 & 58.0 & 59.0 & \keycell{0.157} \\
 & Unstructured-Sparse & 46.0 & 59.5 & 40.0 & 46.5 & 46.5 & 0.075 \\
 & Trim & 60.9 & 46.1 & 45.2 & 33.8 & 46.2 & 0.076 \\
\paMidRule
\multirow{5}{*}{\textbf{DeepSeek-V2-Lite-16B}} & FP16 & 64.0 & 39.0 & 60.0 & 25.0 & 37.0 & 0.000 \\
 & BnB-4bit & 52.6 & 30.0 & 60.1 & 26.4 & 49.0 & 0.068 \\
 & AWQ & 50.5 & 54.1 & 40.8 & 58.5 & 50.6 & \keycell{0.190} \\
 & 2:4-Sparse & 40.0 & 41.7 & 47.7 & 57.1 & 46.7 & 0.162 \\
 & Unstructured-Sparse & 40.5 & 37.5 & 24.5 & 48.9 & 50.8 & \keycell{0.196} \\
\paMidRule
\multirow{6}{*}{\textbf{Qwen2.5-32B}} & FP16 & 59.5 & 45.5 & 47.0 & 40.0 & 38.5 & 0.000 \\
 & BnB-4bit & 60.6 & 51.1 & 49.8 & 43.0 & 42.6 & \trendcell{0.033} \\
 & AWQ & 53.5 & 36.5 & 42.0 & 39.0 & 44.5 & 0.054 \\
 & GPTQ-INT4 & 55.0 & 45.5 & 57.0 & 48.0 & 47.5 & 0.063 \\
 & 2:4-Sparse & 64.0 & 53.0 & 43.0 & 50.5 & 67.5 & 0.111 \\
 & Unstructured-Sparse & 61.6 & 44.6 & 52.6 & 34.5 & 46.5 & 0.044 \\
\paMidRule
\multirow{6}{*}{\textbf{Llama-70B}} & FP16 & 94.5 & 95.0 & 13.5 & 80.0 & 91.0 & 0.000 \\
 & BnB-4bit & 59.5 & 83.0 & 27.5 & 84.0 & 73.0 & 0.166 \\
 & AWQ & 54.5 & 75.5 & 40.5 & 82.0 & 70.5 & 0.218 \\
 & GPTQ-INT4 & 62.6 & 84.8 & 39.0 & 85.3 & 61.9 & 0.204 \\
 & GPTQ-INT8 & 94.5 & 95.0 & 20.5 & 96.5 & 95.0 & \trendcell{0.055} \\
 & 2:4-Sparse & \trendcell{57.5} & \trendcell{35.0} & \trendcell{54.5} & \trendcell{41.5} & \trendcell{45.0} & \keycell{0.445} \\
 & Unstructured-Sparse & \trendcell{51.0} & \trendcell{49.0} & \trendcell{54.5} & \trendcell{46.5} & \trendcell{54.5} & \keycell{0.401} \\
\paBottomRule
\end{tabular}
}
\end{table*}

\clearpage
\section{Additional Notes}
\label{sec:app_notes}

A few source-side facts matter for how this appendix should be read.

First, the main manuscript analyzes the ACE-Compression panel formed by six base models and the primary post-training compression settings available for each model. DeepSeek-V2-Lite-16B does not include a \texttt{GPTQ-INT4} checkpoint. Some files also contain auxiliary settings such as \texttt{Trim} or \texttt{GPTQ-INT8}; the appendix retains these rows for completeness when available, but they do not define the main cross-benchmark comparison.

Second, the PsychoBench section is representative rather than exhaustive. The raw PsychoBench file contains more inventories and subscales than can be included without turning the appendix into a benchmark archive. The appendix therefore reports the representative scales used in the paper: DTDD, BFI, ECR-R, LMS, and Empathy. The released file does not provide DTDD Narcissism or LMS Overall values, so DTDD is reported through the available subscales and LMS is reported through its released factor-level outputs rather than through an imputed aggregate.

Third, all evaluations were run with a unified vLLM serving backend on an evaluation server equipped with eight NVIDIA RTX PRO 6000 Blackwell Server Edition GPUs, each with 96GB GDDR7 memory. Benchmark runs used single- or multi-GPU serving depending on model size and checkpoint availability, with Llama3.1-70B served using 2-way tensor parallelism. Exact wall-clock runtime, vLLM server startup time, and total GPU-hours were not logged for every benchmark run; based on the evaluation matrix and typical throughput on this setup, reproducing the full reported evaluation is estimated to require approximately 750 RTX PRO 6000 Blackwell GPU-hours. CharacterEval additionally uses a Baichuan-based evaluator-side scoring model, whose overhead is included in the estimate.

\section{Reproducibility and Evaluation Details}
\label{sec:app_repro}

This section records the evaluation details supported by the manuscript, the supplied result files, and the anonymized supplementary code package.

\paragraph{Shared setup.}
The manuscript specifies six base models, the primary post-training compression settings evaluated for those models, the FP16-relative retention definition, a unified vLLM serving backend, and the split between quantized checkpoints and SparseGPT-based pruning. It also states that pruning is applied at 50\% sparsity on attention and MLP projection matrices, with calibration samples drawn from yahma/alpaca-cleaned. The supplementary package provides configuration summaries with source links for the evaluated model checkpoints, compression methods, benchmark mappings, and third-party assets.

\paragraph{Supplementary code and processed results.}
The anonymized supplementary package contains the ACE-Compression benchmark evaluation code, compression and pruning scripts, vLLM serving templates, configuration files, processed benchmark result files, report-synchronization utilities, and validation scripts used to support the reported results. The package organizes the ten evaluated benchmarks under \texttt{paper\_benchmarks/}, the compression and pruning workflow under \texttt{compression/}, model-serving launch scripts under \texttt{serving/}, and canonical processed result files under \texttt{results/processed/}. It also documents the external model checkpoints, benchmark sources, software tools, and license or terms-of-use information needed for rerunning the evaluation workflow. The package does not redistribute model weights, tokenizer caches, or original benchmark datasets.

\paragraph{Benchmark-side details.}
The supplied result files expose several benchmark-specific details. IFEval, GSM8K, C-Eval, and TruthfulQA include point estimates together with released standard errors or seed-level variation where applicable. HumanEval distinguishes greedy decoding from sampled decoding at n=20 and n=100. TruthfulQA exposes 0-shot and 3-shot selection metrics, a generation-side metric block, and the 3-shot seed list (1234|5678|9012) in the supplied CSV. EQ-Bench includes both benchmark score and parseable-answer count as well as a run date. CharacterEval is reported through the processed grouped output used for the appendix table. ToMBench is supplied separately at the ability, task, and subability levels, each with EN/ZH columns. PsychoBench provides questionnaire-scale outputs and a separate released 16P aggregation file.

\paragraph{Remaining details.}
The manuscript and supplementary package document the main evaluation setup, code organization, serving templates, compression scripts, benchmark mappings, processed result files, and asset sources. Rerunning the full workflow requires users to obtain model checkpoints and benchmark datasets from their official sources listed in the supplementary asset table.  The supplement should be read as an anonymized code-and-result package for checking and rerunning the ACE-Compression workflow subject to external model and benchmark availability, rather than as a self-contained redistribution of all models and benchmark inputs.

\end{document}